\documentclass[11pt,letterpaper,dvipsnames]{article}
\usepackage{epsfig}
\usepackage[margin=1in]{geometry}
\usepackage{graphicx} 
\usepackage[utf8]{inputenc}
\usepackage{esvect}
\usepackage{fancyhdr}
\usepackage{enumitem}
\usepackage{tcolorbox}
\usepackage[T1]{fontenc}
\usepackage{calc}
\usepackage{subfiles}
\usepackage{pbsi}
\usepackage{amsmath,amssymb}
\usepackage{blindtext}
\usepackage{booktabs}
\usepackage{amsthm, amsmath, amssymb}
\usepackage{indentfirst}
\usepackage[affil-it]{authblk}
\usepackage{xfrac}
\usepackage{bm}
\usepackage{multirow}
\usepackage{hhline}
\usepackage{float}
\usepackage{tikz}
\usepackage{pgfplots}
\usetikzlibrary{shapes,arrows,positioning}
\usepackage[nottoc, numbib, notlot]{tocbibind}
\usepackage{xcolor}
\usepackage{lmodern}
\usepackage{graphicx}
\usepackage{mathtools}
\usepackage{lastpage}
\usepackage{titling}
\usepackage{subcaption}  
\usepackage{algpseudocode}
\usepackage{relsize}
\usepackage[ruled,vlined,linesnumbered]{algorithm2e}
\usepackage{array}
\usepackage{booktabs} 
\usepackage{calc}
\usepackage{amsthm}
\usepackage{pgfplots} 
\usepackage{float}
\usepackage[font=small,skip=1pt]{caption}
\usepackage{chngcntr}
\usepackage[authoryear,round]{natbib}
\usepackage[colorlinks=true,allcolors=teal]{hyperref}

\counterwithout{equation}{section}      

\hypersetup{
    colorlinks=true,
    linkcolor=violet,
    filecolor=magenta,      
    urlcolor=red,
    citecolor=blue,
    linktoc=all
}

\makeatletter
\newcommand{\mytag}[2]{
  \text{#1}
  \@bsphack
  \begingroup
    \@onelevel@sanitize\@currentlabelname
    \edef\@currentlabelname{
      \expandafter\strip@period\@currentlabelname\relax.\relax\@@@
    }
    \protected@write\@auxout{}{
      \string\newlabel{#2}{
        {#1}
        {\thepage}
        {\@currentlabelname}
        {\@currentHref}{}
      }
    }
  \endgroup
  \@esphack
}
\makeatother

\newtheorem{theorem}{Theorem}
\newtheorem{corollary}{Corollary}
\newtheorem{lemma}{Lemma}
\theoremstyle{definition}
\newtheorem{definition}{Definition}

\newtheorem{prop}{Proposition}

\newtheorem*{lemma*}{Lemma}

\definecolor{mypurple}{RGB}{120,80,200}

\newcommand{\prob}{\mathrm{P}}
\newcommand{\Ex}{\mathbf{E}}
\newcommand{\real}{\mathbb{R}}
\newcommand{\Fil}{\mathcal{F}}

\newcommand{\Exs}{\mathbb{E}}

\newcommand{\bound}{\lambda_T}
\newcommand{\Z}{Z}
\newcommand{\z}{z}
\newcommand{\Tfrac}{\frac{T}{4}}
\newcommand{\Rew}{X}
\newcommand{\X}{V}
\newcommand{\Ztwo}{L}
\newcommand{\Algo}{\mathcal{A}}
\newcommand{\W}{V}

\newcommand{\minimaxfam}{\texttt{MOSS-family~}}
\newcommand{\ucb}{UCB-1~}
\newcommand{\taileve}{\mathcal{E}_T}
\newcommand{\vopt}{\operatorname{Var}_G}
\newcommand{\Eone}{\mathcal{E}_1(T)}
\newcommand{\Ezero}{\mathcal{E}_0(T)}
\newcommand{\E}{\mathcal{E}(T)}
\newcommand{\algo}{$\mathcal{A}$~}

\usepackage{enumitem}

\colorlet{shadecolor}{gray!10}
\colorlet{lightyellow}{yellow!10}

\title{ \textbf{On Instability of Minimax Optimal Optimism-Based Bandit Algorithms}}

\author{
\large Samya Praharaj$^{\dagger}$ \quad
Koulik Khamaru$^{\dagger}$ \\
\vspace{0.2cm}
\normalsize $^{\dagger}$Department of Statistics, Rutgers University \\
}

\begin{document}

\maketitle

\begin{abstract}
\noindent
Statistical inference from data generated by multi-armed bandit (MAB) algorithms is challenging due to their adaptive, non-i.i.d.\ nature. A classical manifestation is that sample averages of arm-rewards under bandit sampling may fail to satisfy a central limit theorem. Lai and Wei’s \emph{stability} condition~\cite{lai1982least} provides a sufficient—and essentially necessary—criterion for asymptotic normality in bandit problems. While the celebrated Upper Confidence Bound (UCB) algorithm satisfies this stability condition, it is not minimax-optimal, raising the question of whether minimax optimality and statistical stability can be achieved simultaneously. In this paper, we analyze the stability properties of a broad class of bandit algorithms that are based on the \emph{optimism-principle}. We establish general structural conditions under which such algorithms violate the Lai–Wei stability criterion. As a consequence, we show that widely used minimax-optimal UCB-style algorithms—including MOSS, Anytime-MOSS, Vanilla-MOSS, ADA-UCB, OC-UCB, KL-MOSS, KL-UCB++, KL-UCB-SWITCH and Anytime KL-UCB-SWITCH—are unstable. We further complement our theoretical results with numerical simulations demonstrating that, in all these cases, the sample means fail to exhibit asymptotic normality. 

Overall, our findings suggest a fundamental tension between stability and minimax-optimal regret, raising the question of whether it is possible to design bandit algorithms that achieve both. Understanding whether such simultaneously stable and minimax-optimal strategies exist remains an important open direction. 
\end{abstract}

\section{Introduction}

\noindent Sequential inference (\citet{wald1992sequential}, \citet{lai1985asymptotically}, \citet{siegmund2013sequential}) and online decision making (\citet{supancic2017tracking}, \citet{bastani2025improving}, \citet{bastani2020online}) concern the study of how learners update their beliefs and choose actions in real time as new information arrives. Unlike batch settings, where all data is available at once, sequential settings require processing streaming data efficiently, often under uncertainty and resource constraints. Multi-armed bandit algorithms (MAB) are a class of online strategies that have found wide applicability across a diverse range of domains which include online advertising (\citet{wen2017online}, \citet{vaswani2017model}), adaptive clinical trials (\citet{bastani2020online}), personal recommendation systems (\citet{bouneffouf2012contextual}, \citet{bouneffouf2013contextual}, \citet{zhou2017large}) and financial portfolio optimization (\citet{shen2015portfolio}, \citet{huo2017risk}). 

The general $K$-armed MAB problem can be formulated as follows. A learner has $K$ reward distributions with means $\mu_1,\mu_2,\ldots,\mu_K$, respectively. The set of indices $\{1,2,\ldots,K\}$ is called the \emph{action space}. At time $t$, the learner observes the history $\mathcal{F}_{t-1} := \sigma(A_1,X_1,\ldots,A_{t-1},X_{t-1})$---the $\sigma$-field generated by the data collected up to time $t-1$---and chooses an action $A_t$. The function mapping the history to the action space is called a \emph{policy}. If $A_t = a$, then the learner receives a reward $X_t$ from arm $a$. Formally,
\begin{align}
\label{eqn:bandits}
    X_t = \mu_a + \epsilon_{a,t} \qquad \text{if } A_t = a.
\end{align}
The zero-mean noise $\epsilon_{a,t}$ is assumed to be independent of the history $\mathcal{F}_{t-1}$ at every round. To make the presentation succinct, we assume that all arms have the same variance, i.e., $\Exs[\epsilon^2_{a,t}] = \sigma^2$ for all arms $a \in [K]$ and rounds $t \ge 1$. We use $n_{a,t}$ to denote the number of pulls of arm $a$ up to time~$t$, and $T$ to denote the total number of interactions with the environment. The loss function for this problem is called the \emph{regret}, defined as
\begin{align}
    \operatorname{Reg}(T) := \sum_{t=1}^{T} \left( \mu^\star - \mu_{A_t} \right),
\end{align}
where $\mu^\star := \max_{a \in [K]} \mu_a$ is the maximum arm mean.

The core challenge in the MAB problem is the exploration-exploitation tradeoff. As the reward distributions are unknown, non-adequate arm pulls may lead to sub-optimal arm choice, whereas over-pulling of sub-optimal arms may lead to higher regret. To address this issue, a popular class of bandit algorithms called Upper Confidence Bound (UCB) was proposed by Lai and Robbins (\citet{lai1985asymptotically}) and later popularized by \cite{lai1987adaptive,katehakis1995sequential,auer2002finite}. In the UCB algorithm at time $t$, the learner creates high probability upper confidence bounds $U_a(t)$ of the mean $\mu_a$ for each arm $a$, and selects the arm with highest bound. For different choice of $U_a(t)$ one obtains different UCB strategies: some prominent examples include UCB (\citet{lai1987adaptive}, \citet{auer2002finite}), MOSS (\citet{audibert2009minimax}), UCB-V (\citet{audibert2009exploration}) and KL-UCB (\citet{pmlr-v19-garivier11a}).

While bandit algorithms are traditionally motivated by regret minimization, in several studies experimenters and scientists are also interested to conduct statistical inference from the data collected via running a bandit algorithm (\citet{chow2005statistical,berry2012adaptive,zhang2022statistical,trella2023reward,trella2024oralytics,zhang2024replicable,trella2025deployed}). However, due to the adaptive structure of the bandit strategies the data collected is highly non-i.i.d. As a consequence, standard techniques to conduct inference may lead to erroneous conclusions. In particular, the sample means---computed from data collected from adaptive algorithms---may not converge to normal distribution (\citet{zhang2020inference,deshpande2023online,khamaru2021near,lin2023statistical,lin2025semiparametric}). To tackle this issue, a growing line of research (\citet{kalvit2021closer},\citet{khamaru2024inference,fan2024precise,han2024ucb,halder2025stable}) focuses on a notion called \textit{stability} of bandit algorithms which was introduced by \citet{lai1982least}. It is defined as follows:
\begin{definition}
  A K-armed bandit algorithm $\mathcal{A}$ is called stable if for all arm $a \in [K]$, there exist \emph{non-random} scalars $n^{*}_{a,T}(\mathcal{A})$ such that
  \begin{equation}\label{defn-stability}
      \dfrac{n_{a,T}(\mathcal{A})}{n^{*}_{a,T}(\mathcal{A})} \xrightarrow[]{\prob} 1 \ \ \ \text{and} \ \ \ n^{*}_{a,T}(\mathcal{A}) \rightarrow \infty \ \ \text{as} \ \ T \rightarrow \infty.
  \end{equation}
 Here $n_{a,T}(\mathcal{A})$ denotes the number of times arm $a$ has been pulled up to time $T$.
\end{definition}

Stability provides a sufficient condition ensuring that sample means exhibit asymptotic normality as $T \rightarrow \infty$ (see Section \ref{sec-stab-mab} for a detailed discussion). Conversely, as discussed in Appendix \ref{append-instab-non-norm}, the absence of stability may prevent the applicability of a central limit theorem, due to non-normality. Thus, determining whether a given bandit algorithm is stable is a question of both theoretical and practical significance. While the stability of UCB-1 has been established (\citet{fan2022typical,khamaru2024inference,han2024ucb,chen2025characterization}), this algorithm does not attain the minimax lower bound of $\Theta(\sqrt{KT})$ (\cite{han2024ucb}, \citet{lattimore2020bandit}). In this paper we consider a popular class of bandit algorithms which are minimax optimal (\cite{audibert2009minimax}), which we call the~\minimaxfam. Our goal is to  investigate  the extent to which this \minimaxfam of bandit strategies exhibit the stability notion (equation~\eqref{defn-stability}).

\subsection{Related Work}

\noindent When data is collected adaptively, local dependencies in sample points lead to non-i.i.d structure where standard techniques can fail. Such problems are known to fail to hold when applied to data collected from time-series and econometric studies (\cite{dickey1979distribution,white1958limiting,white1959limiting}). In particular, for bandit algorithms \cite{lai1982least} introduced the notion of stability and established that asymptotic normality holds under this condition. Building on this foundation, subsequent research has leveraged stability to construct asymptotically valid confidence intervals (\citet{kalvit2021closer}, \citet{khamaru2024inference}, \citet{fan2022typical}, \citet{fan2024precise}, \citet{han2024ucb}, \citet{halder2025stable}).

 This work aims to bridge the gap in understanding between the class of minimax-optimal bandit algorithms and asymptotic inference through Lai's stability condition. The MOSS algorithm was the first to be shown to achieve minimax regret (\citet{audibert2009minimax}). Since then, several variants of MOSS (\citet{garivier2016explore}, \citet{menard2017minimax}, \citet{lattimore2020bandit}), \citet{lattimore2015optimally}) have been proposed to address its limitations (see Section \ref{sec-MOSS-fam} for details). We note that results concerning the instability of multi-armed bandit algorithms remain relatively scarce. Recently \citet{fan2024precise} have identified that the UCB-V algorithm can be unstable in some cases.  

We note that there exists some alternate approaches to construction of valid confidence intervals (CI). 
One is the non-asymptotic approach which constructs finite-sample CI via the application of concentration inequalities for self-normalized martingales. This line of work is build on the foundational analysis of de la Pena et al (\citet{de2004self}, \citet{de2009self}),  which hold uniformly over time (see  \citet{abbasi2011improved,howard2020time,waudby2024anytime}, \citet{waudby2024anytime}, and references therein). However, the CI obtained via stability are often substantially shorter than those derived from anytime-valid approaches.

As the data collected via bandit algorithms are adaptive in nature, the sample means $\widehat{\mu}_a$ is biased. To tackle this issue, several researchers have studied bandits from the debiasing or doubly-robust perspective (\citet{zhang2014confidence,krishnamurthy2018semiparametric,kim2019doubly,chen2022debiasing,kim2023double}). Furthermore, in the context of Off Policy Evaluation (OPE) the uncertainty quantification problem has been approached via inverse-propensity scores and importance sampling techniques (\citet{hadad2021confidence,zhang2022statistical,nair2023randomization,leiner2025adaptive}).

\subsection{Contributions}
\label{sec:contrib}

\noindent 
In this paper, we study the instability properties of a broad class of optimism principle~\eqref{algo:optimism} based multi-armed bandit algorithms. This class includes the UCB-style minimax-optimal algorithms in Table~\ref{tab:upd-rule}—such as MOSS, Anytime-MOSS, Vanilla-MOSS, ADA-UCB, and OC-UCB—as well as the Kulback-Leibler (KL) divergene based minimax optimal algorithms in Table~\ref{tabtwo:upd-rule}, including KL-MOSS, KL-UCB++, KL-UCB-SWITCH, and Anytime-KL-UCB-switch. These algorithms are introduced formally in Section~\ref{sec-MOSS-fam}.
Section~\ref{sec-stab-mab} provides a detailed discussion of the Lai–Wei stability condition for multi-armed bandits and its implications for asymptotic inference. Our main theoretical result, Theorem~\ref{thm:Instability}, establishes general sufficient conditions under which an optimism-based algorithm fails to satisfy the Lai–Wei stability criterion. As a consequence, Corollary~\ref{cor:instab-Table-1}  shows that all algorithms in Tables~\ref{tab:upd-rule} are unstable when the reward distribution is $1$ sub-Gaussian. Furthermore, Corollary~\ref{cor:instab-Table-2} shows that algorithms listed in Table~\ref{tabtwo:upd-rule} are unstable under one parameter exponential family of distribution. Finally, in Section~\ref{sec:sim}, we complement our theoretical findings with numerical experiments demonstrating that the sample means produced by these algorithms fail to exhibit asymptotic normality.

\section{The MOSS Family of Algorithms}\label{sec-MOSS-fam}
\noindent 
The multi-armed bandit and reinforcement learning based algorithms have gained renewed interest in modern times due to their various applications. One popular class of algorithms is the upper confidence bound family. An important and widely used method in this class is the Upper Confidence Bound (UCB)-$1$ algorithm. At every round $t$, the UCB-$1$ algorithm~\cite{auer2002finite} selects arm $A_{t+1}$ according to
\begin{align}
\label{algo:UCB}
A_{t+1}
= \arg\max_{a \in [K]}
\left\{
\widehat{\mu}_{a}(t)
+ \sqrt{\frac{2\log T}{n_{a,t}}}
\right\},
\end{align}
where $\widehat{\mu}_{a}(t)$ is the empirical mean of the observed rewards for arm $a$ up to time $t$. The regret of the UCB-$1$ algorithm is $O(\sqrt{KT \log T})$, which is sub-optimal by a factor $\sqrt{\log T}$. The Minimax Optimal Strategy in the Stochastic case (MOSS) algorithm, introduced by~\citet{lai1987adaptive} and later refined and popularized by~\citet{audibert2009minimax}, resolves this sub-optimality and attains a minimax regret of $O(\sqrt{KT})$. The MOSS algorithm selects arms using the update rule
\begin{align}
\label{algo:MOSS-eqn}
A_{t+1}
:= \arg\max_{a \in [K]}
\left\{
\widehat{\mu}_a(t)
+ \sqrt{
\frac{\max\{0,\log((T/K)/n_{a,t})\}}
     {n_{a,t}}
}
\right\}.
\end{align}

At a high level,
\begin{itemize}
    \item In the \emph{early stages}---when $n_{a,t}$ is small---the MOSS algorithm behaves similarly to the UCB-$1$ algorithm.
    \item In the \emph{later stages}---when $n_{a,t}$ is large---MOSS reduces the uncertainty for heavily explored arms. This modification leads to improved regret bounds compared to UCB-$1$.
\end{itemize}

We detail the MOSS algorithm in Algorithm~\ref{alg:moss}.

\begin{algorithm}[H]
\caption{MOSS (Minimax Optimal Strategy in the Stochastic case)}
\label{alg:moss}
\DontPrintSemicolon
\SetKwInOut{Input}{Input}
\SetKwInOut{Output}{Output}
\Input{Number of arms $K$; time horizon $T$.}
\Output{Chosen arms $(A_t)_{t=1}^T$.}
\BlankLine 
For arm $a$ at round $t$, let $n_{a,t}$ be pulls so far, $\widehat{\mu}_a(t)$ its mean, and define
\[
U_a(t)=\widehat{\mu}_a(t)+\sqrt{\frac{\max\{0,\log\!\big(\frac{T/K}{n_{a,t}}\big)\}}{n_{a,t}}}.
\]
\textbf{Init:} Pull each arm once.\;
\For{$t=K{+}1$ \KwTo $T$}{
  Select $A_t\in\arg\max_a U_a(t-1)$.\;
  Pull $A_t$, observe $\Rew_t$.\;
  $n_{a,t}\leftarrow n_{a,t-1}+1\left\{A_t = a\right\}$, \quad
  $\widehat{\mu}_{a,t}\leftarrow \frac{n_{a,t-1}}{n_{a,t}}\widehat{\mu}_{a,t-1} + \dfrac{\Rew_{t}}{n_{a,t}} \mathbb{I}_{\{A_t = a\}}$.\;
  }
\end{algorithm}

While MOSS attains the minimax optimal regret, it also exhibits several important limitations. 
First, the algorithm requires prior knowledge of the horizon~$T$, which is unavailable in many practical settings. 
Second, MOSS is not \emph{instance-optimal}~\cite{lattimore2020bandit}; despite its favorable minimax performance, it may behave poorly on specific bandit instances. 
These issues have motivated the development of refined variants that retain (or nearly retain) the minimax optimality of MOSS while improving performance in more nuanced regimes. Table~\ref{cor:instab-Table-1} provides detailed descriptions of these algorithms for $1$-sub-Gaussian reward distributions.

\begin{table}[H]
\centering
\caption{Upper Confidence Bound for MOSS variants}
\label{tab:upd-rule}
\renewcommand{\arraystretch}{2.5}  
\setlength{\tabcolsep}{8pt}
\begin{tabular}{p{4cm} p{12cm}}
\toprule
\textbf{Algorithm} & \textbf{Upper Confidence Bound \(U_a(t)\)} \\
\midrule

\textbf{MOSS} & 
\(\widehat{\mu}_a(t) + \sqrt{\dfrac{\max\{0,\;\log \bigl(\tfrac{T}{K n_{a,t}}\bigr)\}}{n_{a,t}}}\) \\[8pt]

\textbf{Anytime-MOSS} & 
\(\widehat{\mu}_a(t) + \sqrt{\dfrac{\max\{0,\; \log\bigl(\tfrac{t}{K n_{a,t}}\bigr)\}}{n_{a,t}}}\) \\[8pt]

\textbf{Vanilla-MOSS} & 
\(\widehat{\mu}_a(t) + \sqrt{\dfrac{\log \bigl(\tfrac{T}{n_{a,t}}\bigr)}{n_{a,t}}}\) \\[8pt]

\textbf{OC-UCB} & 
\(\widehat{\mu}_a(t) + \sqrt{\dfrac{2(1+\epsilon)}{n_{a,t}}\log \big(\frac{T}{t}\big) } \) \\[8pt]

\textbf{ADA-UCB} & 
\(\widehat{\mu}_a(t) + \sqrt{\dfrac{2}{n_{a,t}}\log \bigg(\dfrac{T}{H_a} \bigg)}\), 
where \(H_a :=\sum^{k}_{j=1} \min(n_{a,t},\sqrt{n_{a,t},n_{j,t}})\) \\[8pt]

\bottomrule
\end{tabular}
\end{table}

\citet{lattimore2015optimally} proposed OC-UCB (Optimally Confident UCB), which achieves the optimal worst-case regret $O(\sqrt{KT})$ while also attaining nearly optimal problem-dependent regret in Gaussian bandits. 
Subsequently, \citet{lattimore2018refining} introduced ADA-UCB, which is both minimax-optimal and asymptotically optimal. 
To address the dependence on the horizon, an \emph{anytime} version of MOSS can be obtained by replacing $T$ with $t$ in the exploration term. 
Moreover, Section~9.2 of \citet{lattimore2020bandit} documents that MOSS may incur worse regret than UCB in certain regimes. 
This can be mitigated by removing the dependence on the number of arms $K$ in the exploration function, yielding a variant which we call \emph{Vanilla-MOSS}.

\newcommand{\KL}{\mathrm{KL}}

All such algorithms arise from replacing the Upper Confidence Bound $U_a(t)$ in step~1 of Algorithm~\ref{alg:moss} with suitable modifications, each grounded in the principle of \emph{optimism in the face of uncertainty}~\cite{lattimore2020bandit}. These algorithms are known to perform well when the reward distribution is $\sigma$ sub-Gaussian. For a sub-class of reward distributions, the UCB-$1$ algorithm can be improved by using the notion of Kullback-Leibler (KL) divergence:  
\begin{definition}[Kullback--Leibler Divergence]
Let $P$ and $Q$ be two probability distributions on the same measurable space, with densities $p$ and $q$ with respect to a common dominating measure. The \emph{Kullback--Leibler divergence} of $Q$ from $P$ is defined as
\begin{align*}
\KL(P , Q)
    = \int p(x) \log\!\left( \frac{p(x)}{q(x)} \right) \, dx.
\end{align*}
When the domain of the random variable is discrete, it reduces to
\begin{align*}
\KL(P , Q)
    = \sum_{x} p(x) \log\!\left( \frac{p(x)}{q(x)} \right).
\end{align*}
\end{definition}

\noindent For one-dimensional canonical exponential distributions the KL-divergence has a simplified form. Suppose $P_\theta$ is absolutely continuous w.r.t. some dominating measure $\rho$ on $\mathbb{R}$ with density $\exp(x\theta - b(\theta))$. Then the KL-divergence becomes a function of the population means
\begin{align*}
   \KL(P_{\theta_1}, P_{\theta_2})= b(\theta_2) - b(\theta_1) - b'(\theta_1)(\theta_2 - \theta_1) 
\end{align*}
By reparameterization $b^{'}(\theta_1) = \mu_1,b^{'}(\theta_2) = \mu_2$, we can rewrite the KL-divergence as follows:
\begin{align*}
    \KL(P_{\theta_1}, P_{\theta_2}) = \mathrm{KL}(\mu_1,\mu_2)
\end{align*}

For Bernoulli distribution, one can obtain tighter confidence bounds in terms of the KL-divergence, than the one obtained for general sub-Gaussian distribution. This notion has been extended to bounded and one-parameter exponential family of distributions (\citet{menard2017minimax}). By combining this idea with the UCB-1 algorithm (equation~\eqref{algo:UCB}), we have the KL-UCB algorithm ~\citep{filippi2010optimistic,maillard2011finite,garivier2011kl}. Let $I$ be the compact set of possible values of population mean $\mu$. Then at every round $t$, select arm $A_{t+1}$ according to
\begin{align}
\label{algo:KL-UCB}
A_{t+1}
= \arg\max_{a \in [K]}
\left\{
\sup \left\{ q \in I :
\KL(\widehat{\mu}_{a,t},q)
\leq \frac{f(t)}{n_{a,t}}
\right\}
\right\},
\end{align}
where $f(t) = \log t + \log \log t$

As the minimax regret of KL-UCB is also sub-optimal, subsequent work combined insights from both MOSS and KL-UCB, yielding refined algorithms such as KL-MOSS, KL-UCB-SWITCH, and KL-UCB++~\citep{garivier2016explore,menard2017minimax,garivier2022kl}. KL-UCB-SWITCH and Anytime KL-UCB-SWITCH achieve optimal distribution-free worst-case regret, while KL-UCB++ is both minimax-optimal and asymptotically optimal for single-parameter exponential families. Table~\ref{tabtwo:upd-rule} summarizes their UCB indices.

\begin{table}[H]
\centering
\caption{Upper Confidence Bound for KL-based MOSS variants}
\label{tabtwo:upd-rule}
\renewcommand{\arraystretch}{2.5}  
\setlength{\tabcolsep}{8pt}
\begin{tabular}{p{4cm} p{12cm}}
\toprule
\textbf{Algorithm} & \textbf{Upper Confidence Bound \(U_a(t)\)} \\
\midrule

\textbf{KL-MOSS} & 
\(\displaystyle  \sup\Big\{ q \in I : \,\KL(\widehat{\mu}_a(t),q) \le \frac{\log^{+}(T / n_{a,t})}{n_{a,t}} \Big\}\) \\[8pt]

\textbf{KL-UCB++} & 
\(\displaystyle \sup\{ q \in I : \,\KL(\widehat{\mu}_a(t),q) \le  \dfrac{h(n_{a,t})}{n_{a,t}}\}\), \\[8pt]
& where \(h(n):= \log^{+} \bigg(\dfrac{T/K}{n}  \left(\log^{+} \dfrac{T/K}{n} \right)^2 + 1  \bigg)\) \\[8pt]

\textbf{KL-UCB-Switch} & 
\(\displaystyle
\begin{cases} 
U^{KM}_a(t), & n_{a,t} \le \lfloor (T/K)^{1/5} \rfloor \\
U^M_a(t), & n_{a,t} > \lfloor (T/K)^{1/5} \rfloor
\end{cases}\), \\
&where $U^{KM}_a(t),U^M_a(t)$ are the UCB of KL-MOSS and MOSS respectively \\[12pt]

\textbf{Anytime--KL-UCB-Switch} &
\(\displaystyle 
\begin{cases}
\displaystyle 
\sup\Bigl\{ q \in I :\,
\KL\!\left(\widehat{\mu}_a(t), q\right)
\le 
\dfrac{\phi\!\left(\tfrac{t}{K n_{a,t}}\right)}{n_{a,t}}
\Bigr\},
& \text{if } n_{a,t} \le f(t,K), \\[10pt]
\displaystyle 
\widehat{\mu}_a(t) 
+ \sqrt{\dfrac{1}{2 n_{a,t}}\, \phi\!\left(\tfrac{t}{K n_{a,t}}\right)},
& \text{if } n_{a,t} > f(t,K),
\end{cases}
\) 
\\[10pt]
& where \(\phi(x) := \log^{+}(x)\bigl(1 + (\log^{+} x)^2\bigr)\) and $f(t,K) = [(t/K)^{1/5}]$. \\[8pt]
\bottomrule
\end{tabular}
\end{table}

\section{Inference in  Multi-Armed Bandits via Stability}\label{sec-stab-mab}
The ability to conduct valid inference is of particular importance in settings such as clinical trials~\citep{marschner2024confidence,trella2023reward,trella2024oralytics,zhang2024replicable,trella2025deployed} and A/B testing ( \citep{dubarry2015confidence}). A central difficulty is that data generated by multi-armed bandit algorithms are inherently non-i.i.d., rendering classical estimators biased and inconsistent in general~\citep{zhang2020inference,deshpande2018accurate}. Consequently, the direct application of standard asymptotic theory for hypothesis testing and uncertainty quantification can lead to severely distorted inference.

To address these challenges, recent work has focused on the notion of \emph{stability} in bandit algorithms, a concept originally introduced in \citet{lai1982least}. Informally, a multi-armed bandit algorithm~$\mathcal{A}$ is stable if the number of times each arm is pulled concentrates around certain algorithm-dependent deterministic quantities. Formally, we recall from the definition(~\ref{defn-stability}) that
a $K$-armed bandit algorithm $\mathcal{A}$ is called stable if, for all arms $a \in [K]$, there exist \emph{non-random} scalars $n^{*}_{a,T}(\mathcal{A})$ such that
  \begin{equation}
      \frac{n_{a,T}(\mathcal{A})}{n^{*}_{a,T}(\mathcal{A})} \xrightarrow[]{\mathbb{P}} 1,
      \qquad 
      n^{*}_{a,T}(\mathcal{A}) \rightarrow \infty 
      \quad \text{as } T \rightarrow \infty.
  \end{equation}
  Here $n_{a,T}(\mathcal{A})$ denotes the number of times arm $a$ has been pulled up to time $T$.

A key implication of stability is that when data are collected via a stable bandit algorithm~$\mathcal{A}$, classical asymptotic inference becomes valid despite the adaptive data collection. In particular, one obtains an i.i.d.-like asymptotic normality result:
\begin{lemma}\label{lemma-stab-norm}
    Suppose we have a stable $K$-armed MAB algorithm with sub-Gaussian reward distributions of variance~$1$ for all arms. Then the following distributional convergence holds:
   \begin{equation}\label{eqn-stab-norm}
       \begin{bmatrix}
           \sqrt{n_{1,T}(\mathcal{A})}\,(\widehat{\mu}_{1,T} - \mu_1) \\
           \vdots \\
           \sqrt{n_{K,T}(\mathcal{A})}\,(\widehat{\mu}_{K,T} - \mu_K)
       \end{bmatrix}
       \xrightarrow{d} N(0, I_K).
   \end{equation}
\end{lemma}
This result follows from \cite[Theorem~3]{lai1982least}, combined with the martingale central limit theorem~\citep{hall2014martingale,dvoretzky1972asymptotic}.

A growing line of recent work has examined the stability of widely used bandit algorithms. \citet{kalvit2021closer} establish stability of the UCB-$1$ algorithm~\citep{auer2002finite} in the two-armed case under certain conditions on the arm means, with extensions to the unequal-variance setting~\citet{fan2024precise,chen2025characterization}. \citet{fan2022typical} prove stability of UCB-$1$ and Thompson Sampling under the assumption of a unique optimal arm. The most general results to date are due to \citet{khamaru2024inference} and \citet{han2024ucb}, who establish stability for a broad class of UCB-type algorithms with potentially growing numbers of arms~$K$, arbitrary arm means, and provide rates of convergence for the central limit theorem in Lemma~\ref{lemma-stab-norm}.

\pgfplotsset{compat=1.18} 

\begin{figure}[H]
    \centering

    \begin{subfigure}[t]{0.42\textwidth}
        \includegraphics[width=\textwidth,trim={1cm 0.5cm 1cm 0.5cm}]{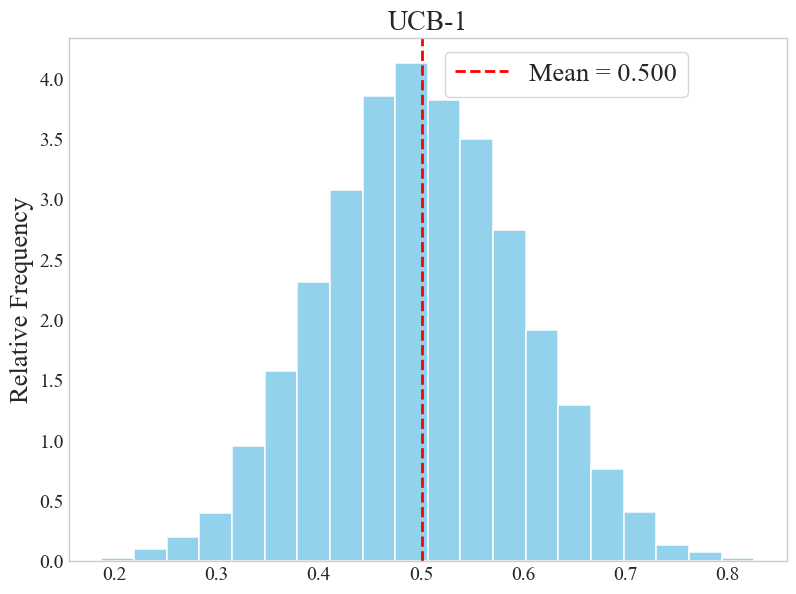} 
        \caption*{$\frac{n_{1,T}}{T}$}%
        \label{fig:ucbs}
    \end{subfigure}
    \hfill
    \begin{subfigure}[t]{0.42\textwidth}
        \includegraphics[width=\textwidth,trim={1cm 0.5cm 1cm 0.5cm}]{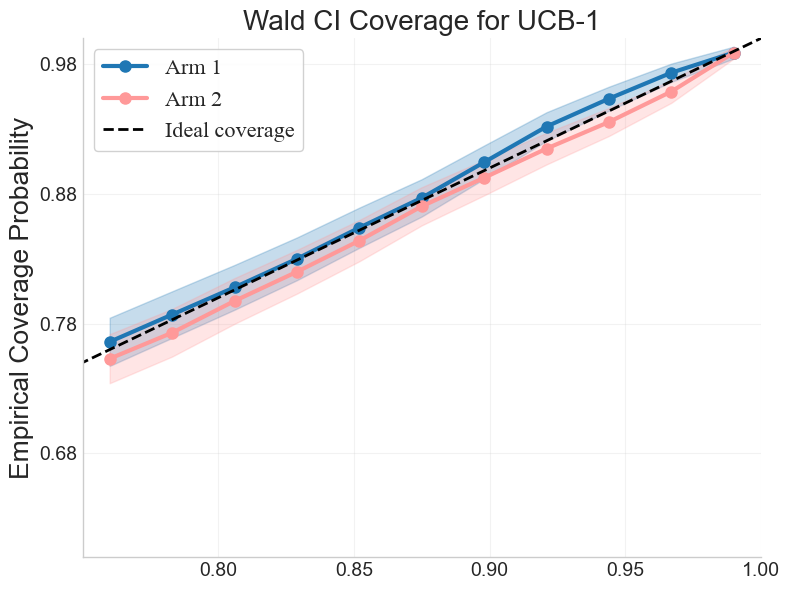} 
         \caption*{Nominal Coverage}
    \end{subfigure}
    \\[4pt]
    \caption{\textbf{Left}: Empirical results for UCB-$1$  based on $5000$ independent sample runs with $T = 10000$ pulls. \textbf{Right}: Coverage of Wald confidence interval of MOSS with $T = 10000$.}
\end{figure}

\subsubsection*{Is Stability Necessary for Asymptotic Normality?}
\noindent Lemma~\ref{lemma-stab-norm} shows that stability is a sufficient condition to achieve asymptotic normality. This naturally leads to the question of whether stability is, in some sense, also necessary for a central limit theorem to hold. To understand this, it is useful to peek at the proof of Lemma~\ref{lemma-stab-norm}. Consider a $K$-armed bandit problem from~\eqref{eqn:bandits}. Given a bandit algorithm~\algo, and for an arm $a \in [K]$, the standardized estimation error admits the decomposition
\begin{align}
    \sqrt{n_{a,T}(\mathcal{A})}\,\bigl(\widehat{\mu}_{a,T}-\mu_a\bigr)
    &=
    \frac{1}{\sqrt{n_{a,T}(\mathcal{A})}}
    \sum_{t=1}^{T} \xi_{a,t}\,\mathbb{I}\{A_t = a\} \nonumber\\[10pt]
    &=
    \underbrace{\sqrt{\frac{n^{*}_{a,T}(\mathcal{A})}{n_{a,T}(\mathcal{A})}}}_{\approx 1}
    \cdot
    \underbrace{\frac{1}{\sqrt{n^{*}_{a,T}(\mathcal{A})}}
    \sum_{t=1}^{T} \xi_{a,t}\,\mathbb{I}\{A_t = a\}}_{\text{(normalized sum of an MDS)}}.
    \label{eq:stab-mds-decomp}
\end{align}

\noindent The first factor in~\eqref{eq:stab-mds-decomp} converges to $1$ in probability by stability. The second factor is a normalized sum of martingale differences adapted to $(\mathcal{F}_t)$. Its conditional variance is given by
\begin{align}\label{eqn-condvar-mds}
    \sum_{t=1}^{T}
    \operatorname{Var}\!\left(
        \frac{1}{\sqrt{n^{*}_{a,T}(\mathcal{A})}}\,
        \xi_{a,t}\,\mathbb{I}\{A_t = a\}
        \,\middle|\,\mathcal{F}_{t-1}
    \right)
    =
    \frac{n_{a,T}(\mathcal{A})}{n^{*}_{a,T}(\mathcal{A})},
\end{align}
which again converges in probability to $1$ under stability. Thus, in the multi-armed bandit setting, stability is precisely the condition that the sum of conditional variances of the underlying martingale difference sequence stabilizes, allowing a martingale central limit theorem to yield asymptotic normality.

Conversely, violations of such variance stabilization are well known to lead to non-Gaussian limits in classical martingale settings, including Gaussian mixture limits (\citet{aldous1978mixing}; \citet{hall2014martingale}; \citet{lai1982least}) and limits arising from Brownian motion indexed by a stopping time (\cite{hall2014martingale}). In the context of bandit algorithms, \cite{zhang2020inference} show that several standard batched policies exhibit non-normal limiting behavior. As we discuss in Appendix~\ref{append-instab-non-norm}, these algorithms fail the stability condition.

\section{Instability of MOSS and it's Variants}

\noindent While \ucb  is provably stable~\cite{khamaru2024inference,han2024ucb}, its regret is not optimal~\cite{lattimore2020bandit,han2024ucb}.  This motivates the search for bandit algorithms which are both stable and attains the minimax-optimal regret of $\sqrt{KT}$. A class of bandit algorithms that attain the minimax optimal regret is the MOSS-family (see Table~\ref{tab:upd-rule}). The main result of this paper is to prove that MOSS family of algorithms, while attaining minimax-optimal regret, are not stable for a $K$-armed bandit problem. Furthermore, in Section~\ref{sec:sim},  we also demonstrate through extensive simulations that the normality-based confidence intervals do not provide correct coverage for the MOSS-family of algorithms. 

Our main result applies to a wide class of algorithms 
that select an arm according to the following~\emph{optimism-based} rule
\begin{align}
\label{algo:optimism}
A_t := \arg\max \limits_{a \in [K]} U_{a}(t) \qquad \texttt{(Optimism-rule)}
\end{align}
where $U_{a}(t)$ is an appropriate upper confidence bound of arm mean $\mu_a$ at time $t$. We assume that our optimism-based rule~\eqref{algo:optimism} satisfies the following conditions:

\paragraph{Assumption A.}
\begin{enumerate}[label=(A\arabic*), ref=~(A\arabic*)]
\item \label{assump:UCB-func} For any arm $a\in[K]$ at any time $t\in [T]$,  the UCB $U_a(t)$ is a function of $\{ \widehat{\mu}_a(t): \ a \in [K]\}$ and $\{n_{a,t}: a\in [K] \}$.
\item \label{assump:H-G} Fix a time $t \in [T]$ and arm $a \in [K]$. If arm $a$ is not pulled at time $t$, then $U_a(t) \leq U_a(t-1)$. 
\item \label{assump:ucb-bound}
We assume that on the event defined below:
\begin{subequations}
\begin{align}
\label{eqn:tail-event}
\taileve:= \left\lbrace
n_{a,\frac{T}{2K}} \in \left[ \frac{T}{4K^2} , \; \frac{3T}{4K^2} \right] \;\; \text{for all} \;\; a \in [K]
\right\rbrace
\end{align}
the following sandwich relation holds for $t\geq T/2K$
\begin{equation}
    \label{eqn-UCB-bound}
        \widehat{\mu}_{a,t} +  \frac{\beta_1}{\sqrt{n_{a,t}}}
        \leq \  U_a(t) \ \leq \ \widehat{\mu}_{a,t} +   \frac{\beta_2}{\sqrt{n_{a,t}}}.
\end{equation} 
\end{subequations} 
Here $\beta_1, \beta_2$ are suitable real (algorithm dependent) constants. 
\end{enumerate}

Assumption~\ref{assump:UCB-func} states that once we know the sample means $\widehat{\mu}_{a,t}$ and the number of arm pulls $n_{a,t}$ for each arm $a$ at time $t$, then we exactly know which arm to pull at next step $t+1$.  We note that if for some algorithm $\Algo$ this assumption is not satisfied, our proof may still go through provided we have  more explicit information on the UCB of $\Algo$. Assumption~\ref{assump:H-G} posits a mild assumption on the behavior of the upper confidence bound $U_a(t)$ of arms that are not pulled. This assumption is satisfied for all Algorithms in Table~\ref{tab:upd-rule} except Anytime MOSS. Our instability proofs also work for Anytime-MOSS with a slight modification of our argument. The requirement in Assumption~\ref{assump:ucb-bound} is more subtle. This assumption requires that for large value of $t$, the upper confidence bound (UCB) $U_{a}(t)$ for all arms are sandwiched between two UCBs with \emph{constant bonus} factors. We demonstrate that all algorithms in Table~\ref{tab:upd-rule} satisfy this condition. A closer look at the proof of  Theorem~\ref{thm:Instability} reveals that this constant-bonus sandwich behavior of the UCB's in the later stage of the algorithm leads to their instability; we return to this point in the comments following Theorem~\ref{thm:Instability}. 

\vspace{0.8pt}

\begin{tcolorbox}
\begin{subequations}
\begin{theorem}
\label{thm:Instability}
    Consider a $K$ armed bandit problem where all arms have identical $1$-sub-Gaussian reward distribution. Then any optimism based rule~\eqref{algo:optimism} satisfying Assumptions~\ref{assump:UCB-func},~\ref{assump:H-G} and~\ref{assump:ucb-bound} is unstable. 
\end{theorem}
\end{subequations}
\end{tcolorbox}

\vspace{1.2pt}

\noindent The proof of this theorem is provided in Section~\ref{sec:Instability-proof}. Let us now discuss how Assumption~\ref{assump:ucb-bound} breaks the stability of \emph{optimism-based} rule~\ref{algo:optimism}. Observe that under Assumption~\ref{assump:ucb-bound} for large values of $t$ the upper confidence bound is safely approximated (sandwiched) by an optimism rule with the following UCB: 
\begin{align}
\label{eqn:constant-ucb}
    \widetilde{U}_{a,\beta}(t) =  \widehat{\mu}_{a,t} +  \frac{\beta}{\sqrt{n_{a,t}}}
\end{align}
where $\beta$ is an appropriate constant that is independent on $n_{a,t}$ and $T$. 
It is instructive to compare the rule~\ref{eqn:constant-ucb} with the UCB-$1$ algorithm analyzed in~\cite{khamaru2024inference,han2024ucb}. The authors show that any UCB style algorithm with UCB's of the form
\begin{align}
\label{algo:UCB1-rule}
    \mathrm{UCB}_a(t):= \widehat{\mu}_{a,t} +  \frac{q_T}{\sqrt{n_{a,t}}}
\end{align}
is stable if $q_T \gg \sqrt{2\log \log T}$. This class of algorithms include for instance the popular UCB-$1$ algorithm by~\cite{auer2002finite}. The term $\sqrt{2\log \log T}$ originates from the law of iterated logarithm (LIL) which captures the fluctuation of the sample mean from the population mean over $T$ rounds. In the proof of Theorem~\ref{thm:Instability} we show that if the bonus term $q_T$ in rule~\ref{algo:UCB1-rule} is replaced by a constant (independent of $T$), then the resulting optimism based rule becomes unstable.  

All algorithms considered in Tables~\ref{tab:upd-rule} and~\ref{tabtwo:upd-rule} satisfy Assumption~\ref{assump:UCB-func}. Furthermore, a simple inspection of algorithms in Table~\ref{tab:upd-rule} satisfy Assumptions~\ref{assump:H-G} and~\ref{assump:ucb-bound} except Anytime MOSS. In Appendix~\ref{append-aux-lemma} we prove that a simple modification of the proof argument of Theorem~\ref{thm:Instability} also shows the instability of Anytime Moss. Overall we conclude.

\begin{corollary}
    \label{cor:instab-Table-1}
     Consider a $K$ armed bandit problem where all arms have identical $1$-sub-Gaussian reward distribution. Then algorithms stated in Table~\ref{tab:upd-rule}, while achieving minimax optimal regret, are unstable. 
\end{corollary}

Note that the KL-divergence based algorithms in Table~\ref{tabtwo:upd-rule} (except Anytime KL-UCB-SWITCH) satisfy Assumption~\ref{assump:H-G}. However, Assumption~\ref{assump:ucb-bound} is not necessarily satisfied for any sub-Gaussian reward distribution. Here we outline a sub-class of reward distributions for which Assumption~\ref{assump:ucb-bound} holds. Suppose the reward distribution is such that the following inequality is satisfied for some constant $c>0$ (which is distribution dependent):
\begin{align}\label{eqn-pinsker}
    \KL(\widehat{\mu}_a(t), q) \;\geq\; c\,\bigl(\widehat{\mu}_a(t) - q\bigr)^2.
\end{align}

\citet{garivier2019explore} show that any distribution supported on $[0,1]$ satisfies~\eqref{eqn-pinsker}. In particular, Bernoulli and uniform distributions fall into this class. Moreover, within canonical exponential families, inequality~\eqref{eqn-pinsker} also holds whenever the mean parameter space $I$ is bounded and the variance is uniformly bounded. This contains a large class of distributions including Bernoulli and Gaussian random variables (with $I$ bounded). Now, for any such reward distribution, the KL-based algorithms—KL-MOSS, KL-UCB++, and KL-UCB-SWITCH—listed in Table~\ref{tabtwo:upd-rule} obey the sandwich bound:
\begin{align}
   \widehat{\mu}_a(t) - \sqrt{\frac{f(T/n_{a,t})}{n_{a,t}}}
   \;\leq\; U_a(t) \;\leq\;
   \widehat{\mu}_a(t) + \sqrt{\frac{f(T/n_{a,t})}{n_{a,t}}}
\end{align}
for some appropriate choices of $f$. Finally, note that the UCB indices of Anytime KL-UCB-SWITCH are sandwiched between the UCB of two \emph{unstable} algorithms which are Anytime in nature. For a rigorous argument see Appendix~\ref{append-aux-lemma}. The discussion above leads us to our second corollary.

\begin{corollary}
\label{cor:instab-Table-2}
     Consider a $K$-armed bandit problem where all arms have identical $1$-sub-Gaussian reward distributions satisfying equation~\eqref{eqn-pinsker}. Then the algorithms stated in Table~\ref{tabtwo:upd-rule}, while achieving minimax optimal regret, are unstable.
\end{corollary}

\subsection{Illustrative simulations}\label{sec:sim}
\noindent In this section, we present simulation results that illustrate both the instability of the algorithms in Tables~\ref{cor:instab-Table-1} and~\ref{cor:instab-Table-2} and the failure of CLT-based confidence intervals (Wald-type intervals) for estimating the arm means. All experiments consider a two-armed bandit with identical reward distributions $\mathcal{N}(0,1)$. For each algorithm, we report two plots. The left panel displays the empirical \emph{relative frequencies} of arm pulls, i.e.,~$n_{1,T}/T$. The right panel reports the empirical coverage of the standard Wald confidence interval for the mean reward. In every case, the Wald interval exhibits substantial undercoverage, indicating significant deviations from normality. These findings provide empirical support for our theoretical results establishing instability of these algorithms, as well as direct evidence that the central limit theorem does not hold for the sample arm-means.

\begin{figure}[H]\label{fig-ucb}
    \centering

    \begin{subfigure}[t]{0.42\textwidth}
        \includegraphics[width=\textwidth,trim={1cm 0.5cm 1cm 0.5cm}]{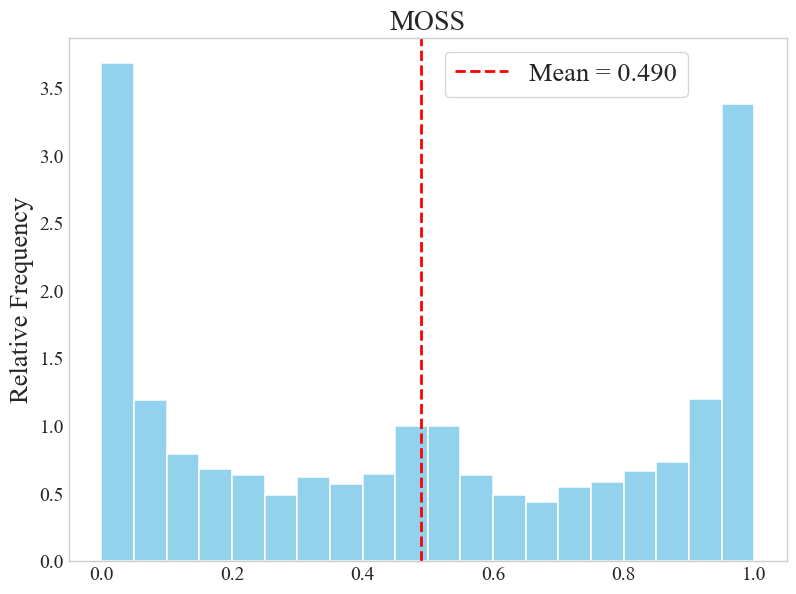}
        \caption*{$\frac{n_{1,T}}{T}$}
        \label{fig:moss}
    \end{subfigure}
    \hfill
    \begin{subfigure}[t]{0.42\textwidth}
        \includegraphics[width=\textwidth,trim={1cm 0.5cm 1cm 0.5cm}]{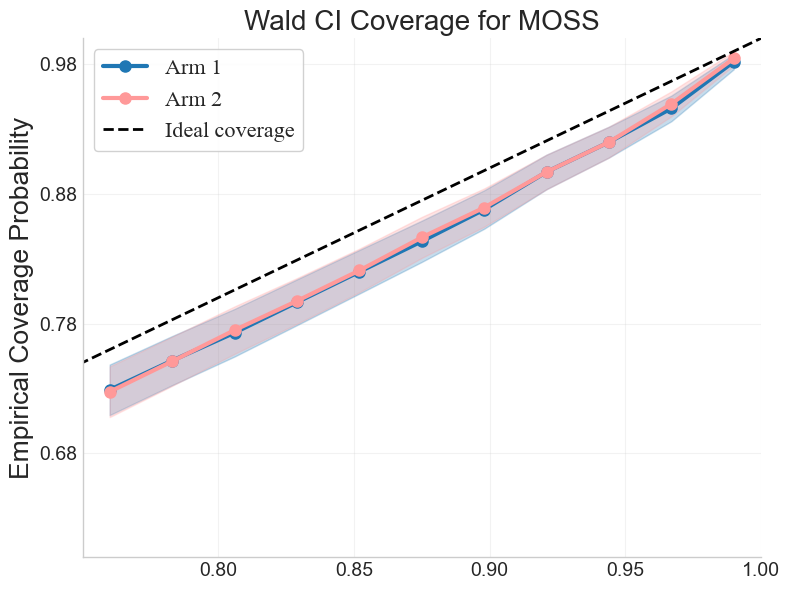}
        \caption*{Nominal Coverage}
    \end{subfigure}
    \\[4pt]
    \caption{\textbf{Left}: Empirical results for MOSS  based on $5000$ independent sample runs with $T = 10000$ pulls. \textbf{Right}: Coverage of Wald confidence interval of MOSS with $T = 10000$.}
     \label{fig:moss-cov}
\end{figure}

\begin{figure}[H]
    \centering

    \begin{subfigure}[t]{0.42\textwidth}
        \includegraphics[width=\textwidth,trim={1cm 0.5cm 1cm 0.5cm}]{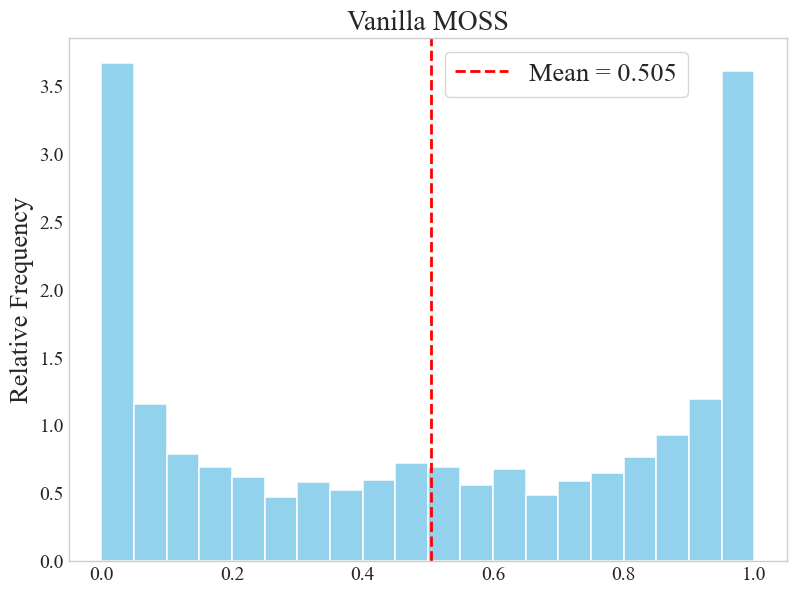} 
        \caption*{$\frac{n_{1,T}}{T}$}%
        \label{fig:vmoss}
    \end{subfigure}
    \hfill
    \begin{subfigure}[t]{0.42\textwidth}
        \includegraphics[width=\textwidth,trim={1cm 0.5cm 1cm 0.5cm}]{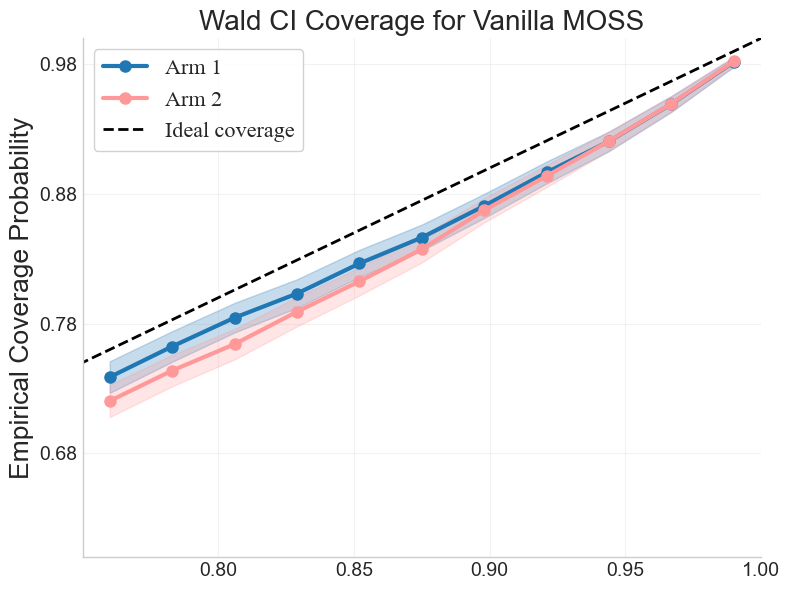} 
         \caption*{Nominal Coverage}
    \end{subfigure}

    \caption{\textbf{Left}: Empirical results for Vanilla MOSS  based on $5000$ independent sample runs with $T = 10000$ pulls. \textbf{Right}: Coverage of Wald confidence interval of Vanilla MOSS with $T = 10000$.}
     \label{fig:vmoss-cov}
\end{figure}

\begin{figure}[H]
    \centering

    \begin{subfigure}[t]{0.42\textwidth}
        \includegraphics[width=\textwidth,trim={1cm 0.5cm 1cm 0.5cm}]{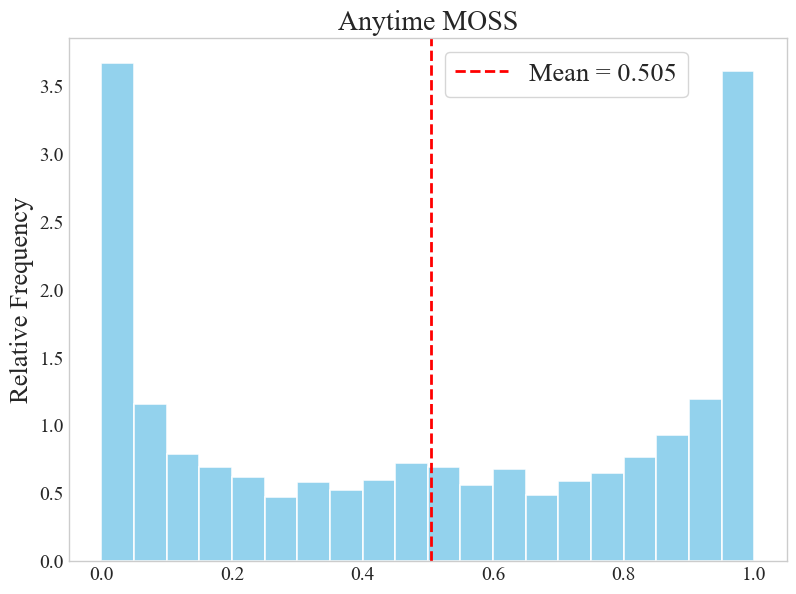} 
        \caption*{$\frac{n_{1,T}}{T}$}%
        \label{fig:any_moss}
    \end{subfigure}
    \hfill
    \begin{subfigure}[t]{0.42\textwidth}
        \includegraphics[width=\textwidth,trim={1cm 0.5cm 1cm 0.5cm}]{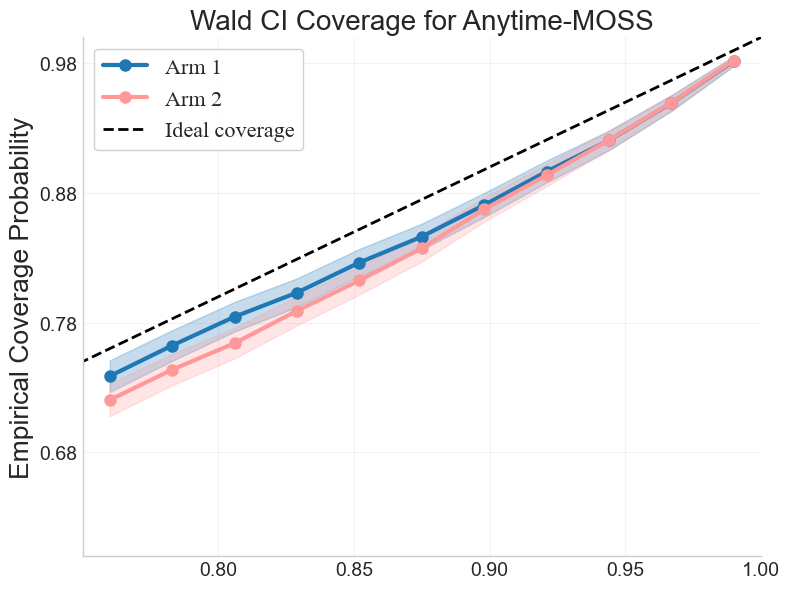} 
         \caption*{Nominal Coverage}
        \label{fig:any_moss-cov}
    \end{subfigure}

    \caption{\textbf{Left}: Empirical results for Anytime MOSS  based on $5000$ independent sample runs with $T = 10000$ pulls. \textbf{Right}: Coverage of Wald confidence interval of Anytime MOSS with $T = 10000$.}
    \label{fig:1x2}
\end{figure}

\begin{figure}[H]
    \centering

    \begin{subfigure}[t]{0.42\textwidth}
        \includegraphics[width=\textwidth,trim={1cm 0.5cm 1cm 0.5cm}]{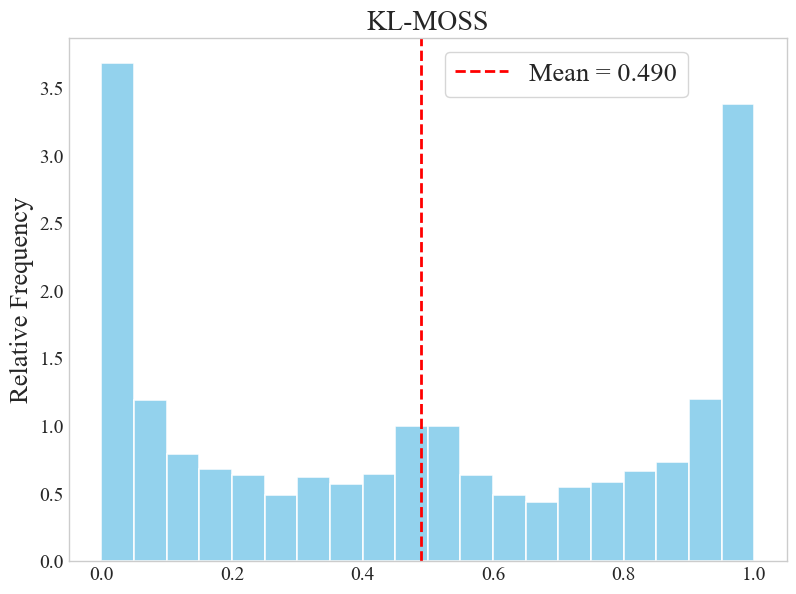} 
        \caption*{$\frac{n_{1,T}}{T}$}%
        \label{fig:kl-moss}
    \end{subfigure}
    \hfill
    \begin{subfigure}[t]{0.42\textwidth}
        \includegraphics[width=\textwidth,trim={1cm 0.5cm 1cm 0.5cm}]{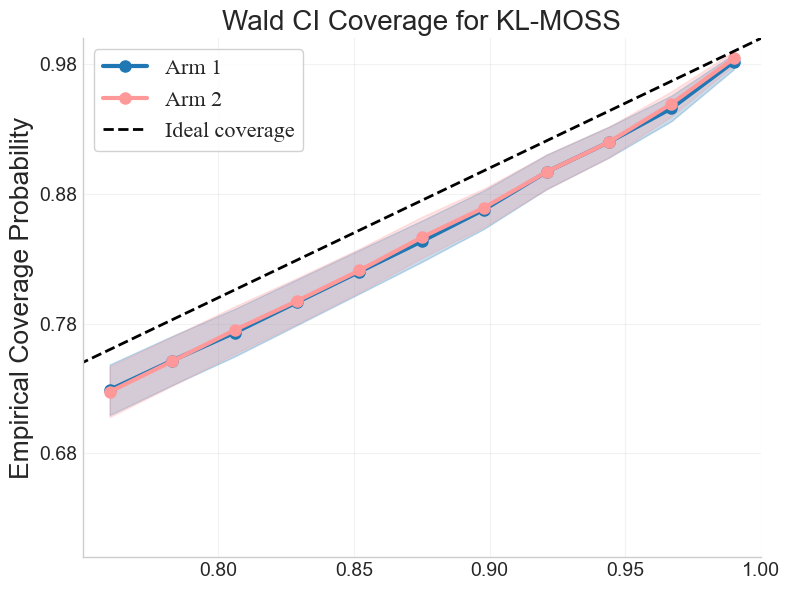} 
     \caption*{Nominal Coverage}
    \end{subfigure}
    \caption{\textbf{Left}: Empirical results for KL-MOSS  based on $5000$ independent sample runs with $T = 10000$ pulls. \textbf{Right}: Coverage of Wald confidence interval of KL-MOSS with $T = 10000$.}
     \label{fig:kl-moss-cov}
\end{figure}

\begin{figure}[H]
    \centering

    \begin{subfigure}[t]{0.42\textwidth}
        \includegraphics[width=\textwidth,trim={1cm 0.5cm 1cm 0.5cm}]{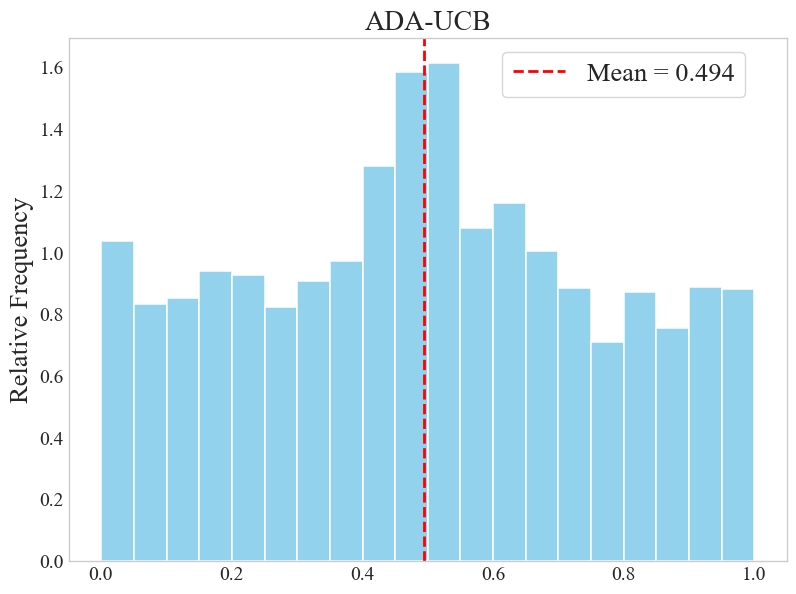} 
        \caption*{$\frac{n_{1,T}}{T}$}%
        \label{fig:ada_ucb}
    \end{subfigure}
    \hfill
    \begin{subfigure}[t]{0.42\textwidth}
        \includegraphics[width=\textwidth,trim={1cm 0.5cm 1cm 0.5cm}]{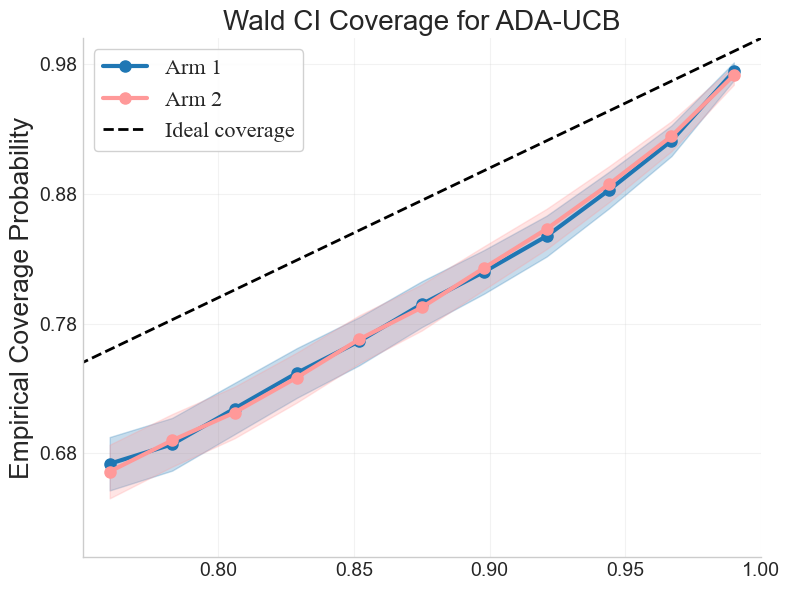}
      \caption*{Nominal Coverage}   
    \end{subfigure}

    \caption{\textbf{Left}: Empirical results for ADA-UCB  based on $5000$ independent sample runs with $T = 10000$ pulls. \textbf{Right}: Coverage of Wald confidence interval of ADA-UCB with $T = 10000$.}
    \label{fig:ada_ucb-cov}
\end{figure}

\begin{figure}[H]
    \centering

    \begin{subfigure}[t]{0.42\textwidth}
        \includegraphics[width=\textwidth,trim={1cm 0.5cm 1cm 0.5cm}]{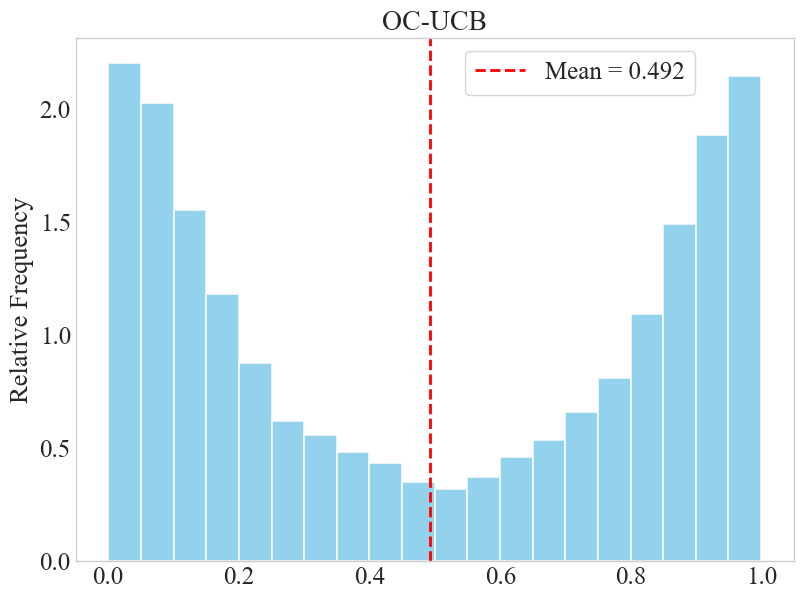} 
        \caption*{$\frac{n_{1,T}}{T}$}%
        \label{fig:oc_ucb}
    \end{subfigure}
    \hfill
    \begin{subfigure}[t]{0.42\textwidth}
        \includegraphics[width=\textwidth,trim={1cm 0.5cm 1cm 0.5cm}]{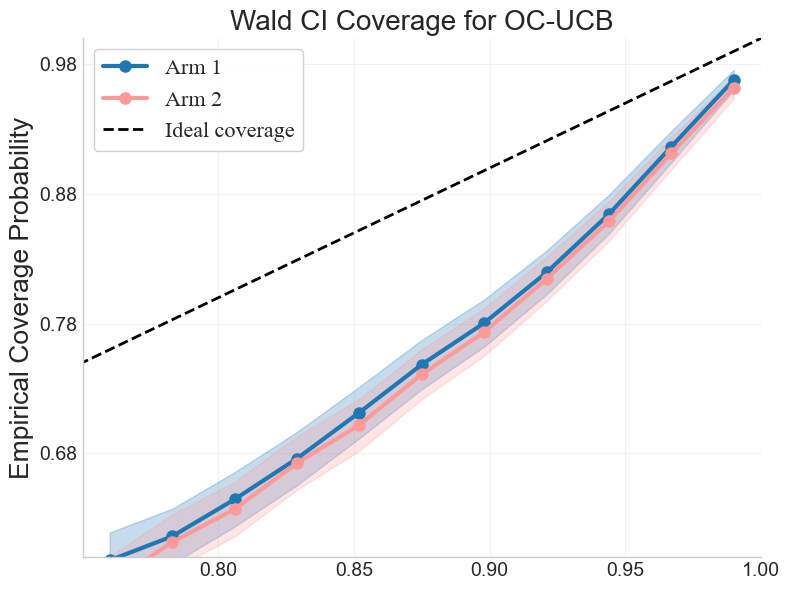} 
         \caption*{Nominal Coverage}
        \label{fig:oc_ucb-cov}
    \end{subfigure}

    \caption{\textbf{Left}: Empirical results for OC-UCB  based on $5000$ independent sample runs with $T = 10000$ pulls. \textbf{Right}: Coverage of Wald confidence interval of OC-UCB with $T = 10000$. We have chosen $\epsilon = 0.1$.}
\end{figure}

\begin{figure}[H]
    \centering

    \begin{subfigure}[t]{0.42\textwidth}
        \includegraphics[width=\textwidth,trim={1cm 0.5cm 1cm 0.5cm}]{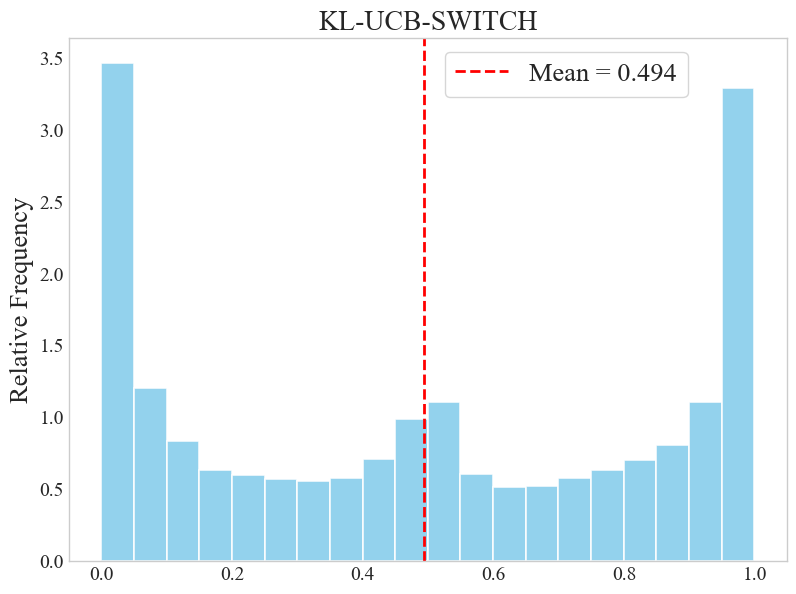} 
        \caption*{$\frac{n_{1,T}}{T}$}%
        \label{fig:kl_ucb_switch}
    \end{subfigure}
    \hfill
    \begin{subfigure}[t]{0.42\textwidth}
        \includegraphics[width=\textwidth,trim={1cm 0.5cm 1cm 0.5cm}]{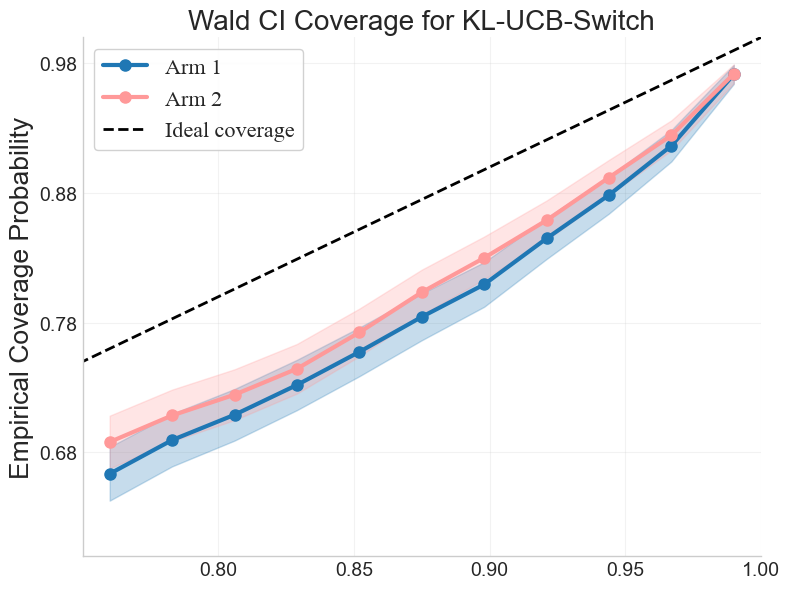} 
        \caption*{Nominal Coverage}
        \label{fig:kl_ucb_switch-cov}
    \end{subfigure}

    \caption{\textbf{Left}: Empirical results for KL-UCB-Switch  based on $5000$ independent sample runs with $T = 10000$ pulls. \textbf{Right}: Coverage of Wald confidence interval of KL-UCB-Switch with $T = 10000$.}
\end{figure}

\begin{figure}[H]
    \centering

    \begin{subfigure}[t]{0.42\textwidth}
        \includegraphics[width=\textwidth,trim={1cm 0.5cm 1cm 0.5cm}]{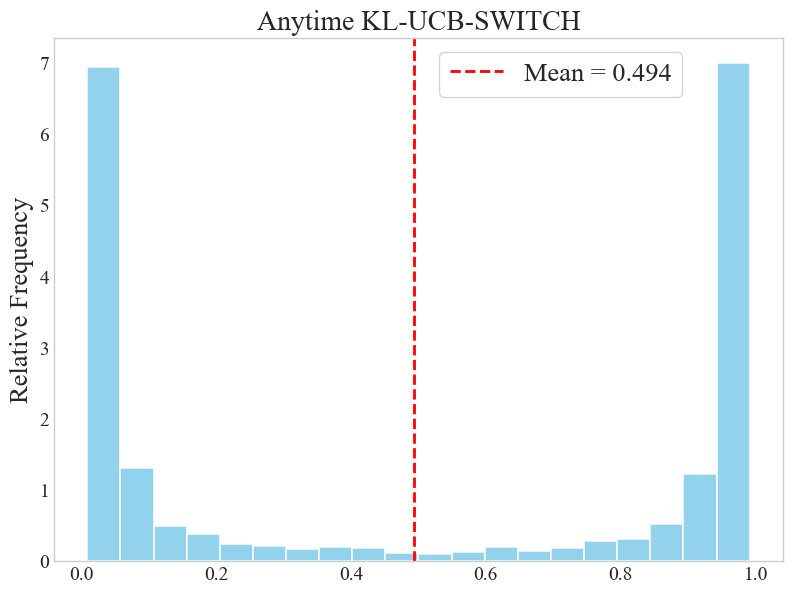} 
        \caption*{$\frac{n_{1,T}}{T}$}%
        \label{fig:any_kl_ucb_switch}
    \end{subfigure}
    \hfill
    \begin{subfigure}[t]{0.42\textwidth}
        \includegraphics[width=\textwidth,trim={1cm 0.5cm 1cm 0.5cm}]{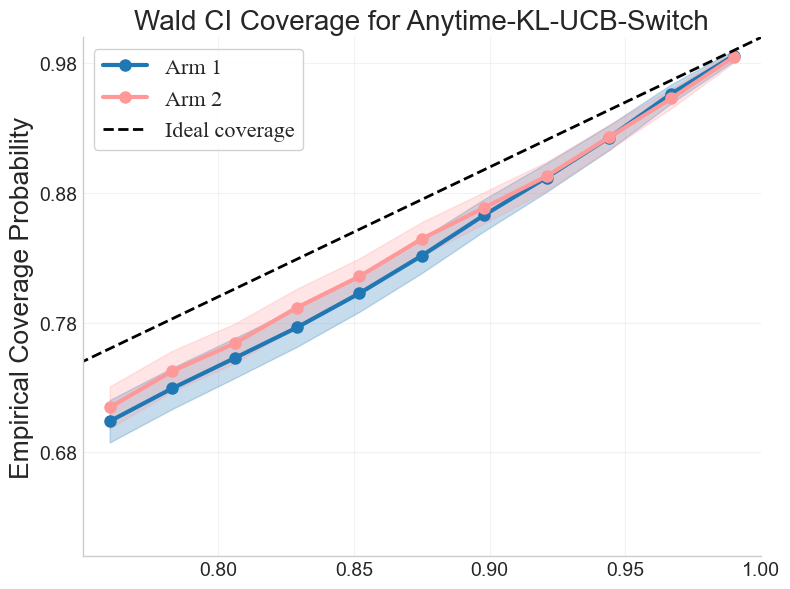} 
        \caption*{Nominal Coverage}
        \label{fig:any_kl_ucb_switch-cov}
    \end{subfigure}

    \caption{\textbf{Left}: Empirical results for Anytime-KL-UCB-Switch  based on $5000$ independent sample runs with $T = 10000$ pulls. \textbf{Right}: Coverage of Wald confidence interval of KL-UCB-Switch with $T = 10000$.}
\end{figure}

\begin{figure}[H]
    \centering

    \begin{subfigure}[t]{0.42\textwidth}
        \includegraphics[width=\textwidth,trim={1cm 0.5cm 1cm 0.5cm}]{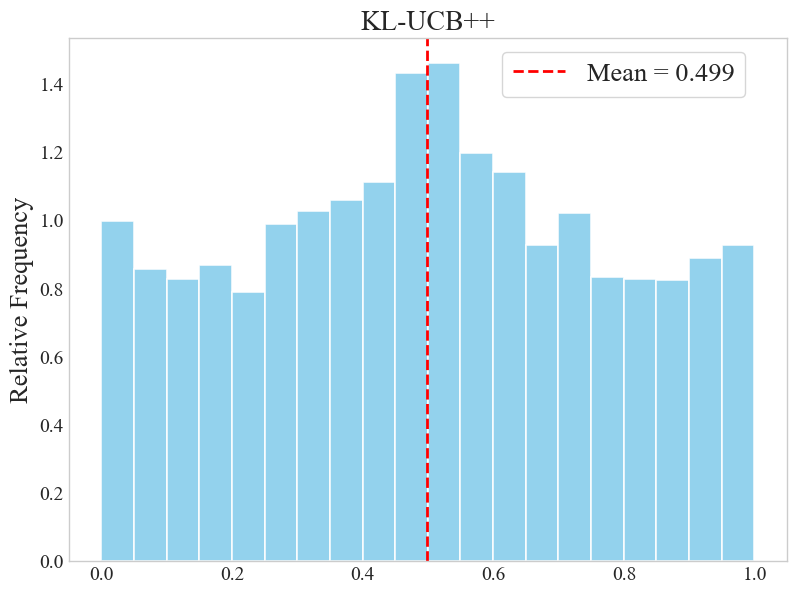} 
        \caption*{$\frac{n_{1,T}}{T}$}%
        \label{fig:kl_ucb_pplus}
    \end{subfigure}
    \hfill
    \begin{subfigure}[t]{0.42\textwidth}
        \includegraphics[width=\textwidth,trim={1cm 0.5cm 1cm 0.5cm}]{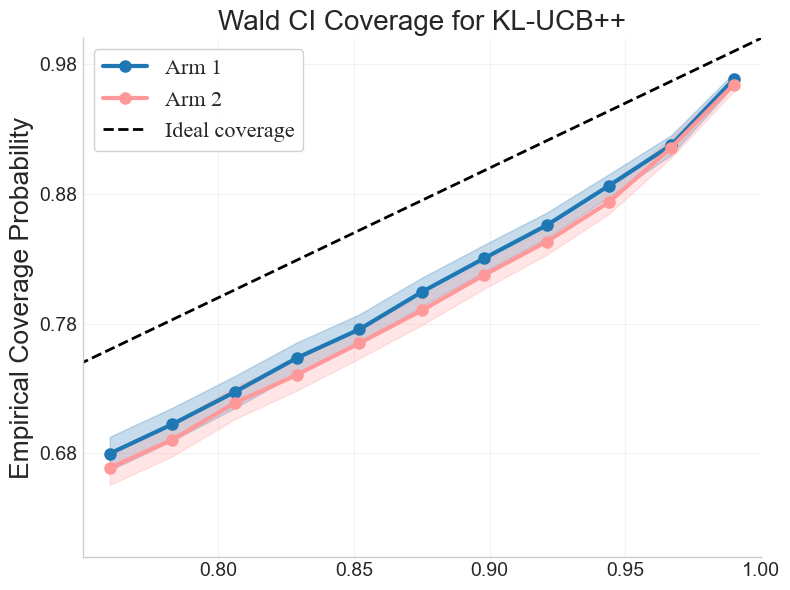} 
        \caption*{Nominal Coverage}
        \label{fig:kl_ucb_pplus-cov}
    \end{subfigure}

    \caption{\textbf{Left}: Empirical results for KL-UCB++  based on $5000$ independent sample runs with $T = 10000$ pulls. \textbf{Right}: Coverage of Wald confidence interval of KL-UCB++ with $T = 10000$.}
\end{figure}

\section{Proof of Theorem~\ref{thm:Instability} } \label{sec:Instability-proof}

\noindent We prove this theorem by contradiction. We use \algo to denote a generic optimism rule~\eqref{algo:optimism}. We prove Theorem~\ref{thm:Instability} in two steps: 
\begin{itemize}
    \item[(a)] In Lemma~\ref{lemma-stab-seq}, we show that if an algorithm \algo is stable under the set up of Theorem~\ref{thm:Instability}, then \algo must be stable with $n^\star_{a, T} = T/K$ for all arm $a$. 
    \item[(b)] Next we argue that if algorithm \algo is stable in the first $T/2K$ rounds, then under Assumptions~\ref{assump:UCB-func},~\ref{assump:H-G} and~\ref{assump:ucb-bound} the algorithm \algo must violate the stability condition~\eqref{defn-stability} in later rounds $(t \geq T/2K)$ with positive probability.  
\end{itemize}

\begin{lemma}\label{lemma-stab-seq}
Under the set up of Theorem~\ref{thm:Instability}, any stable algorithm~\algo must be stable with $n^\star_{a, T} = T/K$ for all arm  $a \in [K]$.
\end{lemma}
\noindent 
We prove Lemma~\ref{lemma-stab-seq} in Appendix~\ref{append-aux-lemma}. Given this lemma at hand it suffices to prove that 
\begin{equation}\label{eqn-main-obj}
    \liminf_{T \to\infty}\prob\bigg(\frac{n_{1,T}(\mathcal{A})}{T} \geq \frac{K+1}{2K},\ \frac{n_{a,T}(\mathcal{A})}{T} \leq \frac{1}{2K} \;\; \text{for all} \;\; a \in \{2,\ldots,K\} \bigg)>0.
\end{equation}
Indeed, it is easy to see that the last bound violates the stability condition~\eqref{defn-stability}. Throughout we assume that $\beta_1, \beta_2 > 0$ in Assumption~\ref{assump:ucb-bound}. The same proof goes through with arbitrary reals $\beta_1, \beta_2$. Let the mean of all the arms be $\mu$ and define the boundary $\bound := \mu-c/\sqrt{T}$ for some $c>0$;
we select the value of $c$ at  a later stage of the proof. Now consider the event
\begin{equation}\label{defn-Ezero}
    \Ezero := \left\{U_a \left(\frac{T}{2K} \right)<\bound \ \forall a \in \{2,\ldots,K\},\ \min_{\frac{T}{2K} \leq t \leq T } U_1(t)> \bound\right\}.
\end{equation}

\noindent Observe that on the event 
$\left\lbrace U_a(T/2K)<\bound \;\;   \text{for all} \;\;  a\ge 2 \;\; \text{and} \;\;  U_1(T/2K)>\bound \right\rbrace$, 
arm $1$ is pulled at time  $T/(2K)+1$. Consequently, we have for all arms $a\geq 2$,
\begin{align*}
    n_{a,\frac{T}{2K}+1}(\mathcal{A}) = n_{a,\frac{T}{2K}}(\mathcal{A}), \qquad  \text{and} 
    \qquad
    \widehat{\mu}_{a,\frac{T}{2K}+1} = \widehat{\mu}_{a,\frac{T}{2K}}.
\end{align*}

\noindent Hence, it follows from Assumption~\ref{assump:H-G} that
\begin{align*}
    U_a \Big(\frac{T}{2K}+1\Big) \leq U_a \Big(\frac{T}{2K}\Big) < \bound.
\end{align*}

\noindent By applying this argument iteratively it follows that on event $\Ezero$ arm $1$ is pulled at every step from $T/(2K)$ to $T$. Hence,
\begin{align}\label{eqn-ezero-cont}
    \Ezero \subset \left\{\frac{n_{1,T}(\mathcal{A})}{T} \geq \frac{K+1}{2K},\ 
    \frac{n_{a,T}(\mathcal{A})}{T} \leq \frac{1}{2K} \ \forall a \ge 2 \right\}.
\end{align}

\noindent Now, let us define the event
\begin{align*}
    \E = \Big\{
    n_{a,\frac{T}{2K}} \in \left[ \frac{T}{4K^2}, \frac{3T}{4K^2} \right]
    \ \text{for all } a\in[K]
    \Big\}.
\end{align*}

\noindent Now suppose that an algorithm satisfying Assumptions~\ref{assump:UCB-func},~\ref{assump:H-G} and~\ref{assump:ucb-bound} is \emph{stable}. By Lemma~\ref{lemma-stab-seq} we have
\begin{align*}
    \frac{n_{a,T/2K}}{T/2K^2} \xrightarrow{\prob} 1
\end{align*}
Consequently, we have $\prob(\E)\to 1$ as $T\to\infty$, and for any sequence of events $(A_T)_{T\ge1}$,
\begin{align}
    \label{eqn:stability-event}
    \liminf_{T\to\infty}\prob(A_T)
    = 
    \liminf_{T\to\infty}\prob(A_T \cap \E).
\end{align}

\noindent In particular, for the choice of $A_T = \Ezero$ we obtain,
\begin{align}\label{eqn-Ezero-E}
    \liminf_{T\to\infty}\prob(\Ezero)
    =
    \liminf_{T\to\infty}\prob(\Ezero \cap \E).
\end{align}

\noindent On the event $\Ezero \cap \E$, Assumption~\ref{assump:ucb-bound} implies that the upper confidence bounds $U_a(t)$ for all $a\in[K]$ lie between $\widetilde{U}_{a,\beta_1}(t)$ and $\widetilde{U}_{a,\beta_2}(t)$ where,
\begin{align*}
    \widetilde{U}_{a,\beta}(t) = \widehat{\mu}_{a,t} + \frac{\beta}{\sqrt{n_{a,t}}}
\end{align*}
\noindent Moreover, on $\E$, there exist constants $\gamma_1,\gamma_2$ (depending on $\beta_1,\beta_2$) such that for each arm $a \in [K]$,
\begin{align}\label{eqn-UV}
\widehat{\mu}_{a,t} + \frac{\gamma_1}{\sqrt{t}}
        \leq \widetilde{U}_{a,\beta_1}(t)
        \leq U_a(t)
        \leq \widetilde{U}_{a,\beta_2}(t)
        \leq \widehat{\mu}_{a,t} + \frac{\gamma_2}{\sqrt{t}}.    
\end{align}

\noindent Motivated by this, we define the \emph{pseudo upper confidence bound} below
\begin{align*}
    \W_{a,\gamma}(t) := \widehat{\mu}_{a,t} + \frac{\gamma}{\sqrt{t}}.
\end{align*}

\noindent Consider the event
\begin{equation}\label{defn-Eone}
    \Eone:= \left\{V_{a,\gamma_2} \left(\frac{T}{2K} \right)<\bound\ \forall a \in \{2,\ldots,K\},\ 
    \min_{\frac{T}{2K}+1 \leq t \leq T } V_{1,\gamma_1}(t)> \bound\right\}.
\end{equation}

\noindent Note that equation~\eqref{eqn-UV} and Assumption~\ref{assump:ucb-bound} imply that  
\begin{align}\label{eqn-EEone-subset}
  \Eone \cap \E \subset \Ezero \cap \E.  
\end{align}
Therefore we obtain the following string of inequalities.  
\begin{align}\label{eqn-Eone-lbd}
    \notag
    &\liminf_{T \to\infty}\prob\bigg(\frac{n_{1,T}(\mathcal{A})}{T} \geq \frac{K+1}{2K},\ \frac{n_{a,T}(\mathcal{A})}{T} \leq \frac{1}{2K} \;\; \text{for all} \;\; a \in \{2,\ldots,K\} \bigg)\\[8pt]\notag
    & \overset{(i)}{\geq}\liminf_{T\to\infty}\prob(\Ezero)\\[8pt]\notag
    & \overset{(ii)}{=}\liminf_{T\to\infty}\prob(\Ezero \cap \E)\\[8pt]\notag
    &\overset{(iii)}{\geq}
    \liminf_{T\to\infty}\prob(\Eone \cap \E) \\[8pt]
    &\overset{(iv)}{=}\liminf_{T\to\infty}\prob(\Eone).
\end{align}

\noindent Inequalities $(i)$ and $(iii)$ follow from equation~\eqref{eqn-ezero-cont} and~\eqref{eqn-EEone-subset}, whereas equalities $(ii)$ and $(iv)$ follow from equations~\eqref{eqn:stability-event} and~\eqref{eqn-Ezero-E}.  Now we apply the following lemma, which states that $\prob(\Eone)$ can be written in terms of probabilities of three different events which we then analyze separately.

\begin{lemma}\label{lemma-MC-decom}
Assume that assumption~\ref{assump:UCB-func} holds. Then we have
\begin{equation}\label{eqn-MC-decom}
    \prob(\mathcal{E}_1(T))
    =
    \frac{D_{1,T}}{D_{2,T}}\, D_{3,T},
\end{equation}
    where
    \begin{align}\label{defn-delta}
        &D_{1,T}
        =
        \prob \bigg(
        V_{a,\gamma_2} \left(\frac{T}{2K} \right)<\bound \ \forall a\ge2,\ 
        V_{1,\gamma_1} \left(\frac{T}{2K} \right)>\bound,\ 
        V_{1,\gamma_1} \left(\frac{T}{2K} +1 \right)>\bound
        \bigg), 
        \notag\\[8pt]
        &D_{2,T}
        =
        \prob \bigg(
        V_{1,\gamma_1} \left(\frac{T}{2K} +1 \right) > \bound
        \bigg),
        \notag\\[8pt]
        &D_{3,T}
        =
        \prob \bigg(
        \min_{\frac{T}{2K}+1 \le t \le T} V_{1,\gamma_1}(t)>\bound
        \bigg).
    \end{align}
\end{lemma}

\noindent Lemma~\ref{lemma-MC-decom} enables us to obtain a non-degenerate lower bound on the term in equation~\eqref{eqn-Eone-lbd} by controlling each of the sequences $D_{i,T}$ individually. This reduction is achieved using the following lemma.

\begin{lemma}\label{lemma-Deltas}
Assume that Assumption~\ref{assump:ucb-bound} holds and suppose $D_{1,T},D_{2,T}$ and $D_{3,T}$ are as defined in equations~\eqref{defn-delta}. Then there exist positive constants $\kappa_1,\kappa_2,\kappa_3$ such that
    \begin{equation}
       \liminf_{T\to\infty} D_{i,T} > \kappa_i \quad \text{for each } i\in\{1,2,3\}.
    \end{equation}
\end{lemma}
\noindent We provide the proof of Lemmas~\ref{lemma-MC-decom} and~\ref{lemma-Deltas} in Appendix~\ref{append-il}. This completes the proof.

\section{Discussion}
\noindent 
Stability is a sufficient and nearly necessary condition for a central limit theorem to hold in the multi-armed bandit problem. This paper establishes general and easily verifiable conditions under which an optimism-based $K$-armed bandit algorithm becomes unstable. These conditions cover a broad class of minimax-optimal UCB and KL-UCB type methods. In particular, we show that widely used minimax-optimal algorithms—including MOSS, Anytime-MOSS, Vanilla-MOSS, ADA-UCB, OC-UCB, KL-MOSS, KL-UCB++, KL-UCB-SWITCH, and Anytime KL-UCB-SWITCH—are unstable. In all these cases, we complement our theoretical guarantees with numerical simulations demonstrating that the sample means fail to exhibit asymptotic normality.

\noindent
Our theoretical results and numerical experiments raise an interesting question: does there exist any optimism-based bandit algorithm that is both minimax optimal and stable? Additional directions include quantifying how instability depends on the reward gap~$\Delta$, and determining whether modifications of MOSS or related minimax strategies could achieve stability under alternative model assumptions. Overall, our findings highlight the need for new adaptive data collection mechanisms that preserve stability while retaining optimal regret performance.

\paragraph{Acknowledgment.}
This work was partially supported by the National Science Foundation Grant DMS-2311304 to Koulik Khamaru.

\bibliographystyle{plainnat}
\bibliography{ref}

\newpage

\appendix 

\section{Proofs of Important Lemmas}\label{append-il}

\noindent In this section, we provide the proofs of Lemma~\ref{lemma-MC-decom} and~\ref{lemma-Deltas} which are the key ingredients in the proof of Theorem~\ref{thm:Instability} in Section~\ref{sec:Instability-proof}. 
\subsection*{Proof of Lemma~\ref{lemma-MC-decom}}

\noindent We begin the proof by fixing two arbitrary constant real numbers $\gamma_1$ and $\gamma_2$. For simplicity, we first prove the theorem for 2-arms and then generalize it for the $K$-arm case.

\subsubsection*{2-arm case:}
\noindent Consider the following process:
\begin{align}\label{defn:Z}
    \textbf{\Z}_t(\gamma_1,\gamma_2) := \begin{bmatrix}
        \W_{1,\gamma_1}(t) \\  \W_{2,\gamma_2}(t)\\ \widehat{\mu}_{1,t} \\  \widehat{\mu}_{2,t} \\ n_{1,t} \\ n_{2,t}
    \end{bmatrix}
\end{align}

\noindent For better readibility of the proofs, we use the shorthand $Z_t$ for $Z_t(\gamma_1, \gamma_2)$.
To see why the introduction of this new process is advantageous, let us consider two projection maps  $g_1$ and $g_2$ on  $\real$, which are defined below. 
\begin{align*} 
    g_{1}(\textbf{\z}) := \textbf{e}_{1}^{T}\textbf{\z} \quad \text{and} \quad g_{2}(\textbf{\z}) := \textbf{e}_{2}^{T}\textbf{z}
\end{align*}

 where $\textbf{e}_{i}^{T}$ is the $i^{th}$ orthonormal basis vector of the standard basis in $\real^{6}$.
 
\noindent Now, let us define sets $A_T,B_T,C_T$ such that $A_T := g^{-1}_2((-\infty,\bound))$, $C_T := g^{-1}_1((\bound, \infty))$ and $B_T := A_T \bigcap C_T$. Using these sets, we can rewrite the following three sets of our interest in terms of the process $(\textbf{Z}_t)_{t \geq 1}$:
\begin{align} \label{defn-markov-sets}
   &\left\{ \W_{1,\gamma_1}(t)>\bound \right\} \equiv \left\{\textbf{\Z}_t \in C_T\right\} \text{,} \notag \\[8pt] &\left\{ \W_{2,\gamma_2}(t)<\bound \right\} \equiv \left\{\textbf{\Z}_t \in A_T\right\} \ \
    \text{and,} \notag \\[8pt] &\ \left\{ \W_{2,\gamma_2}(t)<\bound< \W_{1,\gamma_1}(t) \right\} \equiv \left\{\textbf{Z}_t \in B_T\right\} 
\end{align}

\noindent This reformulation allows us to rewrite the event $\Eone$ of our interest in the following form:
\begin{align}
    \Eone &\equiv \left\{ \W_{2,\gamma_2} \left(\Tfrac \right)<\bound,\  \W_{1,\gamma_1} \left(\Tfrac \right)> \bound , \min_{\Tfrac+2 \leq t \leq T } \W_{1,\gamma_1} \left( \Tfrac \right)> \bound  \right\} \notag \\[6pt]
    &=\left\{\textbf{Z}_\Tfrac \in B_T, \ \textbf{Z}_{\Tfrac+1} \in C_T,\textbf{Z}_t \in C_T \ \ 
    \forall \ \ t \in \bigg[\Tfrac+2,T \bigg] \right\}
    \label{eqn:Eone-decomp}
\end{align}

\noindent The decomposition~\eqref{eqn:Eone-decomp} is useful because of a Markov property of the process $ \{\textbf{Z}_t \}_{t \geq 0}$ which is a Markov chain. We now formally state these Markov properties below:
\begin{lemma}\label{lemma-MC}
    Suppose $\left\{\textbf{Z}_t(\gamma_1, \gamma_2)\right\}_{t \geq 1}$ is the process defined in equation~\eqref{defn:Z}. Then under the assumption~\ref{assump:UCB-func} it is a discrete time Markov chain, where $n_{a,t}$ is the number of times arm $a$ has been pulled up to time $t$, $\widehat{\mu}_{a,t}$ and $U_a(t)$, respectively, denote the sample mean and the UCB of arm $a$ at time $t$.
\end{lemma}

\noindent Lemma~\ref{lemma-MC} allows us to apply Markov property which states that given the present, past and future events are independent (\citet{durrett2019probability}). Concretely, 
\begin{lemma} \label{lemma-mc-indep}
     Suppose that $(\textbf{Q}_t)_{t \geq 1}$ is a finite-dimensional, discrete time Markov Chain. Let $A,B$ and $C$ be arbitrary sets comprising of states in the state space. Then, we have the following equality holds for any fixed value of $m$:
    \begin{align*}
         &\prob \left(\textbf{Q}_{t-1} \in A, \textbf{Q}_{t+i} \in C, i = 1,\ldots,m \;\; \mid \;\; \textbf{Q}_{t} \in B \right) \\[8pt]
         &= \prob\left(\textbf{Q}_{t+i} \in C, i = 1,\ldots,m|\ \textbf{Q}_{t} \in B\right) \times \prob\left( \textbf{Q}_{t-1} \in A \; \mid \;   \textbf{Q}_{t} \in B \right)
     \end{align*}
\end{lemma}

\noindent We prove Lemmas~\ref{lemma-MC} and~\ref{lemma-mc-indep} in Appendix~\ref{append-aux-lemma}. Applying Lemma~\ref{lemma-mc-indep} with $\textbf{Q}_t = \textbf{Z}_t$, $A = B_T , B = C_T , C = C_T$ and $t = \frac{T}{4} + 1$ yields
\begin{align*}
    &=\prob\bigg(\textbf{Z}_{\Tfrac}\in B_T, \ \textbf{Z}_{{\Tfrac+1}} \in C_T,\ \textbf{Z}_{t} \in C_T \   \forall t \ \in \bigg[\Tfrac+2,T \bigg]\bigg)\\[8pt]
    &=\prob\bigg(\textbf{Z}_{\Tfrac} \in B_T, \textbf{Z}_{t} \in C_T \  \forall \ t \in \bigg[{\Tfrac+2},T \bigg] \  \bigg| \ \ \textbf{Z}_{{\Tfrac+1}} \in C_T\bigg) \times \prob\bigg(\textbf{Z}_{\Tfrac+1} \in C_T\bigg)\\[8pt] 
    &\overset{(i)}{=}\prob\bigg(\textbf{Z}_{\Tfrac} \in B_T \ \bigg|\ \ \textbf{Z}_{\Tfrac+1} \in C_T \bigg) \times \prob\bigg(\textbf{Z}_{\Tfrac+1} \in C_T\bigg)\\[8pt]
    &  \times \prob\bigg(\textbf{Z}_{t} \in C_T \ \forall  \ t \in \bigg[\Tfrac+2,T \bigg] \ \bigg| \ \ \textbf{Z}_{\Tfrac+1} \in C_T\bigg)\\[8pt]
    &= \dfrac{\prob\bigg(\textbf{Z}_{\Tfrac} \in B_T,\textbf{Z}_{{\Tfrac+1}} \in C_T \bigg) }{\prob\bigg(\textbf{Z}_{{\Tfrac+1}} \in C_T \bigg)} \times \prob\bigg(\textbf{Z}_{t} \in C_T \ \forall \ t \in \bigg[\Tfrac+1,T \bigg]\bigg)  \\[8pt]
    &= \frac{D_{1,T}}{D_{2,T}} \  D_{3,T}
\end{align*}

\noindent where equality $(i)$  follows from Lemma~\ref{lemma-mc-indep}. In the last equality, we have defined three real sequences $D_{1,T}, D_{2,T}$ and $D_{3,T}$ which are as follows:
\begin{align*}
    & D_{1,T} = \prob\bigg(\textbf{Z}_{\Tfrac} \in B_T, \ \textbf{Z}_{{\Tfrac+1}} \in C_T \bigg) 
    = \prob \bigg( \W_{2,\gamma_2} \left(\Tfrac \right)<\bound,\  \min_{\Tfrac \leq t \leq \Tfrac+1 } \W_{1,\gamma_1}(t)> \bound \bigg) \\[8pt]
    & D_{2,T} = \prob\bigg(\textbf{Z}_{{\Tfrac+1}} \in C_T \in C_T \bigg) = \prob \bigg( \W_{1,\gamma_1} \left(\Tfrac+1 \right)> \bound \bigg)\\[8pt]
    & D_{3,T} = \prob \bigg(\textbf{Z}_{t}  \in C_T, \ \forall \ t \in \bigg[\Tfrac+1,T \bigg] \bigg) = \prob \bigg(  \min_{\Tfrac+1 \leq t \leq T } \W_{1,\gamma_1}(t)> \bound \bigg)
\end{align*}

\subsubsection*{$K$-arm case:}

\noindent The extension to the general $K$ arm case follows analogously by defining the following vectors:
\begin{align*}
    \textbf{\W}_t(\textbf{$\gamma$}) := \begin{bmatrix}
        \W_{1,\gamma_1}(t)\\ \ldots\\ \W_{K,\gamma_K}(t)
    \end{bmatrix}, \quad 
     \textbf{$\widehat{\mu}$}_t(\textbf{$\gamma$}) := \begin{bmatrix}
       \widehat{\mu}_{1,t}\\ \ldots\\ \widehat{\mu}_{K,t}
    \end{bmatrix}, \quad \text{and} \quad 
    \textbf{$n_t(\gamma)$} := \begin{bmatrix}
       n_{1,t}\\ \ldots\\ n_{K,t}
    \end{bmatrix}
\end{align*}

\noindent Now the sane steps of the proof follow by considering the projection maps on the following Markov Chain in the extended state space:
\begin{align*}
    \textbf{Z}_t(\gamma) := \begin{bmatrix}
       \textbf{\W}_{1,\gamma_1}(t)\\ \textbf{$\widehat{\mu}$}_t(\textbf{$\gamma$})\\
       \textbf{$n_t(\gamma)$}
    \end{bmatrix}
\end{align*}
\noindent The proof of the $K$-arm case is now immediate using a similar argument as the $2$-arm case. with the following choices of $D_{1,T}, D_{2,T}$ and $D_{3,T}$.
\begin{align*}
        &D_{1,T} = \prob \left( \W_{a,\gamma_2} \left(\frac{T}{2K} \right)<\bound \ \forall a \in \{2,\ldots,K \},\  \W_{1,\gamma_1} \left(\frac{T}{2K} \right)> \bound,  \W_{1,\gamma_1} \left(\frac{T}{2K} +1 \right)> \bound  \right)  \\[8pt]
        & D_{2,T} = \prob \left( \W_{1,\gamma_1} \left(\frac{T}{2K} \right)> \bound \right) \\[8pt]
        & D_{3,T} = \prob \left(\min_{\frac{T}{2K}+1 \leq t \leq T } \W_{1,\gamma_1}(t)> \bound \right)
    \end{align*}
This completes the proof.

\subsection*{Proof of Lemma \ref{lemma-Deltas}}
\noindent 
Before moving on to the proof of Lemma~\ref{lemma-Deltas}, we start a high level proof ideas and recall a few definitions. 

\paragraph{Notations and high-level proof sketch:}
In this section, we prove that each of the real sequences defined before in equation~\eqref{defn-delta} are lower bounded (in the limit) by a positive constant. For simplicity, we  provide a detailed proof for the two arms case. The proof for the general $K$ arm case follows using a similar argument. Before we begin the proof, let us recall a few key observations which shall be used repeatedly. Invoking Assumption~\ref{assump:ucb-bound}, we note that  on the event $\{n_{a,\Tfrac} \in [T/16,3T/16], \ \text{for} \;\;  a  = 1, 2 \}$ and for any $t \geq T/4$ we have:
\begin{align}\label{eqn-algo-uplow}
     \widehat{\mu}_{a,t} +  \dfrac{\gamma_1}{\sqrt{t}} \leq \  U_a (t) \ \leq \ \widehat{\mu}_{a,t} +   \dfrac{\gamma_2}{\sqrt{t}}
\end{align}
Without loss of generality, let us assume that both $\gamma_1$ and $\gamma_2$ are positive as the same proof goes through for any arbitrary real numbers. Furthermore, for a given value of $\gamma$ define a \emph{pseudo upper confidence bounds} $\W_{a,\gamma}(t)$ for each arm $a$ as:  
\begin{align}
    \W_{a,\gamma}(t):=\widehat{\mu}_{a,t}+g_{\gamma}(t) \;\; 
\end{align}

\noindent The main idea of each of the proofs is that under stability, one can sandwich $U_a(t)$ by the two processes $\W_{a,\gamma_1}(t)$ and $\W_{a,\gamma_2}(t)$ with high probability for large $T$. Finally, we recall the important events:
\begin{subequations}
\begin{align}
   &\E = \Big\{
    n_{a,\frac{T}{2K}} \in \left[ \frac{T}{4K^2}, \frac{3T}{4K^2} \right]
    \ \text{for all } a\in[K]
    \Big\}. \label{eqn:E-recall}  \\[8pt]
    &\Eone:= \left\{V_{a,\gamma_2} \left(\frac{T}{2K} \right)<\bound\ \forall a \in \{2,\ldots,K\},\ 
    \min_{\frac{T}{2K}+1 \leq t \leq T } V_{1,\gamma_1}(t)> \bound\right\}.
    \label{eqn:Eone-recall}
\end{align}
\end{subequations}

\subsubsection*{Main argument:}
\noindent Without loss of generality we assume $\mu = 0$. We are now ready to prove the following three claims from Lemma~\ref{lemma-Deltas}:
\begin{subequations}
\begin{align}
    \liminf_{T \rightarrow \infty} D_{1,T} &> 0, \label{eqn-D-one} \\[6pt]
    \liminf_{T \rightarrow \infty} D_{2,T} &> 0, \label{eqn-D-two} \\[6pt]
    \liminf_{T \rightarrow \infty} D_{3,T} &> 0. \label{eqn-D-three}
\end{align}
\end{subequations}

\subsubsection*{Proof of claim~\eqref{eqn-D-one}:}

\noindent Consider the event
\begin{align}
    \Big\{ \W_{2,\gamma_2} \left(\Tfrac \right) < \bound,\; \W_{1,\gamma_1} \left(\Tfrac \right) > \bound \Big\} \bigcap \E.
\end{align}
On this event, it follows that at step $\Tfrac+1$, arm $1$ is pulled and the empirical mean of arm~$1$ is updated as
\begin{equation*}
    \widehat{\mu}_{1,\Tfrac+1} = \frac{n_{1,\Tfrac}}{n_{1,\Tfrac}+1} \widehat{\mu}_{1,\Tfrac} + \frac{\Rew_{\Tfrac+1}}{n_{1,\Tfrac}+1}.
\end{equation*}

\noindent Under the  event $\{ \W_{2,\Tfrac}(\gamma_2)<\bound, \W_{1,\Tfrac}(\gamma_1)>\bound \} \cap \E$ we obtain the following
\begin{align}\label{eqn-W-approx}
    \W_{1,\gamma_1} \left(\Tfrac +1 \right)
    &= \widehat{\mu}_{1,\Tfrac+1} + \dfrac{\gamma_2}{\sqrt{\Tfrac +1}} \notag\\[8pt]
    &= \frac{n_{1,\Tfrac}}{n_{1,\Tfrac}+1} \widehat{\mu}_{1,\Tfrac} + \frac{\Rew_{\Tfrac+1}}{n_{1,\Tfrac}+1} + \dfrac{\gamma_2}{\sqrt{\Tfrac +1}} \notag\\[8pt]
\end{align}
Therefore from equation~\eqref{eqn-W-approx} we observe that on event $\E$
\begin{align}\label{eqn-V-approx}
    \W_{1,\gamma_1} \left(\Tfrac +1 \right)
    &= \frac{n_{1,\Tfrac}}{n_{1,\Tfrac}+1} \widehat{\mu}_{1,\Tfrac} + \frac{\Rew_{\Tfrac+1}}{n_{1,\Tfrac}+1} + \dfrac{\gamma_1}{\sqrt{\Tfrac +1}} \notag\\[8pt]
    & = \frac{n_{1,\Tfrac}}{n_{1,\Tfrac}+1} \W_{1,\gamma_1} \left(\Tfrac \right) + \frac{\Rew_{\Tfrac+1}}{n_{1,\Tfrac}+1} + g_{\gamma_1} \left(\Tfrac +1\right) -  \frac{n_{1,\Tfrac}}{n_{1,\Tfrac+1}+1}  g_{\gamma_1} \bigg(\Tfrac \bigg) \notag \\[8pt]
    & = \frac{n_{1,\Tfrac}}{n_{1,\Tfrac}+1} \W_{1,\gamma_1} \left(\Tfrac \right) + \frac{\Rew_{\Tfrac+1}}{n_{1,\Tfrac}+1} + \phi(T)
\end{align}

where
\begin{equation*}
    \phi(T) := g_{\gamma_1} \left(\Tfrac +1\right) -  \frac{n_{1,\Tfrac}}{n_{1,\Tfrac+1}+1}  g_{\gamma_1} \bigg(\Tfrac \bigg)
\end{equation*}

\medskip
\noindent By substituting the expression~\eqref{eqn-V-approx} under the joint event $\{ \W_{2,\gamma_2} \left(\Tfrac \right)<\bound, \W_{1,\gamma_1} \left(\Tfrac \right)>\bound \} \cap \E$, we can write
\begin{align*}
    &\prob\Big( \W_{2,\gamma_2} \left(\Tfrac \right)<\bound,\; \W_{1,\gamma_1} \left(\Tfrac \right)>\bound,\; \W_{1,\gamma_1} \left(\Tfrac +1 \right)>\bound,\; \E \Big)\notag\\[8pt]
    & = \prob\Big( \W_{2,\gamma_2} \left(\Tfrac \right)<\bound,\; \W_{1,\gamma_1} \left(\Tfrac \right)>\bound,\; \frac{n_{1,\Tfrac}}{n_{1,\Tfrac}+1}\W_{1,\gamma_1} \left(\Tfrac \right) + \frac{\Rew_{\Tfrac}}{n_{1,\Tfrac}+1} + \phi(T) > \bound,\; \E \Big)\notag \notag\\
\end{align*}

\noindent The last inequality follows from equations~\eqref{eqn-V-approx}. To simplify further, we note the equivalences
\begin{align*}
    &\Big\{\W_{1,\gamma_1} \left(\Tfrac \right)>\bound \Big\} 
    = \Big\{ \frac{n_{1,\Tfrac}}{n_{1,\Tfrac}+1} \W_{1,\gamma_1} \left(\Tfrac \right) > \bound \frac{n_{1,\Tfrac}}{n_{1,\Tfrac}+1} \Big\},\\[8pt]
    &\Big\{ \frac{n_{1,\Tfrac}}{n_{1,\Tfrac}+1}\bound + \frac{\Rew_{\Tfrac +1}}{n_{1,\Tfrac}+1} > \bound - \phi(T) \Big\} = \Big\{ \Rew_{\Tfrac +1} > \bound - (n_{1,\Tfrac}+1)\phi(T) \Big\}, 
\end{align*}

Therefore we have,
\begin{align}
    &\liminf_{T\to\infty} \prob\Big( \W_{2,\gamma_2} \left(\Tfrac \right)<\bound,\; \W_{1,\gamma_1} \left(\Tfrac \right)>\bound,\; \W_{1,\gamma_1} \left(\Tfrac +1 \right)>\bound,\; \E \Big) \notag\\[8pt]
    &\;\;= \liminf_{T\to\infty} \prob\Big( \W_{2,\gamma_2} \left(\Tfrac \right)<\bound,\; \W_{1,\gamma_1} \left(\Tfrac \right)>\bound,\; \Rew_{\Tfrac+1}>\bound + (n_{1,\Tfrac}+1)\phi(T),\; \E \Big).\notag
\end{align}

\noindent Now we shall show that the correction term $(n_{1,\Tfrac+1}+1)\phi(T)$ is asymptotically negligible. Evaluating it explicitly gives
\begin{align*}
    (n_{1,\Tfrac}+1)\phi(T) &= n_{1,\Tfrac} \left[ g_{\gamma_1} \left(\Tfrac +1\right) -  g_{\gamma_1} \bigg(\Tfrac \bigg) \right] + g_{\gamma_1} \left(\Tfrac +1\right) \notag\\[8pt]
    &= -\frac{ 4n_{1,\Tfrac}}{\sqrt{T(T+4)}} \frac{2 \gamma_1}{\sqrt{T} + \sqrt{T+4}} + \frac{2\gamma_1}{\sqrt{T+4}}. \notag
\end{align*}

\noindent Now on the event $\E$, we observe that,
\begin{align*}
   & -(n_{1,\Tfrac}+1) \phi(T) \leq \frac{3}{4}\frac{ T}{\sqrt{T(T+4)}} \frac{2 |\gamma_1|}{\sqrt{T} + \sqrt{T+4}} 
   +\frac{2|\gamma_1|}{\sqrt{T+4}}.
\end{align*}

\noindent Define,
\begin{align*}
    r_T 
    &:= \frac{3}{4}\frac{ T}{\sqrt{T(T+4)}} \frac{2 |\gamma_1|}{\sqrt{T} + \sqrt{T+4}} 
    +\frac{2|\gamma_1|}{\sqrt{T+4}}
\end{align*}

\vspace{2pt}
\noindent Therefore, on the event $\E$, $ -(n_{1,\Tfrac}+1)\phi(T) \leq r_T$ and $r_T$ is a real sequence converging to $0$ as $T\rightarrow \infty$. This implies that on the joint event $\left\{ \W_{2,\gamma_2} \left(\Tfrac \right)<\bound,\;\W_{1,\gamma_1} \left(\Tfrac \right)>\bound,\; \E  \right\}$,
\begin{align*}
    \left\{\Rew_{\Tfrac+1} > \bound + r_T \right\} \subseteq \left\{ \Rew_{\Tfrac+1}>\bound - (n_{1,\Tfrac}+1)\phi(T) \right\} 
\end{align*}

\vspace{4pt}

\noindent Substituting back and applying the stability property~\eqref{eqn:stability-event}, we finally obtain
\begin{align*}
    &\liminf_{T\to\infty} \prob\Big(  \W_{2,\gamma_2} \left(\Tfrac \right)<\bound,\; \W_{1,\gamma_1} \left(\Tfrac \right)>\bound,\; \Rew_{\Tfrac+1}>\bound - (n_{1,\Tfrac}+1)\phi(T),\; \E \Big) \notag\\[8pt]
    &\geq \liminf_{T\to\infty} \prob\Big(  \W_{2,\gamma_2} \left(\Tfrac \right)<\bound,\; \W_{1,\gamma_1} \left(\Tfrac \right)>\bound,\; \Rew_{1\Tfrac+1} > \bound + r_T,\; \E \Big) \notag\\[8pt]
     &= \liminf_{T\to\infty} \prob\Big(  \W_{2,\gamma_2} \left(\Tfrac \right)<\bound,\; \W_{1,\gamma_1} \left(\Tfrac \right)>\bound,\; \Rew_{\Tfrac+1} > \bound + r_T \Big) \notag\\[8pt]
    &= \liminf_{T\to\infty} \prob\Big( \W_{2,\gamma_2} \left(\Tfrac \right)<\bound,\; \W_{1,\gamma_1} \left(\Tfrac \right)>\bound \Big) \times \liminf_{T\to\infty} \prob\Big( \Rew_{\Tfrac+1} > \bound + o(1) \Big). 
\end{align*}

\vspace{8pt}

\noindent The last equality follows because $\Rew_{1.\Tfrac+1}$ is independent~\footnote{See the argument in the proof of Lemma~\ref{lemma-MC}, equation~\ref{eqn-indep}} of history up to time $\Tfrac$. Let $\Rew$ be a identical copy of $\Rew_{\Tfrac}$. Then, from the last equality we conclude that :
\begin{equation}\label{eqn-delta1}
  \liminf_{T \rightarrow \infty} D_{2,T} \geq \Phi \bigg(\frac{\gamma_1-c}{2} \bigg)\Phi \bigg(\frac{c-\gamma_2}{2} \bigg) \prob\bigg(\Rew>0 \bigg) \quad \text{as} \quad T \rightarrow \infty 
\end{equation}

\noindent The proof of the $K$ armed case follows using an exactly same argument, where we  analyze the event 
\begin{align*}
   \prob\Big(  \W_{a,\gamma_2} \left(\Tfrac \right)<\bound \ \forall \ a \in \{2,\ldots K \},\;  \W_{1,\gamma_1} \left(\Tfrac \right)>\bound,\; \W_{1,\gamma_1} \left(\Tfrac + 1 \right)>\bound \Big) 
\end{align*}

\subsubsection*{ Proof of claim~\eqref{eqn-D-two}}

\noindent The proof follows by noting: 
\begin{align*}
    &\left\{\W_{a,\gamma_2} \left(\frac{T}{2K} \right)<\bound \ \forall a \in \{2,\ldots,K \},\  \W_{1,\gamma_1} \left(\frac{T}{2K} \right)> \bound,  \W_{1,\gamma_1} \left(\frac{T}{2K} +1 \right)> \bound \right\} \\
    &\subseteq \left\{\W_{1,\gamma_1} \left(\frac{T}{2K} \right)>\bound \right\}\\
\end{align*}

\subsubsection*{ Proof of claim~\eqref{eqn-D-three}}

\noindent We prove the claim for the two-arm bandit scenario. Extension to the $K$ arm case follows via analogous argument. Now, to quantify the limiting behavior of $(D_{3,t})_{t \geq 1}$, consider the following :
\begin{align}
    D_{3,T}
    & = \prob \bigg( \min_{\frac{T}{2K}+1 \leq t \leq T } \W_{1,\gamma_1} \left(t \right)>\bound \bigg) \notag \\[8pt]
    & = \prob \bigg( \min_{\frac{T}{2K}+1 \leq t \leq T } \W_{1,\gamma_1} \left(t \right)>\bound, \  n_{1,\Tfrac} \in \bigg[\frac{T}{16},\frac{3T}{16} \bigg] \bigg) + o(1) \notag
\end{align}

\noindent In the two armed MAB problem, at step $t+1$ a reward is drawn from one of the arms where the decision to choose the reward distribution is based on the UCB strategy. Consider the reward generating processes for arm $a$ be $(\Rew_{a,t})_{t \geq 1}$. This process is a sequence of i.i.d random variables which are independent of the bandit algorithm. Now for arm $a$ let,
\begin{align}\label{defn-mubar}
    \bar{\mu}_{a,t} := \frac{1}{t}\sum^{t}_{k=1} \Rew_{a,k}
\end{align}
\noindent Before we continue, let us recall $g_\gamma(t) := \gamma/\sqrt{t}$ and define one additional entity,
\begin{align}\label{defn-vstar}
    \W^{*}_{a,\gamma}(t) := \bar{\mu}_{a,t} + g_\gamma(t)
\end{align}

\noindent Therefore, from the definition of $\widehat{\mu}_{a,t}$ and equations~\eqref{defn-mubar} it follows that,
\begin{align}\label{eqn-mustart-mu}
    \widehat{\mu}_{a,t} = \bar{\mu}_{a,n_{a,t}}
\end{align}

\noindent Consequently if $n_{a,t}  \in  \left[ \frac{T}{16}, \frac{3T}{16}  \right] $, from equation~\eqref{eqn-mustart-mu} it follows that,
\begin{align} \label{eqn-muhat-interval}
    \widehat{\mu}_{a,t} \ \in \ \left\{\bar{\mu}_{a,t} \bigg| t \in  \bigg[ \dfrac{T}{16}, \dfrac{3T}{16}  \bigg]\right\}
\end{align}

\noindent Therefore, from equations~\eqref{defn-vstar} and~\eqref{eqn-muhat-interval} it follows that: 
\begin{align*}
     \liminf_{T \rightarrow \infty}D_{3,T}
    & \stackrel{(i)}{\geq}  \liminf_{T \rightarrow \infty} \prob\bigg(\min_{\frac{T}{4}+1 \leq t \leq T }\W_{1,\gamma_1} \left(t \right)>\bound, n_{1,\Tfrac} \in \bigg[\frac{T}{16},\frac{3T}{16} \bigg]  \bigg)\\[8pt]
    & \stackrel{(ii)}{\geq}  \liminf_{T \rightarrow \infty}\prob\bigg(\min_{\frac{T}{16} \leq t \leq T }\W^\star_{1,\gamma_1} \left(t \right)>\bound, n_{1,\Tfrac} \in \bigg[\frac{T}{16},\frac{3T}{16} \bigg]  \bigg)\\[8pt]
    & \stackrel{(iii)}{=} \liminf_{T \rightarrow \infty}\prob\bigg(\min_{\frac{T}{16} \leq t \leq T } \W^\star_{1,\gamma_1} \left(t \right)>\bound  \bigg)\\[8pt]
    & = 1-\limsup_{T \rightarrow \infty}\prob\bigg(\min_{\frac{T}{16} \leq t \leq T }\W^\star_{1,\gamma_1} \left(t \right) \leq \bound  \bigg)\\[8pt]
    & = 1-\limsup_{T \rightarrow \infty}\prob\bigg(\max_{\frac{T}{16} \leq t \leq T } (-\W^\star_{1,\gamma_1} \left(t \right)) \geq -\bound  \bigg)
\end{align*}
\noindent Inequality $(i)$ and equality $(iii)$ follow from stability~\eqref{eqn:stability-event}. Inequality $(ii)$ follows from equation~\eqref{eqn-muhat-interval}. Now, the lemma stated below completes the proof.
\begin{lemma}\label{lemma-max2}
Suppose $(\Rew_{a,t})_{t\geq 1}$ are centered $1$-subgaussian random variables and $-\X^{*}_{1,t}(\beta_1)$ is as defined in equation~\eqref{defn-vstar}. Then there exists an $\alpha \in (0,1)$ and a $c_{\alpha}>0$ such that for $\bound = -c_{\alpha}/\sqrt{T}$ we have :
    \begin{equation*}
        \limsup_{T \rightarrow \infty}\prob\bigg(\max_{\frac{T}{16} \leq t \leq T } (-\W^\star_{1,\gamma_1} \left(t \right)) \geq -\bound  \bigg) \leq \alpha
    \end{equation*}
\end{lemma}

\noindent We note that the random process $(\X^\star_{1,t}(\beta_1))_{t \geq 1}$ is neither a sub-martingale nor a super-martingale, and as consequence, standard maximal concentration inequalities like Doob's inequality cannot be applied. To prove this lemma we shall apply Theorem~\ref{thm-maximal} and the proof is provided in the next section.

\subsection*{Proof of Lemma~\ref{lemma-max2}}

\noindent Let us recall from equations~\eqref{defn-mubar} and~\eqref{defn-vstar} that
\begin{align*}
    \W^{*}_{a,\gamma}(t) := \bar{\mu}_{a,t} + g_\gamma(t) = \frac{1}{t}\sum^{t}_{k=1} \Rew_{a,t} + g_\gamma(t) 
\end{align*}
where $(\Rew_{a,t})_{t\geq 1}$ are centered $1$-subgaussian random variables and $g_\gamma(t) := \gamma/\sqrt{t}$. To prove this lemma we shall use the following maximal inequality for un-centered sampled means which we state below.
\begin{lemma}\label{lemma-maximal}
    Suppose $X_1,\ldots,X_n$ are i.i.d. $1$-subgaussian random variables with mean $0$ and variance $\sigma_X$. Let $g(n)$ be any real decreasing function and fix $\lambda>0$. Then there exists absolute constant $\tilde{C}>0$ for which we have, 
    \begin{align}
    \notag
     \lambda \prob \bigg(\max_{\alpha N \leq n \leq \beta N} \left[ \bar{X}_{n+1} + g(n+1)\right] \geq \lambda \bigg)  
     &\leq \frac{1}{\sqrt{N}} \left( \tilde{C}\sqrt{\frac{2}{\pi}} e^{-Ng(\beta N)^2/2} + 2\tilde{C}\right) \ +\ \bigg( g(\alpha N) - g(\beta N) \bigg) \\[8pt] 
    & + g(\beta N) \bigg(1-2\Phi\bigg( -\frac{\sqrt{N}g(\beta N)}{\sigma_X}\bigg) \bigg)+ \sum ^{\beta N -1}_{ k =\alpha N}\frac{1}{(k+1)\sqrt{k}} \sqrt{\frac{2}{\pi}} 
\end{align}
\end{lemma}

\noindent Therefore, for the choice of $\bound = -c/\sqrt{T}$ we obtain the following :
\begin{align}\label{eqn-upp-bdd1}
    -\frac{1}{\lambda_T\sqrt{T}} \left( \tilde{C}\sqrt{\frac{2}{\pi}} e^{-Tg_{\gamma_1}(T)^2/2} + 2\tilde{C}\right) = \frac{1}{c}\bigg[ \tilde{C} \sqrt{\frac{2}{\pi}} e^{-(\beta_1)^2/2} + 2\tilde{C} + \gamma \bigg(1-2\Phi\bigg( -\beta_1\bigg) \bigg) \bigg]
\end{align}

\noindent and,
\begin{align}\label{eqn-upp-bdd2}
     \frac{\sqrt{T}}{c}\sqrt{\frac{2}{\pi}} \sum ^{ T -1}_{ k =T/16} \frac{1}{(k+1)\sqrt{k}}\leq \frac{4}{c}\sqrt{\frac{2}{\pi}}\frac{16T}{T+16} \leq \frac{4}{c}\sqrt{\frac{2}{\pi}}
\end{align}

\noindent The second last inequality in the above equation holds because $k\geq T/16$. Finally, we have :
\begin{align}\label{eqn-upp-bdd3}
    \frac{\sqrt{T}}{c}\bigg(g \left(\frac{T}{16} \right) - g(T) \bigg) = \frac{\sqrt{T}}{c} \bigg(\frac{16}{\sqrt{T}} - \frac{1}{\sqrt{T-1}}\bigg)
    \leq \dfrac{16}{c}
\end{align}
     
\noindent Therefore,for any choice of $\alpha \in (0,1)$, it follow from equations~\eqref{eqn-upp-bdd1}, \eqref{eqn-upp-bdd2} and~\eqref{eqn-upp-bdd3} it follows that by taking $c$ sufficiently large,      
\begin{align*}
    \prob\bigg(\text{max}_{t \in [\frac{T}{16},T ]} (-\W^{*}_{a,\gamma_1}(t)) \geq -\bound  \bigg) < \alpha
\end{align*}

\noindent This completes our proof. Lemma~\ref{lemma-maximal} is derived in the next section, which requires application of a generalized Doob's inequality. Before we conclude this section here is an important remark. From equation~\eqref{eqn-norm-mean} it follows that for \textit{any} choice of $\gamma$ in $\real$ it follows that,
\begin{align} \label{eqn-norm-mean-gam}
     \sqrt{\frac{2}{\pi}} e^{-\gamma^2/2} + \gamma \bigg(1-2\Phi( -\gamma) \bigg) >0
\end{align}

\noindent This is because equation~\eqref{eqn-norm-mean-gam} corresponds to equation~\eqref{eqn-norm-mean} when $\mu = \gamma$ and $\sigma = 1$ as $\Ex[|V|]>0$. This observation allows us to be flexible in the choice of constants $\gamma_1$ and $\gamma_2$ in Theorem~\ref{thm:Instability}.

\newpage

\section{Maximal inequality for sample means}

\noindent In this section we derive a maximal inequality for centered sample means where the reward distribution is sub Gaussian. As noted in the earlier section, the process $(\bar{X}_n)$ is not a martingale, sub-martingale or super-martingale. Consequently, standard maximal inequality of Doob and Kolmogorov cannot be applied. Here we first derive a maximal inequality for a general class of processes and then apply that result to derive our desired inequality.

\subsection{Maximal inequality for a general class of processes}

\noindent In this section we discuss a result for random processes of a specific kind which generalize the standard Doob's maximal inequality. Consider a stochastic process $(Q_t)_{t \geq 1}$ and let $(\Fil_t)_{t \geq 1}$ be the natural filtration w.r.t this process. Assume that a decomposition of the following kind holds:
 \begin{equation} \label{eqn-decom}
         \Ex\bigg[Q_{t+1} \bigg| \Fil_{t}\bigg] = Q_{t} + R_{t} 
     \end{equation}
where $R_t\ \in \ \Fil_t$.

\noindent We note that if $R_t \leq 0$ (almost surely), then $(Q_t,\Fil_t)_{t \geq 1}$ is a sub-maringale (analogously we have super-martingales). However, it may happen that $R_t$ is a random variable which can take both positive and negative values then the process $(Q_t,\Fil_t)_{t \geq 1}$ is neither a sub-martingale nor a super-martingale. Hence, our setup is more general. For such processes, we have the following result.

\begin{tcolorbox}
\begin{subequations}
\begin{prop}
\label{thm-maximal}
    Fix $\lambda>0$ and suppose that $\alpha<\beta$ are distinct, arbitrary constants in $(0,1]$, and $Q_k,R_k$ are as defined in Lemma~\ref{eqn-decom} for each $k$. Then, we have the following maximal inequality :
    \begin{equation}\label{eqn-maximal}
        \prob \bigg(\max_{\alpha T \leq t \leq \beta T} Q_{t} \geq \lambda \bigg) \leq \frac{1}{\lambda} \bigg( \Ex \bigg[ |Q_{\beta T}| \bigg] + \sum ^{\beta T -1}_{ k =\alpha T}\Ex \bigg[  |R_{k} |\bigg] \bigg)
    \end{equation} 
 where 
\end{prop}
\end{subequations}
\end{tcolorbox}

\noindent We note that when $R_t  = 0$ almost surely then this result reduces to Doob's maximal inequality. The proof of this proposition follows the standard approach. Let us begin by defining a stopping time $\tau$, defined as follows :
\begin{align}\label{defn-tau}
    \tau := \text{min} \bigg(\text{inf} \left\{k \geq \alpha T  \ \bigg| \max_{\alpha T \leq t \leq k} Q_{t} \geq \lambda  \right\}, \beta T \bigg)
\end{align}

\noindent Now, consider the following lemma :
\begin{lemma}\label{lemma-M-alpha}
Consider the stopping time $\tau$ as defined in equation~\eqref{defn-tau} and let $\Fil_{\tau}$ be the stopped $\sigma$-field. Then we have the following two results:
    \begin{enumerate}
        \item[a] $\left\{ \max_{\alpha T \leq t \leq \beta T} Q_{t} \geq \lambda \right\}$ belongs in the stopped $\sigma$ field $\Fil_{\tau}$.
        \item[b] $\left\{ \max_{\alpha T \leq t \leq \beta T} Q_{t} \geq \lambda \right\} = \left\{Q_{\tau} \geq \lambda\right\}$ 
    \end{enumerate}    
\end{lemma}

\noindent By applying Lemma~\ref{lemma-M-alpha}, we obtain the following :
\begin{equation}\label{eqn-maximal-apply}
   \lambda\prob \bigg(\max_{\alpha T \leq t \leq \beta T} Q_{t} \geq \lambda \bigg) 
    = \Ex \bigg[ \lambda  \mathbb{I}_{\left\{ \max_{\alpha T \leq t \leq \beta T} Q_{t} \geq \lambda\right\}} \bigg]
    = \Ex \bigg[ \lambda  \mathbb{I}_{\left\{Q_{\tau} \geq \lambda\right\}} \bigg]
    \leq \Ex \bigg[Q_{\tau}  \mathbb{I}_{\left\{Q_{\tau} \geq \lambda \right\}} \bigg] 
\end{equation}

\begin{lemma}\label{thm-ost}
     Suppose $\tau_1,\tau_2$ are two stopping times such that $\prob(\tau_1 \leq \tau_2 \leq n) = 1$ (where $n$ is fixed). Then, for any event $A\in \Fil_{\tau_1}$, we have :
    \begin{equation}\label{eqn-OST}
        \Ex\bigg [Q_{\tau_1} \mathbb{I}_{\left\{A\right\}}\bigg] = \Ex\bigg [Q_{\tau_2} \mathbb{I}_{\left\{A\right\}} - \bigg(\sum^{\tau_2 -1}_{\tau_1} R_k \bigg) \mathbb{I}_{{\left\{A\right\}}} \mathbb{I}_{\left\{\tau_2 > \tau_1\right\}} \bigg]
    \end{equation} 
    where $\Fil_{\tau_1}$ is the stopped $\sigma$-field (w.r.t $\tau_1$) defined as $\left\{ A : A\cap\left\{\tau_1 = k\right\} \in \Fil_k, \text{for all} \quad k \in \mathcal{N}\right\}$.
\end{lemma}

\noindent This lemma is a direct generalization of the standard OST result. Now, suppose this lemma holds. Then by replacing $\tau_1 = \tau$ and $\tau_2 = \beta T$ in Lemma~\ref{thm-ost} we have :
\begin{equation}\label{eqn-ex-decom}
    \Ex \bigg[ Q_{\tau} \mathbb{I}_{\left\{Q_{\tau} \geq \lambda\right\}} \bigg] = \Ex\bigg [Q_{\beta T} \mathbb{I}_{{\left\{Q_{\tau} \geq \lambda \right\}}} - \bigg(\sum^{\beta T -1}_{k=\tau}  R_{k-1} \bigg) \mathbb{I}_{{\left\{Q_{\tau} \geq \lambda\right\}}} \mathbb{I}_{{\left\{\beta T > \tau\right\}}} \bigg]
\end{equation}

\noindent Therefore, from equations~\eqref{eqn-maximal-apply} and~\eqref{eqn-ex-decom} we have the following chain of inequalities :
\begin{align*}
   \lambda\prob \bigg(\max_{\alpha T \leq t \leq \beta T} Q_{t} \geq \lambda \bigg) 
   & \leq \Ex\bigg [Q_{\beta T} \mathbb{I}_{{\left\{Q_{\tau} \geq \lambda \right\}}} - \bigg(\sum^{\beta T -1}_{k=\tau} R_{1,k} \bigg) \mathbb{I}_{{\left\{Q_{\tau} \geq \lambda\right\}}} \mathbb{I}_{{\left\{\beta T > \tau\right\}}} \bigg]\\[8pt]
   & \leq \Ex\bigg [ \bigg| Q_{\beta T} \mathbb{I}_{{\left\{Q_{\tau} \geq \lambda \right\}}} - \bigg(\sum^{\beta T -1}_{k=\tau}  R_{k} \bigg) \mathbb{I}_{{\left\{Q_{\tau} \geq \lambda\right\}}} \mathbb{I}_{{\left\{\beta T > \tau\right\}}}  \bigg]\\[8pt]
   & \leq \Ex\bigg [ | Q_{\beta T}| \mathbb{I}_{{\left\{Q_{\tau} \geq \lambda \right\}}} + \bigg(\sum^{\beta T -1}_{k=\tau} | R_{k}| \bigg) \mathbb{I}_{{\left\{Q_{\tau} \geq \lambda\right\}}} \mathbb{I}_{{\left\{\beta T > \tau\right\}}} \bigg]\\[8pt]
   & \leq \Ex\bigg [ | Q_{\beta T}| \mathbb{I}_{{\left\{Q_{\tau} \geq \lambda \right\}}} \bigg] + \Ex\bigg[\bigg(\sum^{\beta T -1}_{k=\tau} | R_{k}| \bigg) \mathbb{I}_{{\left\{Q_{\tau} \geq \lambda\right\}}} \mathbb{I}_{{\left\{\beta T > \tau\right\}}} \bigg]\\[8pt]
   & \leq \Ex\bigg [ | Q_{\beta T}| \bigg] + \Ex\bigg[\sum^{\beta T -1}_{k=\tau} | R_{k}| \bigg]
\end{align*}
This completes the proof of Lemma~\ref{thm-maximal}, once we prove Lemma~\ref{thm-ost}. We note that as $\tau_1 \leq \tau_2 \leq n$, we have 
\begin{align*}
    \Ex\bigg[\bigg( Q_{\tau_2} - Q_{\tau_1}\bigg) \mathbb{I}_{\{A\}}\bigg]
    =\Ex\bigg[\bigg( Q_{\tau_2} - Q_{\tau_1}\bigg) \mathbb{I}_{\left\{A\right\}} \mathbb{I}_{\left\{ \tau_2 > \tau_1\right\}}\bigg]\\
\end{align*}
This is true because on the event $\left\{ \tau_1 = \tau_2 \right\}$, $ Q_{\tau_2} - Q_{\tau_1} = 0$. This observation leads us to the following set of equalities.
\begin{align*}
    \Ex\bigg[\bigg( Q_{\tau_2} - Q_{\tau_1}\bigg) \mathbb{I}_{\left\{A\right\}}\bigg]
    &=\Ex\bigg[\bigg( Q_{\tau_2} - Q_{\tau_1}\bigg) \mathbb{I}_{\left\{A\right\}} \mathbb{I}_{\left\{ \tau_2 > \tau_1\right\}}\bigg]\\[8pt]
    &=\Ex\bigg[\sum^{\tau_2}_{\tau_1+1}\bigg( Q_{k} - Q_{k-1}\bigg) \mathbb{I}_{\left\{ \tau_2 > \tau_1, A\right\}} \bigg]\\[8pt]
    & = \Ex\bigg[\sum^{n}_{2}\bigg( Q_{k} - Q_{k-1}\bigg) \mathbb{I}_{\left\{\tau_1 < k \leq \tau_2, A\right\}} \bigg]\\[8pt]
    & = \sum^{n}_{2}\bigg(\Ex\bigg[ Q_{k} \mathbb{I}_{\left\{\tau_1 < k \leq \tau_2, A\right\}}\bigg] - \Ex\bigg[Q_{k-1} \mathbb{I}_{\left\{\tau_1 < k \leq \tau_2, A\right\}} \bigg]\bigg)  
\end{align*}

\noindent Now, as $\tau_1$ and $\tau_2$ are both stopping times, $\left\{\tau_1 \leq k-1\right\}$ and $\left\{\tau_2 \leq k-1\right\}^{c}$ lie in the $\sigma$-field $\Fil_{k-1}$. Furthermore, as $A \in \Fil_{\tau_1}$, it implies that $\left\{\tau_1 \leq k-1\right\} \cap A$ measurable in $\Fil_{k-1}$. Hence, we conclude that $\left\{\tau_1 \leq k-1\right\} \cap A\cap \left\{\tau_2 \leq k-1\right\}^{c}$ belongs to $\Fil_{k-1}$. This is equivalent to the statement $A\cap\left\{\tau_1 < k \leq \tau_2 \right\} \in \Fil_{k-1}$, for any $A \in \Fil_{\tau_1}$. Therefore, from equation~\eqref{eqn-decom} we have:
\begin{equation} \label{eqn-ost-conde}
    \Ex\bigg[Q_{k}1\left\{\tau_1 < k \leq \tau_2, A\right\} \bigg]
    = \Ex\bigg[Q_{k-1}1\left\{\tau_1 < k \leq \tau_2, A\right\} \bigg] 
    + \Ex\bigg[\bigg(  R_{k-1} \bigg) 1{\left\{\tau_1 < k \leq \tau_2, A\right\}}\bigg]
\end{equation}
\noindent Therefore, by applying equation~\eqref{eqn-ost-conde} in the previous chain of inequalities, we obtain :
\begin{align*}
    \Ex\bigg[\bigg( Q_{\tau_2} - Q_{\tau_1}\bigg) \mathbb{I}_{\left\{A\right\}} \bigg]
    &= \sum^{n}_{2} \Ex\bigg[\bigg( R_{k-1} \bigg) \mathbb{I}_{\left\{\tau_1 < k \leq \tau_2, A \right\}} \bigg] \\[8pt]
    & = \Ex\bigg[\sum^{n}_{2}\bigg( R_{k-1}\bigg) \mathbb{I}_{\left\{ \tau_1 < k \leq \tau_2, A \right\}} \bigg]\\[8pt]
    & = \Ex\bigg[\sum^{\tau_2}_{\tau_1+1}\bigg( R_{k-1}\bigg) \mathbb{I}_{\left\{\tau_2 > \tau_1 \right\}} \mathbb{I}_{ \left\{ A\right\}} \bigg]
\end{align*}

\noindent Therefore, by rearranging the terms, we have our result.
\begin{align*}
    \Ex\bigg [Q_{\tau_1} \mathbb{I}_{{\left\{A\right\}}} \bigg] = \Ex\bigg [Q_{\tau_2} \mathbb{I}_{{\left\{A\right\}}} - \bigg(\sum^{\tau_2 -1}_{\tau_1} R_{k}  \bigg) \mathbb{I}_{{\left\{A\right\}}} \mathbb{I}_{{\left\{\tau_2 > \tau_1\right\}}} \bigg]
\end{align*}

\subsection{Application to  sample means}

\noindent Let $g(n)$ be any real valued function and suppose that $X_1,\ldots,X_n$ are iid standard gaussian random variables. We wish to derive a maximal inequality for the process $(\bar{X}_n + g(n))_{n \geq 1}$. 
Then we have Lemma~\ref{lemma-maximal} stated in the earlier section.
\begin{lemma*}
    Suppose $X_1,\ldots,X_n$ are i.i.d. $1$ sub-Gaussian random variables with mean $0$ and variance $\sigma_X$. Let $g(n)$ be any real decreasing function and fix $\lambda>0$. Then there exists absolute constant $\tilde{C}>0$ for which we have, 
    \begin{align}
    \notag
     \lambda \prob \bigg(\max_{\alpha N \leq n \leq \beta N} \left[ \bar{X}_{n+1} + g(n+1)\right] \geq \lambda \bigg)  
     &\leq \frac{1}{\sqrt{N}} \left( \tilde{C}\sqrt{\frac{2}{\pi}} e^{-Ng(\beta N)^2/2} + 2\tilde{C}\right) \ +\ \bigg( g(\alpha N) - g(\beta N) \bigg) \\[8pt] 
    & + g(\beta N) \bigg(1-2\Phi\bigg( -\frac{\sqrt{N}g(\beta N)}{\sigma_X}\bigg)+ \sum ^{\beta N -1}_{ k =\alpha N}\frac{1}{(k+1)\sqrt{k}} \sqrt{\frac{2}{\pi}} 
\end{align}
\end{lemma*}

\noindent The proof is a direct application of Proposition~\eqref{thm-maximal}. We can apply this lemma due to the following observation.
\begin{align}\label{eqn-xbar-decom}
    \bar{X}_{n+1} + g(n+1) \notag
    & = \dfrac{n}{n+1} \bar{X}_n + \dfrac{X_{n+1}}{n+1} + g(n+1) \\[8pt] \notag
    & = \bar{X}_n + \dfrac{X_{n+1} - \bar{X}_n}{n+1} + g(n+1) \\[8pt] 
    & = \left[\bar{X}_n + g(n) \right] +  \dfrac{X_{n+1} - \bar{X}_n}{n+1} + \left[g(n+1)-g(n) \right] 
\end{align}

\noindent Let $\mathcal{F}_n$ be the $\sigma$-field generated by $X_1,\ldots,X_n$ ]]. Then it follows that $\bar{X}_n$ and $g(n)$ is measurable with respect to $\mathcal{F}_n$ and $X_{n+1}$ is independent of $\mathcal{F}_n$. Therefore we have,
\begin{align}\label{eqn-condex-xbar}
   \Ex\bigg[\bar{X}_{n+1} + g(n+1) \bigg| \Fil_{t}\bigg] 
   =\left[\bar{X}_n + g(n) \right] +  \left[-\dfrac{\bar{X}_n}{n+1} + (g(n+1)-g(n)) \right] 
\end{align}

\noindent By comparing equations~\eqref{eqn-condex-xbar} and~\eqref{eqn-decom} and applying Proposition~\ref{thm-maximal} we obtain the following maximal inequality:
\begin{align}
    \notag
    &\prob \bigg(\max_{\alpha N \leq n \leq \beta N} \left[ \bar{X}_{n+1} + g(n+1)\right] \geq \lambda \bigg) \\[8pt] \notag
    &\leq \frac{1}{\lambda} \bigg( \Ex \bigg[ |\bar{X}_{\beta N} + g(\beta N)| \bigg] + \sum ^{\beta N -1}_{ k =\alpha N}\Ex \bigg[  \bigg|-\dfrac{\bar{X}_k}{k+1} + (g(k+1)-g(k)) \bigg| \bigg] \bigg) \\[8pt]
    &\leq \frac{1}{\lambda} \bigg( \Ex \bigg[ |\bar{X}_{\beta N} + g(\beta N)| \bigg] + \sum ^{\beta N -1}_{ k =\alpha N}\Ex \bigg[  \bigg|\dfrac{\bar{X}_k}{k+1}\bigg|\bigg] + \sum ^{\beta N -1}_{ k =\alpha N}\Ex \bigg[ \bigg| (g(k+1)-g(k)) \bigg| \bigg] \bigg)
\end{align}

\noindent If the random variables $X_1, \ldots, X_n$ are Gaussian random variables, we can write the first moment of the absolute values more explicitly. To do so we shall apply the following well known result (\citet{leone1961folded}).
\begin{lemma}\label{eqn-norm-mean}
     If $\X\sim N(\mu,\sigma^2)$, then :
    \begin{equation*}
        \Ex[ |\X|] = \sigma \sqrt{\frac{2}{\pi}} e^{-\mu^2/2\sigma^2} + \mu \bigg(1-2\Phi\bigg( -\frac{\mu}{\sigma}\bigg) \bigg)
    \end{equation*}
\end{lemma}

\noindent By using this lemma, we have:
\begin{equation*}
    \Ex \bigg[ |\bar{X}_{\beta N} + g(\beta N)| \bigg] = \frac{1}{\sqrt{N}} \sqrt{\frac{2}{\pi}} e^{-Ng(\beta N)^2/2} + g(\beta N) \bigg(1-2\Phi\bigg( -\sqrt{T}g(\beta N)\bigg) \bigg)  
\end{equation*}
\noindent and,
\begin{equation*}
    \Ex\bigg[ \bigg|\dfrac{\bar{X}_k}{k+1}\bigg|  \bigg] = \frac{1}{(k+1)\sqrt{k}} \sqrt{\frac{2}{\pi}}
\end{equation*}

\noindent Let us make a simplifying assumption that $g$ is a decreasing function. Then the above calculations yield the explicit upper bound stated below.
\begin{align*}
    & \lambda \prob \bigg(\max_{\alpha N \leq n \leq \beta N} \left[ \bar{X}_{n+1} + g(n+1)\right] \geq \lambda \bigg) \\[8pt]
    & \leq \frac{1}{\sqrt{N}} \sqrt{\frac{2}{\pi}} e^{-Ng(\beta N)^2/2} + g(\beta N) \bigg(1-2\Phi\bigg( -\sqrt{T}g(\beta N)\bigg) + \sum ^{\beta N -1}_{ k =\alpha N}\frac{1}{(k+1)\sqrt{k}} \sqrt{\frac{2}{\pi}} +\bigg( g(\alpha N) - g(\beta N) \bigg)
\end{align*}
\noindent This concludes the proof for the case when the reward distribution is Gaussian. The extension to the case of sub-gaussian rewards follows directly from the following lemma, which is proved in the next section.
\begin{lemma}\label{lemma-subG}
    Suppose $X_1,\ldots,X_n$ are i.i.d. $1$-sub-Gaussian distribution with mean $\mu$. Then we have :
    \begin{align*}
        \Ex[ |\bar{X}_n|] \leq \frac{1}{\sqrt{n}} \bigg( \tilde{C} \sqrt{\frac{2}{\pi}} e^{-\mu^2n/2} + 2\tilde{C} \bigg) + \mu \bigg(1-2\Phi\bigg( -\frac{\mu \sqrt{n}}{\sigma_X}\bigg) \bigg)
    \end{align*}
    where $\sigma_X$ is the variance of $X_1$ and  $\tilde{C} >0$ is a universal constant.
\end{lemma}

\subsection*{Proof of Lemma~\ref{lemma-subG}}

\noindent Suppose $X_1,\ldots,X_n$ are sub-Gaussian random variables. Recall that a random variable $X$ is called $\sigma$-sub-Gaussian, if the following inequality holds $\forall \ \lambda >0$ :
\begin{align*}
    \Ex[e^{\lambda(X-\Ex[X])}] \leq e^{\sigma^2 \lambda^2 /2}
\end{align*}

\noindent Hence if $X$ is a $\sigma$-sub-Gaussian random variable, then it is also $\tilde{\sigma}$-sub-Gaussian for any $\tilde{\sigma}>\sigma$. One can define a unique variance proxy for sub-gaussian random variables. This is called \emph{optimal variance proxy} (\citet{arbel2020strict,vershynin2018high}) which we define below:
\begin{align}
    \operatorname{Var}_G(X) := \inf_{\sigma>0} \left\{\Ex[e^{\lambda(X-\Ex[X])}] \leq e^{\sigma^2 \lambda^2 /2} \ \  \forall \ \ \lambda > 0  \right\}
\end{align}

\noindent It has several well documented properties (\citet{vershynin2018high}), which we state below in form of a lemma.
\begin{lemma}\label{lemma-vopt}
    Suppose $X_1,\ldots,X_n$ are centered sub-Gaussian random with variance $\sigma_{X_i}^2$. Then the optimal variance proxy satisfies the following properties:
    \begin{enumerate}
        \item $\vopt(cX_i) = c^2 \vopt(X_i)$.
        \item $\sigma_{X_i}^2 \leq \vopt(X_i)$ for all $i \in [n]$.
        \item $\vopt(\sum^{n}_{i=1} X_i) \leq \sum^{n}_{i=1}\vopt(X_i) $.
        \item There exists absolute constants $c_1,c_2>0$ such that $c_1 || X_i ||_{\psi_2} \leq \sqrt{\vopt(X_i)} \leq c_2 || X_i ||_{\psi_2}$.
    \end{enumerate}
\end{lemma}

\noindent In property $4$ of $\vopt(X_i)$, $|| X_i ||_{\psi_2}$ is the \emph{Orlicz} norm of $X$ with respect to the function $\psi_2(x):= e^{x^2}-1$. Let us begin our proof. It follows that if we have i.i.d. sub- Gaussian random variables $X_1,\ldots,X_n$ it follows from Lemma~\ref{lemma-vopt} that
\begin{align*}
    \vopt(\bar{X}_n) \leq \frac{1}{n} \vopt(X_1)
\end{align*}

\noindent Furthermore, an alternate characterization for sub gaussian random variable states that if $X$ is any sub-Gaussian random variable then we have
\begin{align}\label{eqn-subG-char}
    \Ex[|X|] \leq C ||X||_{\psi_2}
\end{align}

\noindent Therefore, if $X$ is a sub-Gaussian random variable then the following inequality follows from Equation~\eqref{eqn-subG-char} and Lemma~\ref{lemma-vopt} that there exists absolute constant $\tilde{C}$ such that
\begin{align}
    \Ex[|X|] \leq \tilde{C} \ \sqrt{\vopt(X)}
\end{align}

\noindent Now we shall use this result to obtain an upper bound on $\Ex[|\bar{X}_n + \mu|]$. Let $Z$ be a normal random variable with mean $0$ and variance $\vopt(\bar{X}_n)$. Then we have
\begin{align*}
    \Ex[|\bar{X}_n + \mu|] 
     & \leq \Ex[|Z+\mu|] + \Ex[|\bar{X}_n-Z|] \\[8pt]
     & = \sqrt{\vopt(\bar{X}_n)} \sqrt{\frac{2}{\pi}} e^{-\mu^2/2\vopt(\bar{X}_n))^2} + \mu \bigg(1-2\Phi\bigg( -\frac{\mu}{\sqrt{\vopt(\bar{X}_n)}}\bigg) \bigg)
     + \Ex[|\bar{X}_n-Z|]
\end{align*}

\noindent We control the second term by applying Equation~\eqref{eqn-subG-char} and Lemma~\ref{lemma-vopt} as follows
\begin{align*}
    \Ex[|\bar{X}_n-Z|] 
    & \leq C||\bar{X}_n-Z||_{\psi_2} \\[8pt]
    & \leq \tilde{C} \sqrt{\vopt(\bar{X}_n-Z)} \\[8pt]
    & \leq \tilde{C} \sqrt{\vopt(\bar{X}_n) + \vopt(Z)} \\[8pt]
    & \leq \tilde{C} \left[ \sqrt{\vopt(\bar{X}_n)} + \sqrt{\vopt(Z)} \right] \\[8pt]
    & \leq 2 \tilde{C} \dfrac{\sqrt{\vopt(X_1)}}{\sqrt{n}}
\end{align*}

\noindent Hence we have,
\begin{align*}
    \Ex[|\bar{X}_n-Z|]  
    &\leq \tilde{C} \dfrac{\sqrt{\vopt(X_1)}}{\sqrt{n}} \sqrt{\frac{2}{\pi}} e^{-\mu^2/2\vopt(\bar{X}_n))^2} + \mu \bigg(1-2\Phi\bigg( -\frac{\mu}{\sqrt{\vopt(\bar{X}_n)}}\bigg) \bigg)\\
     &+ \tilde{C} \dfrac{\sqrt{\vopt(X_1)}}{\sqrt{n}}
\end{align*}

\noindent Now since $X_1,\ldots,X_n$ are $1$-sub-Gaussian random variables, it follows from the definition of optimal variance proxy that $\vopt(X_1) \leq 1$. Furthermore, as $\sigma^2_X/n = \operatorname{Var}(\bar{X}_n) \leq \vopt(\bar{X}_n)$. Due to this we note that
\begin{align*}
    \Ex[|\bar{X}_n-Z|]  
    &\leq \tilde{C} \dfrac{1}{\sqrt{n}} \sqrt{\frac{2}{\pi}} e^{-\mu^2n/2} + \mu \bigg(1-2\Phi\bigg( -\frac{\mu \sqrt{n}}{\sigma_X}\bigg) \bigg) + 2\tilde{C} \dfrac{1}{\sqrt{n}} \\[8pt]
    & \leq  \dfrac{1}{\sqrt{n}} \left[\tilde{C}\sqrt{\frac{2}{\pi}} e^{-\mu^2n/2} + 2\tilde{C} \right] + \mu \bigg(1-2\Phi\bigg( -\frac{\mu \sqrt{n}}{\sigma_X}\bigg) \bigg) 
\end{align*}

\newpage 
\section{Instability and Non-normality of Batched Bandits}\label{append-instab-non-norm}

\noindent In this section we shall discuss three batched bandit strategies discussed in \citet{zhang2020inference}. These algorithms where shown to have non-normal limits as proved in \citet{zhang2020inference}. Here we prove that they are also not stable. Hence, these algorithms show that if the stability condition is not satisfied by a bandit algorithm, then it may not satisfy a central limit theorem.

\noindent Let us first discuss the setup. Here we consider a two-armed, batched bandits with two epochs. The first epoch consists of $T/2$ runs followed by the second epoch.  Consider arm $1$, and let us define the first epoch by a sequence of binary actions 
$\{ \mathcal{A}_{1,i},\mathcal{A}_{2,i}\}_{i=1}^{T/2}$. Here for $k \in \{ 1,2 \}$, $\mathcal{A}_{k,i} = 1$ indicates that arm $1$ is pulled at step $i$ during the $k^{th}$ epoch. Consequently, we note that $\mathcal{A}_{1,i} = 0$ indicates that arm $2$ is pulled. Conditionally on the action, the observed reward is 
\begin{equation}
\Rew_i = 
\begin{cases} 
\Rew^{(1)}_{1,i}, & \text{if } \mathcal{A}_{1,i} = 1, \\
\Rew^{(2)}_{1,i}, & \text{if } \mathcal{A}_{1,i} = 0.
\end{cases}
\end{equation}

\noindent We assume that the actions are independent Bernoulli random variables:
\begin{equation}
\mathcal{A}_{1,i} \sim \operatorname{Ber}(p_1), \quad i = 1, \dots, T/2.
\end{equation}

\noindent Let $\mathcal{F}_i := \left\{\mathcal{A}_{1,1}, \Rew_1, \dots, \mathcal{A}_{1,i}, \Rew_i : i = 1, \dots, T \right\}$ denote the history of the first epoch. In the second epoch, we select actions 
\begin{equation}
\{\mathcal{A}_{2,i}\}_{i=1}^{T/2} \sim \operatorname{Ber}(p_2(T)),
\end{equation}
where the success probability $p_2(T)$ is adapted to the history of the first epoch. Hence, by changing the choice of $p_1$ and $p_2(T)$ we obtain different batched bandit strategies. Now as mentioned earlier we shall present three such strategies which were discussed in \cite{zhang2020inference}.

\subsubsection*{$\epsilon$-Greedy ETC}

\noindent Here we describe an $\epsilon$ greedy Explore then Commit (ETC) strategy for some fixed $\epsilon >0$.  Let $\alpha = T/2$ and define $\widehat{\mu}_{1,\alpha T}$ to be the mean rewards in the first epoch for arm $a$. Now, let $p_1 = 1/2$, and $p_2(T)$ is defined as follows:
\begin{equation}
p_2(T) = 
\begin{cases} 
1-\dfrac{\epsilon}{2}, & \text{if } \widehat{\mu}_{1,\Tfrac} > \widehat{\mu}_{2,\Tfrac}, \\[8pt]
\dfrac{\epsilon}{2}, & \text{otherwise }.
\end{cases}
\end{equation}

\noindent Therefore, for the first $T/2$ rounds either the first or the second arm is pulled with equal probability $1/2$. Then the arm which had the higher mean reward in the first epoch is drawn (independently) with higher probability in the second epoch.

\subsubsection*{ Thompson Sampling in batched setting}

\noindent Suppose $p_1 = 1/2$. For each arm $a$ set $n_{a,t}$ as the number of arm pulls at the end of first epoch. Similarly, let $\widehat{\mu}_{a,\frac{T}{2}}$ be the mean reward of arm $a$ at the end of first epoch. Assume we have $\beta_1,\beta_2$ as i.i.d priors $N(0,1)$. Furthermore, let $\tilde{\beta}_1, \ \tilde{\beta}_2$ be the sampled posterior means of arms $1$ and $2$ respectively. Finally, fix $\pi_{\min}$ and $\pi_{\max}$ such that $0 < \pi_{\min} \leq \pi_{\max} < 1$. Now we define $p_2(T)$ as follows:
\begin{equation}
p_2(T) = 
\begin{cases} 
\pi_{\max}, & \text{if } \prob(\tilde{\beta}_1 > \tilde{\beta}_2 | \mathcal{H}_{T/2}) \geq \pi_{\max}, \\[8pt]
\prob(\tilde{\beta}_1 > \tilde{\beta}_2 | \mathcal{H}_{T/2}), & \text{if } \pi_{\min} < \prob(\tilde{\beta}_1 > \tilde{\beta}_2 | \mathcal{H}_{T/2}) \leq \pi_{\max},\\[8pt]
\pi_{\min}, & \text{otherwise }.
\end{cases}
\end{equation}

\noindent Therefore, for the first $T/2$ rounds either the first or the second arm is pulled with equal probability $1/2$. Then the arm which has higher probability of posterior mean in the first epoch is drawn (independently) with higher probability in the second epoch.

\subsubsection*{ UCB in batched setting}

\noindent Here we describe a batch Upper Confidence Bound (UCB) strategy for some fixed $\pi_{\max} >1/2 $.  Let $\alpha = T/2$ and define the upper confidence bound for arm $a$ as follows:
\begin{equation}
     U_a \bigg(\Tfrac \bigg) := \widehat{\mu}_a(t) + \sqrt{\dfrac{\log \Tfrac}{n_{a,\Tfrac}}}
\end{equation}

\noindent Suppose $p_1 = 1/2$. Now we define $p_2(T)$ as follows:
\begin{equation}
p_2(T) = 
\begin{cases} 
\pi_{\max}, & \text{if }  U_1 \left(\Tfrac \right) > U_2 \left(\Tfrac \right), \\[8pt]
1-\pi_{\max}, & \text{otherwise }.
\end{cases}
\end{equation}

\noindent Therefore, for the first $T/2$ rounds each arm is pulled with equal probability $1/2$. Then the arm which had the highest UCB index in the first epoch is drawn (independently) with higher probability in the second epoch.  In Appendix C of \citet{zhang2020inference}, the authors show that for the aforementioned algorithms, $\sqrt{n_{a_1,T}(\mathcal{A})} ( \bar{\mu}_{a_1,T} - \mu_{a_1})$ converges to a non-normal limit. We state it in the form of a lemma below.
\begin{lemma}
Suppose we run either the $\epsilon$-greedy ETC algorithm or the UBC algorithm (in a batched setting). Then, we have:
    \begin{equation}
        \sqrt{n_{a_1,T}(\mathcal{A})} ( \hat{\mu}_{a_1,T} - \mu_{a_1}) \xrightarrow{d} Y
    \end{equation}
    where \\ 
    \begin{align*}
    &Y = \sqrt{\frac{1}{3-\epsilon}}\left(Z_1-\sqrt{2-\epsilon} Z_3\right) \mathbb{I}_{Z_1>Z_2}+\sqrt{\frac{1}{1+\epsilon}}\left(Z_1-\sqrt{\epsilon} Z_3\right) \mathbb{I}_{Z_1<Z_2} , & \text{ for ETC } \\[8pt]
    \end{align*}
    \begin{align*}
       &Y = \left(\sqrt{\frac{\frac{1}{2}}{\frac{1}{2}+\pi_{\text {max }}}} Z_1+\sqrt{\frac{\pi_{\text {max }}}{\frac{1}{2}+\pi_{\text {max }}}} Z_3\right) \mathbb{I}_{\left(Z_1>Z_2\right)}+\left(\sqrt{\frac{\frac{1}{2}}{\frac{3}{2}-\pi_{\text {max }}}} Z_1+\sqrt{\frac{1-\pi_{\text {max }}}{\frac{3}{2}-\pi_{\text {max }}}} Z_3\right) \mathbb{I}_{\left(Z_1<Z_2\right)}\\[8pt] & \text{for UCB, and  }\\[8pt]
        & Y = \sqrt{\dfrac{1/2+1-\pi_*}{1+2\pi_*}} \left(\sqrt{1/2} \ Z_1 + \sqrt{\pi_*}Z_3 \right) - \sqrt{\dfrac{1/2+\pi_*}{3-2\pi_*}} \left(\sqrt{1/2} \ Z_2 + \sqrt{1-\pi_*}Z_4 \right), \quad \text{for Thompson }  
    \end{align*}
    for $Z_1, Z_2, Z_3$ are i.i.d $N(0,1)$ random variables and $\pi_* \ \in \ (0,1)$.
\end{lemma}

\noindent We note that even though stability implies asymptotic normality, it has not been proved yet whether stability is a necessary condition. Consequently, non-normal limiting distribution does not imply that the algorithm is necessarily unstable. However, in the next lemma we show that these algorithms are not stable.
\begin{lemma}\label{lemma-batched-unstable}
    The batched $\epsilon$-greedy ETC algorithm, batched UCB algorithm and batched Thompson algorithm are unstable.
\end{lemma}

\noindent Lemma~\ref{lemma-batched-unstable} demonstrates that instability in batched bandit algorithms gives rise to non-normal limits. Since lemma~\ref{lemma-stab-norm} already establishes that stability is a sufficient condition for asymptotic normality, these examples highlight the central role of stability: its violation can naturally lead to non-normal asymptotic behavior.

\subsection*{Proof of Lemma~\ref{lemma-batched-unstable}}

\noindent Let us first prove it for the $\epsilon$-greedy ETC bandit strategy. We know that for the first $\lfloor T/2 \rfloor$ times, both the arms are pulled with equal probability. For arm $1$ we have:
\begin{align}\label{eqn-armpull-decom}
    n_{1,T} = \sum^{T/2}_{i = 1} \mathcal{A}_{1,i} + \sum^{T/2}_{j = 1} \mathcal{A}_{2,j}
\end{align}
where $\mathcal{A}_{1,i}$ are i.i.d $\operatorname{Ber}(p_1)$ and $\mathcal{A}_{2,i}$ are i.i.d $\operatorname{Ber}(p_2(T))$. We recall that,
\begin{align*}
    p_2(T)=
    \begin{cases}
       1-\dfrac{\epsilon}{2}, & \text{if } \widehat{\mu}_{1,\frac{T}{2}} >  \widehat{\mu}_{1,\frac{T}{2}} , \\[8pt]
\dfrac{\epsilon}{2}, & \text{otherwise }. 
    \end{cases}
\end{align*}

\noindent Now, we define two independent  triangular sequences of i.i.d random variables $(\mathcal{B}_{T,i})_{1 \leq i \leq T/2}$ and $(\mathcal{C}_{T,i})_{1 \leq i \leq T/2}$ such that $\mathcal{B}_{T,i} \sim \operatorname{Ber} \left(1-\dfrac{\epsilon}{2} \right)$ and $\mathcal{C}_{T,i} \sim \operatorname{Ber} \left(\dfrac{\epsilon}{2} \right)$. These random variables are \textit{independent} of the bandit strategy and act as potential outcomes for the random variables $\mathcal{A}_{2,i}$. This means the following
\begin{align}\label{eqn-armpull-epoch2}
    \mathcal{A}_{2,i} = 
    \begin{cases}
        \mathcal{B}_{T,i}, & \text{if } \widehat{\mu}_{1,\frac{T}{2}} >  \widehat{\mu}_{1,\frac{T}{2}} , \\[8pt]
        \mathcal{C}_{T,i}, & \text{otherwise }.
    \end{cases}
\end{align}

\noindent From equation~\eqref{eqn-armpull-decom} it follows that,
\begin{align*}\label{eqn-armpull-decom-sc}
    \dfrac{n_{1,T}}{T} = \dfrac{1}{2} \left( \dfrac{2}{T}\sum^{T/2}_{i = 1} \mathcal{A}_{1,i} \right) +\dfrac{1}{2} \left( \dfrac{2}{T} \sum^{T/2}_{j = 1} \mathcal{A}_{2,j} \right)
\end{align*}

\noindent Fix some arbitrary $\delta>0$. We claim the following:
\begin{align*}
   \lim_{T \rightarrow \infty} \prob \left(\bigg| \dfrac{n_{1,T}}{T} - \left( \dfrac{1}{4} + \dfrac{\epsilon}{2}\right)  \bigg| > \delta \right) = \dfrac{1}{2}
\end{align*}

\noindent Therefore, by applying Lemma~\ref{lemma-stab-seq} the above result implies instability of the algorithm. Now, by applying the WLLN,
\begin{align*}
     \dfrac{2}{T}\sum^{T/2}_{i = 1} \mathcal{A}_{1,i} \xrightarrow{\prob} \dfrac{1}{2}
\end{align*}

\noindent Hence, it follows that it is enough to prove
\begin{align*}
    \lim_{T \rightarrow \infty} \prob \left(\bigg| \dfrac{1}{2} \left( \dfrac{2}{T} \sum^{T/2}_{j = 1} \mathcal{A}_{2,j} \right) -  \dfrac{\epsilon}{2}  \bigg| > \dfrac{\delta}{2} \right) = \dfrac{1}{2}
\end{align*}

\noindent Equation~\eqref{eqn-armpull-epoch2} shows that as $T\rightarrow \infty$ the above holds if and only if $\mathcal{A}_{2,i} = \mathcal{C}_{T,i}$. We prove this rigorously below.
\begin{align*}
    &\prob \left(\bigg| \dfrac{1}{2} \left( \dfrac{2}{T} \sum^{T/2}_{j = 1} \mathcal{A}_{2,j} \right) -  \dfrac{\epsilon}{2}  \bigg| > \dfrac{\delta}{2} \right)\\[8pt]
    &=  \prob \left(  \widehat{\mu}_{2,\frac{T}{2}} >  \widehat{\mu}_{1,\frac{T}{2}}, \ \bigg| \dfrac{1}{2} \left( \dfrac{2}{T} \sum^{T/2}_{j = 1} \mathcal{C}_{T,i} \right) -  \dfrac{\epsilon}{2}  \bigg| > \dfrac{\delta}{2} \right) + o(1)\\[8pt]
    &= \prob \left(  \widehat{\mu}_{2,\frac{T}{2}} >  \widehat{\mu}_{1,\frac{T}{2}} \right) \times \prob \left(  \bigg| \dfrac{1}{2} \left( \dfrac{2}{T} \sum^{T/2}_{j = 1} \mathcal{C}_{T,i} \right) -  \dfrac{\epsilon}{2}  \bigg| > \dfrac{\delta}{2} \right)+ o(1)\\[8pt]
    & = \dfrac{1}{2} \times \prob \left(  \bigg| \dfrac{1}{2} \left( \dfrac{2}{T} \sum^{T/2}_{j = 1} \mathcal{C}_{T,i} \right) -  \dfrac{\epsilon}{2}  \bigg| > \dfrac{\delta}{2} \right) + o(1)
\end{align*}

\noindent The last equality holds true because $\mathcal{C}_{T,i}$ are independent of the bandit strategy and the reward distributions for both the arms are identical (which are chosen with equal probability). This proves that the $\epsilon$-greedy algorithm is unstable.

\noindent Now, the proof of instability of the batched UCB is analogous to that of the ETC. By replacing $1-\pi_{\max}$ in place of $\epsilon/2$ we observe that we only need to prove,
\begin{align*}
   \lim_{T \rightarrow \infty} \prob \left(\bigg| \dfrac{n_{1,T}}{T} - \left( \dfrac{1}{4} + 1-\pi_{\max}\right)  \bigg| > \delta \right) > 0 
\end{align*}

\noindent A careful look at the proof of the ETC indicates that our objective is to prove,
\begin{align*}
    \lim_{T \rightarrow \infty }\prob \left(U_2 \left(\dfrac{T}{2}\right) > U_1 \left(\dfrac{T}{2}\right) \right) >0
\end{align*}

\noindent If possible, suppose stability holds. Then for any $\epsilon >0$, by applying lemma~\ref{lemma-stab-seq} we know that $\prob(n_{1,T/2}/n_{2,T/2} \in (1-\epsilon,1+\epsilon))$ tends to $1$ as $T \rightarrow \infty $. By using this result we have the following chain of equalities: 
\begin{align*}
    &\lim_{T \rightarrow \infty }\prob \left(U_2 \left(\dfrac{T}{2}\right) > U_1 \left(\dfrac{T}{2}\right) \right)\\[8pt]
    &= \lim_{T \rightarrow \infty }\prob \left(\widehat{\mu}_{2,\frac{T}{2}} + \sqrt{\dfrac{2 \log 1/\gamma}{n_{2,T}}} > \widehat{\mu}_{1,\frac{T}{2}} + \sqrt{\dfrac{2 \log 1/\gamma}{n_{1,T}}}\right)\\[8pt]
    & =\lim_{T \rightarrow \infty }\prob \left( \sqrt{n_{2,T}} \ \widehat{\mu}_{2,\frac{T}{2}} + \sqrt{2 \log 1/\gamma} > \dfrac{\sqrt{n_{2,T}}}{\sqrt{n_{1,T}}} \ \sqrt{n_{2,T}} \ \widehat{\mu}_{1,\frac{T}{2}} + \dfrac{\sqrt{n_{2,T}}}{\sqrt{n_{1,T}}} \ \sqrt{2 \log 1/\gamma}\right)\\[8pt]
    &  =\lim_{T \rightarrow \infty }\prob \left( \sqrt{n_{2,T}} \ \widehat{\mu}_{2,\frac{T}{2}} - \dfrac{\sqrt{n_{2,T}}}{\sqrt{n_{1,T}}} \ \sqrt{n_{2,T}} \ \widehat{\mu}_{1,\frac{T}{2}}   >   \dfrac{\sqrt{n_{2,T}}}{\sqrt{n_{1,T}}} \ \sqrt{2 \log 1/\gamma} - \sqrt{2 \log 1/\gamma}\right)
\end{align*}

\noindent From lemma~\ref{lemma-stab-norm} it follows that $\sqrt{n_{2,T}} \ \widehat{\mu}_{2,\frac{T}{2}} - \sqrt{1-\epsilon} \ \sqrt{n_{2,T}} \ \widehat{\mu}_{1,\frac{T}{2}} \xrightarrow{d} N(0,2-\epsilon)$. Therefore, from the last equality we have,

\begin{align*}
    \lim_{T \rightarrow \infty }\prob \left(U_2 \left(\dfrac{T}{2}\right) > U_1 \left(\dfrac{T}{2}\right) \right) = \Phi \left(-\dfrac{\sqrt{1+\epsilon} \ \sqrt{2 \log 1/\gamma} - \sqrt{2 \log 1/\gamma}}{\sqrt{2-\epsilon}} \right) >0
\end{align*}

\noindent This proves that the batched UCB algorithm is also unstable. Now, we shall prove that the batch Thompson sampling algorithm is also unstable. To prove this, we shall invoke proposition $1$ in Appendix C of paper (\citet{zhang2020inference}):
\begin{lemma}
    For identical two armed batch Thompson sampling, we have:
    \begin{equation*}
        \prob(\tilde{\beta}_1 > \tilde{\beta}_2 | \mathcal{H}_{T/2}) \xrightarrow{d} U(0,1) \ \ \text{as} \ \ T \rightarrow \infty
    \end{equation*}
\end{lemma}

\noindent As a consequence, we have:
\begin{align*}
    \prob (\prob(\tilde{\beta}_1 > \tilde{\beta}_2 | \mathcal{H}_{T/2}) \geq \pi_{\max}) \rightarrow 1-\pi_{\max}
\end{align*}

\noindent We observe that $(\mathcal{A}_{2,i})_{1 \leq i \leq T/2}$ is a mixture of i.i.d Bernoulli random variables indexed by the random variable $p_2(T)$. Therefore, there exists a triangular sequence of i.i.d random variables  $(\mathcal{B}_{T,i})_{1 \leq i \leq T/2}$ (independent of the bandit stategy) such that $\mathcal{B}_{T,i} \sim \operatorname{Ber}(\pi_{\max})$ and, 
\begin{align*}
    \mathcal{A}_{2,i} = \mathcal{B}_{T,i} \ \ \ \text{, if}\ \  p_2(T) = \pi_{\max}
\end{align*}

\noindent Now, by applying similar arguments, we observe the following chain of inequalities:
\begin{align*}
    &\lim_{T \rightarrow \infty} \prob \left(\bigg| \dfrac{1}{2} \left( \dfrac{2}{T} \sum^{T/2}_{j = 1} \mathcal{A}_{2,j} \right) -  \pi_{\max}  \bigg| > \dfrac{\delta}{2} \right)\\[8pt]
    & =  \prob \left( \prob(\tilde{\beta}_1 > \tilde{\beta}_2 | \mathcal{H}_{T/2}) \geq \pi_{\max}, \ \bigg| \dfrac{1}{2} \left( \dfrac{2}{T} \sum^{T/2}_{j = 1} \mathcal{B}_{T,j} \right) -  \dfrac{\epsilon}{2}  \bigg| > \dfrac{\delta}{2} \right)\\[8pt]
    & = 1-\pi_{\max} >0
\end{align*}

\noindent This concludes the proof that the batched Thompson Sampling strategy is unstable.

\newpage

\section{ Proofs of Auxiliary Results} \label{append-aux-lemma}

\noindent In this section, we shall first outline the proof of instability for Anytime-MOSS, followed by the proof for KL-UCB-SWITCH. This is followed by proofs of Lemmas~\ref{lemma-stab-seq},\ref{lemma-mc-indep},\ref{lemma-M-alpha} and~\ref{lemma-mc-couple}

\subsection*{Instability of Anytime-MOSS}~\label{proof:anymoss}

\noindent The proof of instability of Anytime MOSS follows along similar lines of argument presented in Section~\ref{sec:Instability-proof} with some slight modification. Let $U^A_a(t),U^V_a(t)$ be the UCB indices of Anytime MOSS and Vanilla Moss (see Table~\ref{tab:upd-rule}). Note that as $t \leq T \leq TK$, the following holds:
\begin{align}\label{eqn-any-van}
   U^A_a(t)
   =\widehat{\mu}_a(t) + \sqrt{\dfrac{\max\{0,\; \log\bigl(\tfrac{t}{K n_{a,t}}\bigr)\}}{n_{a,t}}} \leq 
   \widehat{\mu}_a(t) + \sqrt{\dfrac{\log \bigl(\tfrac{T}{n_{a,t}}\bigr)}{n_{a,t}}}
   = U^V_a(t)
\end{align}

\noindent Now let us consider event $\mathcal{E}^V_0(T)$ for defined as follows
\begin{equation}\label{eqn-Event-AV}
    \mathcal{E}^{A,V}_0(T) := \left\{U^V_a \left(\frac{T}{2K} \right)<\bound \ \forall a \in \{2,\ldots,K\},\ \min_{\frac{T}{2K} \leq t \leq T } U^A_1(t)> \bound\right\}.
\end{equation}

\noindent We note that $U^A_a(T/2K) \leq U^V_a(T/2K)$ and on the event 
$\{ U^V_a(T/2K)<\bound \;\;   \forall a\ge 2\}$ $\cap \{ U^A_1(T/2K)>\bound \}$, we have 
arm $1$ is pulled at time  $T/(2K)+1$. Consequently, we have for all arms $a\geq 2$,
\begin{align*}
    n_{a,\frac{T}{2K}+1}(\mathcal{A}) = n_{a,\frac{T}{2K}}(\mathcal{A}), \qquad  \text{and} 
    \qquad
    \widehat{\mu}_{a,\frac{T}{2K}+1} = \widehat{\mu}_{a,\frac{T}{2K}}.
\end{align*}

\noindent Hence, it follows from Assumption~\ref{assump:H-G} that
\begin{align*}
    U_a^V \Big(\frac{T}{2K}+1\Big) \leq U_a^V \Big(\frac{T}{2K}\Big) < \bound.
\end{align*}

\noindent By applying this argument iteratively it follows that on event $\mathcal{E}^{A,V}_0(T)$ arm $1$ is pulled at every step from $T/(2K)$ to $T$. Hence,
\begin{align}
    \mathcal{E}^{A,V}_0(T) \subset \left\{\frac{n_{1,T}(\mathcal{A})}{T} \geq \frac{K+1}{2K},\ 
    \frac{n_{a,T}(\mathcal{A})}{T} \leq \frac{1}{2K} \ \forall a \ge 2 \right\}.
\end{align}
where algorithm $\Algo$ is Anytime-MOSS. Therefore, as discussed in Section~\ref{sec:Instability-proof}, once we prove 
\begin{align*}
    \liminf_{T\rightarrow\infty}\prob(\mathcal{E}^{A,V}_0(T)) > 0
\end{align*}
then instability of Anytime MOSS will follow. Furthermore, an inspection of the algorithm shows that it satisfies Assumption~\ref{assump:ucb-bound}. Now suppose that Anytime MOSS is \emph{stable}. By Lemma~\ref{lemma-stab-seq} we have
\begin{align*}
    \frac{n_{a,T/2K}}{T/2K^2} \xrightarrow{\prob} 1
\end{align*}
Consequently, we have $\prob(\E)\to 1$ as $T\to\infty$, and for any sequence of events $(A_T)_{T\ge1}$,
\begin{align}
    \liminf_{T\to\infty}\prob(A_T)
    = 
    \liminf_{T\to\infty}\prob(A_T \cap \E).
\end{align}

\noindent In particular, for the choice of $A_T = \mathcal{E}^{A,V}_0(T)$ we obtain,
\begin{align}
    \liminf_{T\to\infty}\prob(\mathcal{E}^{A,V}_0(T))
    =
    \liminf_{T\to\infty}\prob(\mathcal{E}^{A,V}_0(T) \cap \E).
\end{align}

\noindent On the event $\mathcal{E}^{A,V}_0(T) \cap \E$, Assumption~\ref{assump:ucb-bound} implies that the upper confidence bounds $U^A_a(t)$ for all $a\in[K]$ lie between $\widetilde{U}^A_{a,\beta_1}(t)$ and $\widetilde{U}^A_{a,\beta_2}(t)$ where,
\begin{align*}
    \widetilde{U}^A_{a,\beta}(t) = \widehat{\mu}_{a,t} + \frac{\beta}{\sqrt{n_{a,t}}}
\end{align*}
\noindent Moreover, on $\E$, there exist constants $\gamma_1,\gamma_2$ (depending on $\beta_1,\beta_2$) such that for each arm $a \in [K]$,
\begin{align}
\widehat{\mu}_{a,t} + \frac{\gamma_1}{\sqrt{t}}
        \leq \widetilde{U}^A_{a,\beta_1}(t)
        \leq U^A_a(t)
        \leq \widetilde{U}^A_{a,\beta_2}(t)
        \leq \widehat{\mu}_{a,t} + \frac{\gamma_2}{\sqrt{t}}.    
\end{align}

\noindent As done in Section~\ref{sec:Instability-proof}, we define the \emph{pseudo upper confidence bound} below
\begin{align*}
    \W^A_{a,\gamma}(t) := \widehat{\mu}_{a,t} + \frac{\gamma}{\sqrt{t}}.
\end{align*}
We note that the rest of the proof is exactly identical to Section~\ref{sec:Instability-proof} and hence we conclude our proof.

\subsection*{Instability of Anytime KL-UCB-SWITCH}

\noindent The proof of instability of Anytime-KL-UCB-SWITCH borrows ideas from the proof of Anytime MOSS. Let $U_a(t)$ be the UCB for arm $a \in [K]$. As $t \leq T \leq TK$, it follows that
\begin{align}
   U_a(t) \leq 
\begin{cases}
\displaystyle 
\sup\Bigl\{ q \in [0,1] :\,
KL\!\left(\widehat{\mu}_a(t), q\right)
\le 
\dfrac{\phi\!\left(\tfrac{T}{ n_{a,t}}\right)}{n_{a,t}}
\Bigr\},
& \text{if } n_{a,t} \le f(t,K), \\[10pt]
\displaystyle 
\widehat{\mu}_a(t) 
+ \sqrt{\dfrac{1}{2 n_{a,t}}\, \phi\!\left(\tfrac{T}{ n_{a,t}}\right)},
& \text{if } n_{a,t} > f(t,K),
\end{cases} 
\end{align}
 where \(\phi(x) := \log^{+}(x)\bigl(1 + (\log^{+} x)^2\bigr)\) and $f(t,K) = [(t/K)^{1/5}]$.

\noindent Let us define two UCB indices for arm $a$ as follows
\begin{align*}
    U^{(1)}_a(t) := \widehat{\mu}_a(t) + \sqrt{\dfrac{1}{ n_{a,t}}\, \phi\!\left(\tfrac{T}{ n_{a,t}}\right)},
    \quad 
    U^{(2)}_a(t) := \widehat{\mu}_a(t) + \sqrt{\dfrac{1}{2 n_{a,t}}\, \phi\!\left(\tfrac{T}{ n_{a,t}}\right)}
\end{align*}

\noindent Then due to inequality~\ref{eqn-pinsker}, the UCB index $U_a(t) \leq U^{(1)}_a(t)$. Motivated by this observation, we define the following event
\begin{equation}\label{eqn-Event-one}
    \mathcal{E}^{(1)}_0(T) := \left\{U^{(1)}_a \left(\frac{T}{2K} \right)<\bound \ \forall a \in \{2,\ldots,K\},\ \min_{\frac{T}{2K} \leq t \leq T } U_1(t)> \bound\right\}.
\end{equation}

\noindent We note the event $\mathcal{E}^{A,V}_0(T)$ in equation~\eqref{eqn-Event-AV} is similar to $\mathcal{E}^{(1)}_0(T)$ in equation~\eqref{eqn-Event-one}. Now analogous argument as the one provided in the proof of Anytime MOSS completes the proof.

\subsection*{Proof of Lemma~\ref{lemma-stab-seq}}

\noindent We begin the proof by observing that since the reward distributions of both the arms are identical, the following is true due to symmetry.
\begin{align*}
    n_{1,T}(\Algo)  \overset{\text{d}}{=} n_{a,T}(\Algo) \quad \text{for all $a\in \{ 2,\ldots,K\}$}
\end{align*}

\noindent Now, suppose stability holds. This implies that there exists real sequences $(n^{*}_{a,T}(\Algo))$ such that :
\begin{align*}
    \frac{n_{a,T} (\Algo)) }{n^{*}_{a,T} (\Algo)} \xrightarrow{\prob} 1 \quad \text{as}\quad T \rightarrow \infty 
\end{align*}

\noindent This implies that for any arm $a\in \{2,\ldots,K \}$,
\begin{align}\label{eqn-stab-pd}
    \frac{n_{1,T} (\Algo) }{n^{*}_{1,T} (\Algo)} \  \overset{\text{d}}{=} \ \frac{n_{a,T} (\Algo) }{n^{*}_{1,T} (\Algo)} 
\end{align}

\noindent We note that stability implies that $n_{1,T} (\Algo)/n^\star_{1,T} (\Algo)$ converges to $1$ in probability, which further implies in distribution convergence. Therefore from equation~\eqref{eqn-stab-pd} it follows that 
\begin{align}\label{eqn-stab-pd1}
    \frac{n_{a,T} (\Algo) }{n^{*}_{1,T} (\Algo)} \xrightarrow{d} 1 \implies 
    \frac{n_{a,T} (\Algo) }{n^{*}_{1,T} (\Algo)} \xrightarrow{\prob} 1
\end{align}

\noindent Again, from stability we know that for any arm $a\in[K]$,
\begin{align}\label{eqn-stab-pd2}
    \frac{n_{a,T} (\Algo) }{n^{*}_{a,T} (\Algo)} \xrightarrow{\prob} 1
\end{align}

\noindent Therefore from equations~\eqref{eqn-stab-pd1} and~\eqref{eqn-stab-pd2} we obtain,
\begin{align*}
    \frac{n^\star_{1,T} (\Algo) }{n^{*}_{a,T} (\Algo)} \xrightarrow{} 1
\end{align*}

\noindent This implies that we can assume that for each arm $a$, $n^{*}_{1,T}(\Algo) = n^{*}_{a,T}(\Algo)$. This leads us to the following observation.
\begin{align*}
    K =\lim_{T \rightarrow \infty}\sum^{K}_{a=1} \frac{n_{a,T} (\Algo)) }{n^{*}_{a,T} (\Algo)} = \lim_{T \rightarrow \infty} \sum^{K}_{a=1} \frac{n_{a,T} (\Algo)) }{n^{*}_{1,T} (\Algo)} = \lim_{T \rightarrow \infty} \frac{T}{n^{*}_{1,T} (\Algo)}
\end{align*}

\noindent Therefore, as $T\rightarrow \infty$, $n^{*}_{1,T} (\Algo)/T$ converges to $T/K$.

\subsection*{Proof of Lemma~\ref{lemma-MC}}

\noindent We shall prove this result by proving a similar result for Let us a sub-vector $\textbf{\Ztwo}_t$ defined below.
\begin{align*}\label{defn-L}
    \textbf{\Ztwo}_t := \begin{bmatrix}
        \widehat{\mu}_{1,t}  \\ \widehat{\mu}_{2,t}\\ n_{1,t} \\ n_{2,t}
    \end{bmatrix}
\end{align*}

\noindent Therefore, ignoring the order in which the elements appear in the definition of $\textbf{Z}_t$ it follows that:
\begin{equation*}
    \textbf{Z}_t = \begin{bmatrix}
        \W_{1,\gamma_1}(t) \\ \W_{2,\gamma_2}(t) \\ \textbf{\Ztwo}_t
    \end{bmatrix}
\end{equation*}
Now,  we recall that,
\begin{equation*}
    \W_{a,\gamma}(t) := \widehat{\mu}_{a,t} + \gamma\sqrt{\dfrac{\log(T/2n_{a,t})}{n_{a,t}}} 
\end{equation*}

\noindent Now, according to our algorithm structure, when we have two arms $a_1$ and $a_2$ we have the following:
\begin{align*}
    &n_{a_1,t+1}= n_{a_1,t}+1\left\{U_{a_1}(t)>U_{a_2}(t)\right\},\\[8pt]
    & \widehat{\mu}_{a_1,t+1} = \frac{n_{a,t}}{n_{a_1,t+1}}\widehat{\mu}_{a_1,t} + \dfrac{\Rew_{t+1}}{n_{a_1,t+1}}1\left\{U_{a_1}(t)>U_{a_2}(t)\right\}.
\end{align*}
\noindent Now, let $\Rew_{a,t} := \mu_a + \xi_{a,t}$ be the potential reward process for arm $a$. We note that for each $a$, process $(X_{a,t})_{t \geq 1}$ is independent of the algorithm. Furthermore, if arm $a$ is chosen at time $t+1$ then from the definitions it follows that $\Rew_{t+1} = \Rew_{a,n_{a,t} +1}$. We claim that $\Rew_{a_1,n_{a_1,t+1}}$ is independent of the history till time $t$. This may seem counter-intuitive as from the definition it looks like it is dependent on $n_{a,t+1}$. To see this we observe that,
\begin{align*}
   \Rew_{a,n_{a,t+1}} = \sum^{\infty}_{i = 1} \Rew_{a,i} 1\left\{ n_{a,t+1} = i \right\} 
\end{align*}

\noindent This implies that,
\begin{align*}
    &\prob(\Rew_{a,n_{a,t+1}} \in B \  | n_{a,t+1} = k) = \prob(\Rew_{a,k} \in B \  | n_{a,t+1} = k) = \prob(\Rew_{a,k} \in B )= \prob(\Rew_{a,1} \in B )
\end{align*}

\noindent Furthermore,
\begin{align}\label{eqn-indep}
   \notag
   &\prob(\Rew_{a,n_{a,t+1}} \in B )\\[8pt] \notag
   & = \sum^{\infty}_{i = 1} \prob(\Rew_{a,n_{a,t+1}} \in B \  | n_{a,t+1} = i) \times \prob(n_{a,t+1} = i)\\[8pt] \notag
   & = \sum^{\infty}_{i = 1}  \prob(\Rew_{a,1} \in B ) \times \prob(n_{a,t+1} = i)\\[8pt]
   & = \prob(\Rew_{a,1} \in B )
\end{align}

\noindent Hence, for any measurable subset $B$ and $k$ we have:
\begin{align*}
    \prob(\Rew_{a,n_{a,t+1}} \in B \  | n_{a,t+1} = k) = \prob(\Rew_{a_1,1} \in B )
\end{align*}

\noindent Therefore, we have proved that $\Rew_{a_1,n_{a_1,t+1}}$ is independent of $n_{a,t+1}$. Consequently, we can rewrite the adaptive mean of arm $a_1$ as follows:
\begin{align}\label{eqn-muhat-det}
     \widehat{\mu}_{a_1,t+1} = f(\widehat{\mu}_{a_1,t}, n_{a_1,t}, U_{a_1}(t),\Rew_{a_1,n_{a_1,t}})
\end{align}
where $f$ is a deterministic measurable function. Furthermore,we also assume that the UCB indices $U_{a_1}(t)$ are also deterministic functions of the sample means and arm pulls (Assumption~\ref{assump:UCB-func}) as follows:
\begin{align} \label{eqn-UCB-det}
    U_{a_1}(t) = h_{a_1}(\widehat{\mu}_{1,t}, \widehat{\mu}_{2,t}, n_{1,t}, n_{2,t})
\end{align}

\noindent Therefore, from equations~\eqref{eqn-muhat-det} and~\eqref{eqn-UCB-det}  it follows that there exists a real function $\psi:\real^{2}\rightarrow\real$ such that:
\begin{align*}
   \textbf{Z}_t = \begin{bmatrix}
        \psi(\textbf{\Ztwo}_t) \\ \textbf{\Ztwo}_t
    \end{bmatrix} 
\end{align*}

\noindent Now, consider the following lemma.
\begin{lemma}\label{lemma-mc-couple}
    Assume that $(Q_t)_{t \geq 1}$ is a  finite dimensional discrete time Markov chain of dimension $d$. Also, suppose that $\psi$ is any real valued function such that $\psi:\real^{d} \rightarrow \real^k$, where $k \in \mathrm{N}$. Define a new process $M_t$ as follows:
    \begin{align*}
   M_t = \begin{bmatrix}
        \psi(Q_t) \\ Q_t
    \end{bmatrix} 
    \end{align*}
    Then the process $(M_t)_{t \geq 1}$ is also a Markov chain.
\end{lemma}

\noindent Therefore, if we can show that $(\textbf{\Ztwo}_t)_{t \geq 1}$ is a Markov chain we are done. We note that from equations~\eqref{eqn-UCB-det} and \eqref{eqn-muhat-det} it follows that there exists a real measurable function $\eta$ such that,
\begin{equation*}
    \textbf{L}_{t+1} = \eta(\textbf{L}_{t}, \Rew_{1,t}, \Rew_{2,t})
\end{equation*}
where $\Rew_{1,t}, \Rew_{2,t}$ are the respective reward processes independent of $\textbf{L}_t$. Consequently we deduce that, 
\begin{align*}
    &\prob(\textbf{L}_{t+1} = \textbf{s}_{t+1} \mid \textbf{L}_m = \textbf{s}_m, \; m = 1,2,\ldots,t) \\[8pt]
    &= \prob(\eta(\textbf{L}_{t}, \Rew_{1,t}, \Rew_{2,t}) = \textbf{s}_{t+1} \mid \textbf{L}_m = \textbf{s}_m, \; m = 1,2,\ldots,t) \\[8pt]
    &= \prob(\eta(\textbf{s}_t, \Rew_{1,t}, \Rew_{2,t}) = \textbf{s}_{t+1} \mid \textbf{L}_m = \textbf{s}_m, \; m = 1,2,\ldots,t) \\[8pt]
    &= \prob(\eta(\textbf{s}_t, \Rew_{1,t}, \Rew_{2,t}) = \textbf{s}_{t+1}) \\[8pt]
    &= \prob(\eta(\textbf{s}_t, \Rew_{1,t}, \Rew_{2,t}) = \textbf{s}_{t+1} \mid \textbf{L}_t = \textbf{s}_t) \\[8pt]
    &= \prob(\eta(\textbf{L}_t, \Rew_{1,t}, \Rew_{2,t}) = \textbf{s}_{t+1} \mid \textbf{L}_t = \textbf{s}_t) \\[8pt]
    &= \prob(\textbf{L}_{t+1} = \textbf{s}_{t+1} \mid \textbf{L}_t = \textbf{s}_t)
\end{align*}

\noindent The chain of equalities above imply that $(\textbf{L}_{t+1})_{t \geq 1}$ is a Markov chain and hence, we are done.

\subsection*{ Proof of Lemma~\ref{lemma-mc-indep}}

\noindent We know that $(\textbf{Q}_t)_{t \geq 1}$ is a finite - dimensional, discrete time Markov Chain. Let $A,B$ and $C$ be arbitrary sets comprising of states in the state space. We are interested in the probability of the event $\left\{ \textbf{Q}_{t-1} \in A, \textbf{Q}_{t+1} \in C \right\}$, \textit{given} that event $\left\{ \textbf{Q}_{t} \in B \right\}$ holds. Now, the following chain of equalities follow :
\begin{align*}
    &\prob(\textbf{Q}_{t-1} \in A, \textbf{Q}_{t+1} \in C | \textbf{Q}_{t} \in B)\\[8pt]
    & = \dfrac{\prob(\textbf{Q}_{t-1} \in A, \textbf{Q}_{t} \in B, \textbf{Q}_{t+1} \in C )}{\prob( \textbf{Q}_{t} \in B)}\\[8pt]
    & = \prob(\textbf{Q}_{t+1} \in C| \textbf{Q}_{t} \in B, \textbf{Q}_{t-1} \in A) \bigg[\dfrac{\prob(\textbf{Q}_{t} \in B, \textbf{Q}_{t-1} \in A)}{\prob( \textbf{Q}_{t} \in B)}  \bigg]\\[8pt]
    & = \prob(\textbf{Q}_{t+1} \in C| \textbf{Q}_{t} \in B) \times \prob( \textbf{Q}_{t-1} \in A| \textbf{Q}_{t} \in B)
\end{align*}

\noindent The last equality follows from the definition of Markov Chain due to the reason highlighted below.
\begin{align*}
    &\prob(\textbf{Q}_{t+1} \in C| \textbf{Q}_{t} \in B, \textbf{Q}_{t-1} \in A)\\[8pt]
    &=\dfrac{\prob(\textbf{Q}_{t+1} \in C, \textbf{Q}_{t} \in B, \textbf{Q}_{t-1} \in A)}{\prob( \textbf{Q}_{t} \in B, \textbf{Q}_{t-1} \in A)}\\[8pt]
    &= \dfrac{\prob(\textbf{Q}_{t+1} \in C, \textbf{Q}_{t} \in B, \textbf{Q}_{t-1} \in A,\textbf{Q}_{k} \in \real, k = 1,\ldots,t-2)}{\prob( \textbf{Q}_{t} \in B, \textbf{Q}_{t-1} \in A)}\\[8pt]
    &= \dfrac{\prob(\textbf{Q}_{t+1} \in C| \textbf{Q}_{t} \in B) \times \prob(\textbf{Q}_{t} \in B, \textbf{Q}_{t-1} \in A,\textbf{Q}_{k} \in \real, k = 1,\ldots,t-2)}{\prob( \textbf{Q}_{t} \in B, \textbf{Q}_{t-1} \in A)}\\[8pt]
    &= \dfrac{\prob(\textbf{Q}_{t+1} \in C| \textbf{Q}_{t} \in B) \times \prob( \textbf{Q}_{t} \in B, \textbf{Q}_{t-1} \in A)}{\prob( \textbf{Q}_{t} \in B, \textbf{Q}_{t-1} \in A)}\\[8pt]
    &=\prob(\textbf{Q}_{t+1} \in C| \textbf{Q}_{t} \in B)
\end{align*}

\noindent This proves our result when $m=1$. The proof for the general case follows iteratively.

\subsection*{Proof of Lemma~\ref{lemma-M-alpha}}

\noindent Define $M(\alpha T,\beta T):=\max_{\alpha T \leq t \leq \beta T} Q_{t}$. We wish to prove that $\left\{M(\alpha T,\beta T) \geq \lambda\right\} \bigcap \left\{\tau \leq k\right\}$ belongs in the sigma field $\Fil_k$, for each natural number k. Note that $M(r_1,r_2):=\text{max}_{t \in [r_1,r_2 ]} (-X^{*}_{1,t})$ and $\Fil_{1,t} := \sigma(\xi_{1,1},\ldots,\xi_{1,t})$. Now, recall that the definition of $\tau$ is as follows :
\begin{equation*}
    \tau := \text{min} \bigg(\text{inf} \left\{k \geq \alpha T | M(\alpha T, k) \geq \lambda  \right\}, \beta T \bigg)
\end{equation*}

\noindent Fix  an arbitrary natural number $k$. We split the proof in the following cases:
\begin{enumerate}
    \item Suppose $k <\alpha T$. In this case, $\left\{\tau \leq k\right\} = \phi$. Therefore, $\left\{M(\alpha T,\beta T) \geq \lambda\right\} \bigcap \left\{\tau \leq k\right\} = \phi$ which belongs in $\Fil_k$.
    \item Suppose $k > \beta T$. Then $\left\{\tau \leq k\right\} = \Omega$ and hence $\left\{M(\alpha T,\beta T) \geq \lambda\right\} \bigcap \left\{\tau \leq k\right\} = \left\{M(\alpha T,\beta T) \geq \lambda\right\}$ which belongs in $\Fil_{\beta T}$ which lies in $\Fil_k$ as $k > \beta T$.
    \item Suppose $\alpha T \leq k \leq \beta T$. In this case, we note that $\left\{\tau \leq k\right\} = \left\{M(\alpha T,k) \geq \lambda\right\}$ and vice-versa. As a consequence, $\left\{\tau \leq k\right\} = \left\{M(\alpha T,k) \geq \lambda\right\} \subseteq \left\{M(\alpha T,\beta T) \geq \lambda\right\}$. Therefore, $\left\{M(\alpha T,\beta T) \geq \lambda\right\} \bigcap \left\{\tau \leq k\right\} = \left\{M(\alpha T,k) \geq \lambda\right\}$ which depends only on random variables $\xi_{1,\alpha T}, \ldots, \xi_{1,k}$ and hence belongs in $\Fil_k$
\end{enumerate}

\noindent Therefore, we conclude that for any $k \in \mathcal{N}$, $\left\{M(\alpha T,\beta T) \geq \lambda\right\} \bigcap \left\{\tau \leq k\right\}$ belongs in the sigma field $\Fil_k$.

\subsection*{Proof of Lemma~\ref{lemma-mc-couple}}

\noindent We assume that $Q_t$ is a  $d$-dimensional discrete time Markov chain. We recall that,
 \begin{align*}
   M_t = \begin{bmatrix}
        \psi(Q_t) \\ Q_t
    \end{bmatrix} 
    \end{align*}
 where $\psi$ is a real-valued function. Fix an event $A$ in the extended state space of $(M_t)_{t \geq 1}$. Let $\textbf{v}$ be an arbitrary element of $A$ such that,
\begin{align*}
    \textbf{v} = \begin{bmatrix}
        \textbf{v}_1\\ \textbf{v}_2
    \end{bmatrix}
\end{align*}
where $\textbf{v}_1,\textbf{v}_2$ are $d$-dimensional and $k$ dimensional vectors respectively. This implies that,
\begin{align*}
    M_t = \textbf{v} \ \ \text{if and only if } \ \ \psi(Q_t) = \textbf{v}_1 \ \text{,} \ Q_t = \textbf{v}_2
\end{align*}

\noindent This implies that,
\begin{align*}
    \left\{\begin{bmatrix}
        \psi(Q_t) \\ Q_t
    \end{bmatrix} 
    =  \begin{bmatrix}
        \textbf{v}_1\\ \textbf{v}_2
    \end{bmatrix} \right\}
    =
    \begin{cases}
        \left\{ Q_t = \textbf{v}_2 \right\}, \ \ \text{if } \textbf{v}_2 \in \psi^{-1}(\textbf{v}_1)\\[8pt]
        \phi , \ \ \text{otherwise}
    \end{cases}
\end{align*}

\noindent Motivated by this observation, for set $A$ we define set $B_A$ as follows:
\begin{align*}
    B_A := \left\{ \textbf{v}_2: \begin{bmatrix}
        \textbf{v}_1\\ \textbf{v}_2
    \end{bmatrix} \ \in A \ \ \text{and,} \ \textbf{v}_2 \in \psi^{-1}(\textbf{v}_1) \right\}
\end{align*}
\noindent Therefore, 
\begin{align*}
  \left\{\begin{bmatrix}
        \psi(Q_t) \\ Q_t
    \end{bmatrix} 
    \in A \right\}
    =
    \left\{ Q_t \in B_A \right\}
\end{align*}

\noindent Now, let $A_1,A_2,\ldots,A_{t+1}$ be arbitrary sets in the extended state space. Now, suppose for some $i$, $M_i \in A_i$. This implies that there exists sets $B_{A_i}$ such that:
\begin{align*}
    \left\{ M_i \in A_i \right\} = \left\{ Q_i \in B_{A_i} \right\} 
\end{align*}

\noindent As $Q_t$ is a Markov chain, from the equation above we have:
\begin{align*}
    \prob \bigg( M_{t+1} \in A_{t+1} \bigg| M_{m} \in A_{m}, m = 1,2,\ldots,t \bigg)
    & = \prob \bigg( Q_{t+1} \in B_{A_{t+1}} \bigg| Q_{m} \in B_{A_m} \ , m = 1,2,\ldots,t \bigg)\\[8pt]
    & = \prob \bigg( Q_{t+1} \in B_{A_{t+1}} \bigg| Q_{t} \in B_{A_{t}} \bigg)\\[8pt]
    & = \prob \bigg( M_{t+1} \in A_{t+1} \bigg| M_{t} \in A_{t} \bigg)
\end{align*}

\noindent Hence, we conclude that $(M_t)_{\geq 1}$ is also a Markov chain.

\end{document}